\documentclass{optica-article}

\journal{opticajournal} 

\articletype{Research Article}
\usepackage{comment}
\usepackage{xcolor}
\usepackage{float}
\usepackage{graphicx,subfigure}

\begin{document}

\title{Efficient Minimal Solvers for Relative Pose Estimation in Autonomous Driving Applications}

\author{Tao Li,\authormark{1} Liang Liu,\authormark{1,*} Jianli Han,\authormark{1} and Weimin Lv\authormark{1}}

\address{\authormark{1}College of Aerospace Science and Engineering, Naval Aviation University, Yantai 264000, China}

\email{\authormark{*}jayli1585@2925.com}

\begin{abstract*} 
With the advancement of visual sensing systems, computer vision is playing an increasingly important role in autonomous driving and robot navigation. Relative pose estimation in multi-camera systems is essential for accurate vehicle localization and environment perception, demanding high real-time performance and robustness. Existing methods, however, often involve high computational costs and rely heavily on abundant feature matches, limiting their applicability in time-sensitive driving scenarios. To address these limitations, this paper introduces a unified framework for efficient relative pose estimation, built upon a novel translation parameterization and first-order rotation approximation. Within this framework, we propose three efficient minimal solvers specifically designed for autonomous vehicles. The first solver integrates the vertical direction prior from Inertial Measurement Units (IMUs), the second utilizes the rotation axis direction prior during steering maneuvers, and the third is designed for planar motion - a realistic assumption for ground vehicles operating on structured roads. By reducing both the minimal number of point correspondences and the algebraic complexity, our methods enable faster hypothesis generation within RANSAC-based pipelines, improving suitability for real-time systems. Extensive experiments on synthetic datasets and the KITTI autonomous driving benchmark demonstrate that the proposed solvers achieve a favorable balance between speed and accuracy compared to existing state-of-the-art algorithms.
\end{abstract*}

\section{Introduction}
\begin{figure}[H]      
     \centering    \includegraphics[width=0.7\linewidth]{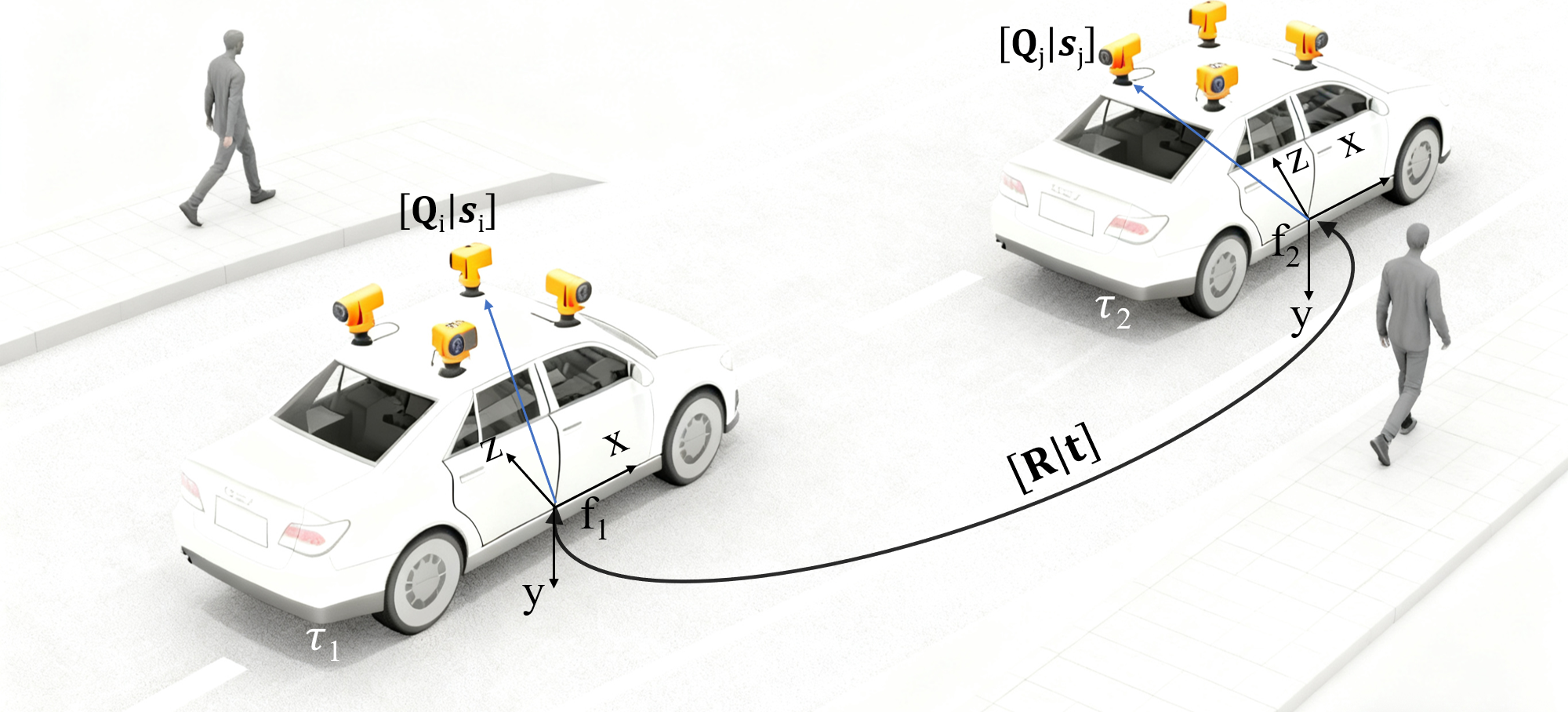}
    \centering
    \caption{The relative pose estimation for multi-camera systems.}
    \label{Multi-Camera}
\end{figure}

The rapid advancement of autonomous systems, such as unmanned aerial vehicles and self-driving cars, has spurred extensive use of camera-based perception. Cameras are typically mounted on these moving platforms to capture environmental information, enabling tasks like navigation, obstacle avoidance, and mapping, thereby supporting fully autonomous operation. In this context, accurate relative pose estimation between consecutive camera views is a critical capability, with key applications in simultaneous localization and mapping (SLAM)~\cite{campos2021orb}, structure from motion (SfM)~\cite{jiang2020efficient}, and autonomous driving~\cite{hee2013motion,huang2025ridge}.
Beyond these applications, pose estimation is also fundamental to other vision tasks including human pose estimation in video \cite{xiang2024dbmht,li2024lmformer} and non-cooperative object tracking \cite{deng2023flexible}, where similar geometric constraints and minimal solver principles are widely employed.
However, in real-world scenarios~\cite{Guan2026DMD,guan2026fusion}, the presence of inaccurate feature correspondences and outliers often severely degrades the robustness of estimation algorithms. To mitigate this, researchers often combine pose estimation with robust frameworks~\cite{fischler1981random,barath2019progressive,barath2020magsac++} to filter out incorrect matches.
The RANSAC~\cite{fischler1981random} algorithm is the most widely used robust framework due to its generality, but its computational cost grows exponentially with the number of matches required by the minimal solver. As a result, traditional methods that depend on more matches incur high computational overhead, limiting their use in real-time systems. This challenge is especially acute in autonomous driving, where dynamic objects like vehicles and pedestrians introduce mismatches, while the system still needs to deliver highly reliable pose estimation under strict timing constraints, as shown in Fig.~\ref{Multi-Camera}. Conventional approaches often struggle to balance accuracy and efficiency under such conditions. Consequently, developing efficient minimal solvers that require very few matches is crucial for enhancing real-time performance in autonomous driving applications.

Compared to monocular systems~\cite{TrifocalTensor2025,huang2025ridge}, multi-camera configurations~\cite{leiAMS2025,tan2026optimal} offer distinct advantages for pose estimation in autonomous driving. 
The most significant advantage lies in their expanded visual coverage and enhanced environmental perception capabilities, a crucial requirement for safe navigation in complex urban environments. The wide field of view provided by multiple cameras ensures robust performance even when individual cameras experience partial occlusion.
 Another critical advantage is their capacity for scale determination. The fixed distances among pre-calibrated cameras can naturally resolve the scale ambiguity problem inherent in monocular methods.
 Additionally, multi-camera systems provide significant architectural flexibility, allowing users to adjust both the number of cameras and their positioning and orientation to meet various needs and application requirements.

Despite their advantages, multi-camera systems introduce significant computational challenges distinct from monocular approaches. Unlike monocular systems, where all light rays converge at a single optical center, multi-camera systems exhibit greater geometric complexity due to their multiple projection centers \cite{pless2003using}. 
This structural difference substantially complicates relative pose estimation.
Existing algorithms for multi-camera systems, such as the minimal 6-point method \cite{henrikstewenius2005solutions} and the linear 17-point method \cite{li2008linear}, struggle to meet the demands of real-world autonomous driving applications. 
The 6-point method, while requiring fewer correspondences, suffers from computational inefficiency due to its reliance on Gröbner basis solvers, resulting in slow per-iteration speeds. 
On the other hand, the 17-point method, despite its linear formulation, necessitates a minimum of 17 point correspondences. When combined with RANSAC, this high sample size leads to an exponential growth in required iterations, making it impractical for real-time processing.
Given these limitations, neither the 6-point nor the 17-point method can adequately satisfy the stringent demands of autonomous driving applications.

This paper therefore focuses on efficient relative pose estimation for multi-camera systems in autonomous driving scenarios. In real-world driving, structural priors are often readily available, such as known vertical directions (typically from IMUs), known instantaneous rotation axis directions (during turns), and planar motion (on structured roads). Exploiting these practical and commonly encountered constraints allows for the development of efficient and robust solvers suitable for real-time deployment.
The main contributions of this work are summarized as follows:
\begin{itemize}
\item We propose a unified framework for multi-camera systems based on depth-parameterized translation and first-order rotation approximation. Within this framework, we reformulate both the generalized epipolar constraint and the epipolar constraint.
\item We introduce three novel minimal solvers: a 4-point solver with known vertical directions, a 4-point solver with known rotation axis directions, and a 3-point solver for planar motion scenarios. Each solver involves only low-degree polynomials, ensuring high efficiency.
\item Comprehensive evaluation on both synthetic and real-world datasets demonstrates that the proposed solvers achieve high computational speed while maintaining competitive estimation accuracy and robustness.
\end{itemize}

The rest of this paper is organized as follows. Section \ref{RELATED_WORK} presents a comprehensive review of state-of-the-art relative pose estimation methods for multi-camera systems. Section \ref{theory} establishes the theoretical foundation by rigorously deriving both the generalized epipolar constraint and the epipolar constraint, based on depth-based translation parameterization under general motion conditions. Section \ref{ourmethod} details our core methodological contributions: three novel solvers for multi-camera systems. Section \ref{sec:Degeneracy} analyzes the degenerate configurations of the proposed algorithms. 
Section \ref{Experiment} conducts an extensive experimental evaluation, validating the proposed methods through systematic benchmarks on synthetic data and real-world datasets. The paper concludes with a summary in Section \ref{Conclusion}.

\section{Related Work}\label{RELATED_WORK}
Unlike monocular systems, multi-camera systems lack a common center of projection, making the classical pinhole camera model inapplicable. Instead, the generalized camera model (GCM) provides a more suitable framework, parameterizing image pixels as spatial rays to represent non-central projection geometry \cite{pless2003using}. 
Research in multi-camera relative pose estimation can be categorized into three main directions according to the available degrees of freedom (DoF) and prior constraints: general 6-DoF solvers, 4-DoF solvers with partial prior information, and specialized planar motion solvers.

\textbf{General 6-DoF Solvers.}
General 6-DoF solvers address the full relative pose estimation problem without relying on prior motion constraints. Early work by Stewénius et al. proposed the first minimal 6-point solver based on Gröbner-basis theory. However, this approach faces notable practical challenges, as it can produce up to 64 potential solutions with slow processing speeds, resulting in high computational burdens \cite{henrikstewenius2005solutions}.
To mitigate the challenges of solution multiplicity and computational complexity, Ventura et al. developed a 6-point solver incorporating first-order rotation approximation, which substantially simplifies the problem while ensuring stability and efficiency \cite{ventura2015efficient}. 
Most recently, Guan et al. further proposed an improved 6-point solver incorporating ray bundle constraints, significantly reducing the solution space without sacrificing the method's generality \cite{guan_IJCV25}. By exhaustively enumerating the topological configurations of camera network observations, they established a geometric model linking platform pose deformation to point correspondences and revealed internal ray bundle constraints within the observation equations. This innovative approach effectively addresses the previously unsolvable problem of pose deformation estimation under non-cross-view configurations, marking a significant advancement with an improvement of over 21\% in accuracy for robust image measurement on dynamic platforms.
 Beyond minimal solvers, Li et al. improved the 17-point linear method through a comprehensive analysis of camera configurations, enabling effective handling of degenerate cases, including locally central and axial camera configurations \cite{li2008linear}.
Kim et al. further investigated the degradation of the 17-point method by decomposing the measurement matrix into ray direction and projection center components \cite{kim2010degeneracy}, while Chen et al. demonstrated the impact of visual overlap on the solvability and precision of the 17-point method \cite{xie202417}.
Complementing these point-correspondence methods, recent work has demonstrated that affine correspondences can also offer superior empirical performance in various scenarios\cite{guan2022affine, guan2021minimal}. 
In particular, by leveraging leveraging intrinsic affine constraints of camera networks, a recent breakthrough enables robust large-depth motion parameter estimation under weak geometric conditions \cite{guan_TPAMI26a}. This work pioneers the theoretical modeling of two novel independent geometric constraint equations, achieving over 38\% improvement in rotation and translation accuracy for non-cooperative targets, and further demonstrates the power of affine correspondences in relative pose estimation.

\textbf{4-DoF Solvers with Partial Prior Information.}
When partial orientation information is available, such as known vertical directions or known rotation axis directions, the problem can be reduced to 4-DoF, leading to more efficient solvers. For multi-camera systems with known vertical direction, Lee et al. developed a 4-point minimal solver that reduces the problem to solving an 8th-degree univariate polynomial \cite{hee2014relative}. 
Building on this, Liu et al. developed a more efficient 4-point algorithm using first-order rotation approximation, further simplifying the problem to a 4th-degree polynomial solution and enabling real-time performance for urban driving applications \cite{liu2017robust}.Similarly, for systems with known rotation axis direction, Sweeney et al. presented another 4-point algorithm that similarly reduces to solving an 8th-degree univariate polynomial \cite{sweeney2014solving}. 

\textbf{Planar Motion Solvers.}
Planar motion, commonly encountered in ground vehicles, further reduces the DoF of the relative pose problem. For monocular camera systems under planar motion~\cite{JEXK202411004}, scale ambiguity reduces the problem to 2-DoF.
Early work by Ortin et al. proposed methods suitable for indoor robots \cite{ortin2001indoor}. Subsequently, Choi et al. introduced the classic 2-point algorithm, which transforms the originally complex problem of solving the intersection of two ellipses into the simpler problem of finding the intersection of a line and a unit circle, enabling fast and reliable minimal solutions \cite{choi2018fast}. Dai et al. explored an alternative approach using the trifocal tensor for estimation, offering a different paradigm to address the problem~\cite{dai2025three}.
In multi-camera systems, scale becomes observable, resulting in a 3-DoF problem. Lee et al. proposed an efficient 2-point method based on the Ackermann motion model, which is commonly used in autonomous vehicles~\cite{hee2013motion}. Guan et al. further demonstrated that a single affine correspondence is sufficient to estimate the relative pose of multi-camera systems, since one affine correspondence can provide three geometric constraints~\cite{guan2023minimal}. 

\section{Generalized Camera Constraints under General Motion}\label{theory}
This section presents the geometric foundation for relative pose estimation in multi-camera systems. We adopt a depth-based translation parameterization, originally inspired by the generalized camera model~\cite{henrikstewenius2005solutions} and extended in recent affine-based methods~\cite{guan_TPAMI26a}. This formulation enables a unified derivation of both the generalized epipolar and epipolar constraints, which underpin the efficient solvers developed in Section~\ref{ourmethod}.
\subsection{Parameterization}
Consider a motion platform equipped with multiple cameras that moves from time $\tau_1$ to $\tau_2$. A 3D point $\mathbf{P}$ is observed by camera $\mathbf{c}_i$ at $\tau_1$ and by camera $\mathbf{c}_j$ at $\tau_2$.
The platform coordinate frames at $\tau_1$ and $\tau_2$ are denoted as $\mathbf{f}_1$ and $\mathbf{f}_2$, respectively. For each relative pose estimation, we define a local world coordinate system $\mathbf{W}$ with its origin placed at the observed 3D point $\mathbf{P}$ and its axes aligned with the initial platform frame $\mathbf{f}_1$. Thus, $\mathbf{f}_1$ and $\mathbf{W}$ differ only by a translation offset. In practice, a high-quality feature point (e.g., from SIFT~\cite{lowe2004distinctive} matching) is selected as $\mathbf{P}$ for each frame pair, and this reference point may change for subsequent estimations, as only the relative transformation, not absolute localization, is needed.
 \begin{figure}[H]              
     \centering    \includegraphics[width=0.6\linewidth]{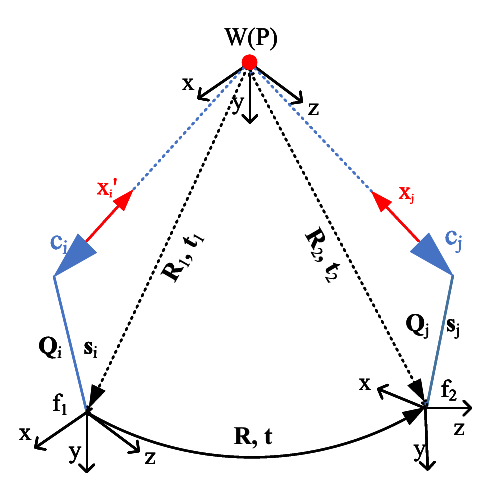}
    \centering
    \caption{The definition of coordinate systems in multi-camera systems.}
    \label{Coordinate}
\end{figure}

 In the multi-camera setup, each camera $\mathbf{c}_k$ has a pre-calibrated extrinsic transformation relative to the platform frame, consisting of rotation $\mathbf{Q}_k$ and translation $\mathbf{s}_k$. For time $\tau_1$, the transformation from camera $\mathbf{c}_i$ to platform frame $\mathbf{f}_1$ is given by $\mathbf{T}_{\mathbf{f}1\mathbf{c}i}=\left[ \begin{matrix} \mathbf{Q}_i & \mathbf{s}_i \\\end{matrix} \right]$. Similarly, for $\tau_2$, the transformation from frame $\mathbf{c}_j$ to frame $\mathbf{f}_2$ is $\mathbf{T}_{\mathbf{f}2\mathbf{c}j}=\left[ \begin{matrix} \mathbf{Q}_j & \mathbf{s}_j \\\end{matrix} \right]$.
 We denote the transformation from the platform frame to the world frame at $\tau_1$ as $\mathbf{T}_{\mathbf{W} \mathbf{f}_1} = [\mathbf{R}_1, \mathbf{t}_1]$, and at $\tau_2$ as $\mathbf{T}_{\mathbf{W} \mathbf{f}_2} = [\mathbf{R}_2, \mathbf{t}_2]$. Due to the axes alignment between $\mathbf{W}$ and $\mathbf{f}_1$, we have $\mathbf{R}_1 = \mathbf{I}$. The platform's motion from $\mathbf{f}_1$ to $\mathbf{f}_2$ is described by $\mathbf{T}_{\mathbf{f}_2 \mathbf{f}_1} = [\mathbf{R}, \mathbf{t}]$, where $\mathbf{R}$ and $\mathbf{t}$ represent the rotation and translation of the platform between $\tau_1$ and $\tau_2$. It is important to note that $\mathbf{R}$ and $\mathbf{t}$ correspond to the platform's ego-motion, not the extrinsic calibration between different cameras. The overall coordinate relationships are illustrated in Fig. \ref{Coordinate}.

Based on the coordinate systems defined above, the transformation from frame $\mathbf{f}_1$ to frame $\mathbf{f}_2$ can be expressed as:
\begin{equation}
	\begin{aligned}
   \mathbf{\bar{T}}_{\mathbf{f}2\mathbf{f}1}&={{\mathbf{\bar{T}}}_{\mathbf{f}2\mathbf{W}}}{{\mathbf{\bar{T}}}_{\mathbf{W}\mathbf{f}1}}=\left[ \begin{matrix}
   {{\mathbf{R}}_{2}} & {{\mathbf{t}}_{2}}  \\
   \mathbf{0} & 1  \\
\end{matrix} \right]\left[ \begin{matrix}
   \mathbf{I} & -{{\mathbf{t}}_{1}}  \\
   \mathbf{0} & 1  \\
\end{matrix} \right] =\left[ \begin{matrix}
   {{\mathbf{R}}_{2}} & -{{\mathbf{R}}_{2}}{{\mathbf{t}}_{1}}+{{\mathbf{t}}_{2}}  \\
   \mathbf{0} & 1  \\
\end{matrix} \right]
        \end{aligned}.
	\label{a1}
\end{equation}
 $\mathbf{\bar{T}}$ denotes the homogeneous representation of the transformation matrix $\mathbf{T}$, constructed by appending the row $[0,0,0,1]$. Based on \eqref{a1}, we can obtain:
\begin{equation}
	\begin{aligned}
   & \mathbf{R}={{\mathbf{R}}_{2}} \\ 
  & \mathbf{t}=-{{\mathbf{R}}_{2}}{{\mathbf{t}}_{1}}+{{\mathbf{t}}_{2}} \\ 
        \end{aligned}.
	\label{a2}
\end{equation}

In the generalized camera model, a correspondence between image points $\mathbf{u}_i$ and $\mathbf{u}_j$ from the same 3D point can be interpreted as the intersection of two Plücker lines. Each line, defined in its respective platform frame, is represented as ${{\mathbf{L}}_{*}}={{\left[   \begin{matrix}   {{\mathbf{x}}_{*}}^{\mathrm{T}} & {{\mathbf{q}}_{*}}^{\mathrm{T}}  \\\end{matrix} \right]}^{\mathrm{T}}}$, where the subscript $*$ denotes the camera index $i$ or $j$. 
The unit direction vector $\mathbf{x}_{*}$ is computed as
$\mathbf{x}_{*} = (\mathbf{Q}_{*} \mathbf{u}_{*})/\begin{Vmatrix}\mathbf{Q}_{*} *\mathbf{u}_{*}\end{Vmatrix}$, and the moment vector is given by ${{\mathbf{q}}_{*}}={{\mathbf{s}}_{*}}\times {{\mathbf{x}}_{*}}$.
According to Plücker line geometry,
any point $\mathbf{U}(\lambda)$ on such a Plücker line can be parameterized as:
\begin{equation}
	\begin{aligned}
\mathbf{U}(\lambda )={{\mathbf{x}}}\times {{\mathbf{q}}}+{{\lambda }}{{\mathbf{x}}},\lambda \in \mathbb{R}
\end{aligned}.
	\label{a3}
\end{equation}

For the world origin $\mathbf{P} = [0,0,0]^\mathrm{T}$, its corresponding Plücker lines are denoted as $\mathbf{L}_1$ and $\mathbf{L}_2$, with direction vectors $\mathbf{x}_1$, $\mathbf{x}_2$ and moment vectors $\mathbf{q}_1$, $\mathbf{q}_2$. Since $\mathbf{P}$ lies at the intersection of these two lines, its coordinates expressed in platform frames $\mathbf{f}_1$ and $\mathbf{f}_2$ must satisfy both line parameterizations:
\begin{equation}
	\begin{aligned}
\mathbf{U}(\lambda_1 )&={{\mathbf{x}}_{1}}\times {{\mathbf{q}}_{1}}+{{\lambda }_{1}}{{\mathbf{x}}_{1}}={{\mathbf{R}}_{1}}{[0, 0, 0]^{\mathrm{T}}}+{{\mathbf{t}}_{1}}\\
\mathbf{U}(\lambda_2 )&={{\mathbf{x}}_{2}}\times {{\mathbf{q}}_{2}}+{{\lambda }_{2}}{{\mathbf{x}}_{2}}={{\mathbf{R}}_{2}}{[0, 0, 0]^{\mathrm{T}}}+{{\mathbf{t}}_{2}}
\end{aligned}.
	\label{a4}
\end{equation}
It should be noted that $\lambda_1$ and $\lambda_2$ here do not represent the depth of $\mathbf{P}$ in the camera frame (as in a depth camera model), but rather denote the distances from the orthogonal projections of the origins of $\mathbf{f}_1$ and $\mathbf{f}_2$ onto lines $\mathbf{L}_1$ and $\mathbf{L}_2$, respectively, to the point $\mathbf{P}$.

From \eqref{a4}, the translation vectors $\mathbf{t}_1$ and $\mathbf{t}_2$ can be expressed as:
\begin{equation}
	\begin{aligned}
{{\mathbf{t}}_{1}}&={{\mathbf{x}}_{1}}\times {{\mathbf{q}}_{1}}+{{\lambda }_{1}}{{\mathbf{x}}_{1}}\\
{{\mathbf{t}}_{2}}&={{\mathbf{x}}_{2}}\times {{\mathbf{q}}_{2}}+{{\lambda }_{2}}{{\mathbf{x}}_{2}}
\end{aligned}.
	\label{a5}
\end{equation}
This shows that the translation vector can be parameterized by two depth variables $\lambda_1$ or $\lambda_2$, rather than by the conventional Cartesian coordinates ${[t_x, t_y, t_z]}^{\mathrm{T}}$. 
Compared to the Cartesian representation, the depth-based parameterization offers two main advantages: (1) it directly relates translation to observable image features, enabling clearer geometric interpretation; (2) it reduces the number of unknowns, using only two scalar depth parameters rather than three translation components. This reduction is particularly beneficial for developing efficient minimal solvers, as it decreases the complexity of the underlying polynomial systems to be solved. The main limitation is its dependence on the accuracy of the reference 3D point (the world origin $\mathbf{P}$), whose correspondence errors may propagate into depth estimates. In practice, this sensitivity is mitigated by selecting high-quality feature points as the reference and integrating the solver within a robust estimation framework. Based on this parameterization, we now derive extended formulations of the generalized epipolar constraint and the epipolar constraint tailored for multi-camera configurations.

\subsection{Generalized Epipolar Constraint}
As ${{\mathbf{L}}_{i}}$ and ${{\mathbf{L}}_{j}}$ are defined in distinct coordinate frames $f_1$ and $f_2$ respectively, a coordinate transformation is required to express them in a common frame. Specifically, when transforming ${{\mathbf{L}}_{i}}$ from frame $f_1$ to frame $f_2$, the resulting line $\mathbf{L}_{i}^{\prime}={{\left[ \begin{matrix}{{\mathbf{x}}_{i}}{{^{\prime }}^{\mathrm{T}}} & {{\mathbf{q}}_{i}}{{^{\prime }}^{\mathrm{T}}}  \\\end{matrix} \right]}^{\mathrm{T}}}$ is obtained via the following transformation:
\begin{equation}
	\begin{aligned} \mathbf{L}_{i}^{\prime}=\left[ \begin{matrix}
   \mathbf{R} & \mathbf{0}  \\
   {{[\mathbf{t}]}_{\times }}\mathbf{R} & \mathbf{R}  \\
\end{matrix} \right]{{\mathbf{L}}_{i}}=\left[ \begin{matrix}
   \mathbf{R}{{\mathbf{x}}_{i}}  \\
   \mathbf{R}{{\mathbf{q}}_{i}}+{{[\mathbf{t}]}_{\times }}\mathbf{R}{{\mathbf{x}}_{i}}  \\
\end{matrix} \right],
\end{aligned}
	\label{b0}
\end{equation}
 where ${{[\mathbf{t}]}_{\times }}$ denotes the skew-symmetric matrix associated with the translation vector $\mathbf{t}$.
 Referencing \cite{pless2003using}, the necessary and sufficient condition for the intersection of lines $\mathbf{L}_{i}^{\prime}$ and ${{\mathbf{L}}_{j}}$ within the common coordinate frame $f_2$ is:
\begin{equation}
	\begin{aligned}{{\mathbf{x}}_{i}}{{^{\prime }}^{\mathrm{T}}}{{\mathbf{q}}_{j}}+{{\mathbf{x}}_{j}}^{\mathrm{T}}{{\mathbf{q}}_{i}}^{\prime }=0
\end{aligned}.
	\label{b1}
\end{equation}

Substituting \eqref{b0} into \eqref{b1} yields the generalized epipolar constraint (GEC):
\begin{equation}
	\begin{aligned}{(\mathbf{R}{{\mathbf{x}}_{i}})}^{\mathrm{T}}{{\mathbf{q}}_{j}}+{{{\mathbf{x}}_{j}}^{\mathrm{T}}}\mathbf{R}{{\mathbf{q}}_{i}}+{{{\mathbf{x}}_{j}}^{\mathrm{T}}}{{[\mathbf{t}]}_{\times }}\mathbf{R}{{\mathbf{x}}_{i}}=0.
    \end{aligned}
	\label{b2}
\end{equation}
Incorporating \eqref{a2} into \eqref{b2} gives:
\begin{equation}
	\begin{aligned}
{{\mathbf{x}}_{i}}^{\mathrm{T}}{{\mathbf{R}}^{\mathrm{T}}}{{\mathbf{q}}_{j}}+{{\mathbf{x}}_{j}}^{\mathrm{T}}\mathbf{R}{{\mathbf{q}}_{i}}+{{\mathbf{x}}_{j}}^{\mathrm{T}}({{[{{\mathbf{t}}_{2}}]}_{\times }}\mathbf{R}-\mathbf{R}{{[{{\mathbf{t}}_{1}}]}_{\times }}){{\mathbf{x}}_{i}}=0
    \end{aligned}.
	\label{b3}
\end{equation}
Finally, substituting \eqref{a5} into \eqref{b3} yields the depth-parameterized formulation of the GEC:
\begin{equation}
	\begin{aligned}
      & -{{\lambda }_{1}}{{\mathbf{x}}_{j}}^{\mathrm{T}}\mathbf{R}{{[{{\mathbf{x}}_{1}}]}_{\times }}{{\mathbf{x}}_{i}}+{{\lambda }_{2}}{{\mathbf{x}}_{j}}^{\mathrm{T}}({{[{{\mathbf{x}}_{2}}]}_{\times }}\mathbf{R}){{\mathbf{x}}_{i}}+{{\mathbf{x}}_{j}}^{\mathrm{T}}\mathbf{R}{{\mathbf{q}}_{i}} +  \\ 
 & {{\mathbf{x}}_{i}}^{\mathrm{T}}{{\mathbf{R}}^{\mathrm{T}}}{{\mathbf{q}}_{j}}+{{\mathbf{x}}_{j}}^{\mathrm{T}}({{[{{\mathbf{x}}_{2}}\times {{\mathbf{q}}_{2}}]}_{\times }}\mathbf{R}-\mathbf{R}{{[{{\mathbf{x}}_{1}}\times {{\mathbf{q}}_{1}}]}_{\times }}){{\mathbf{x}}_{i}}=0 \\ 
    \end{aligned}.
	\label{b4}
\end{equation}
Here, $\mathbf{x}_i$, $\mathbf{x}_j$, $\mathbf{q}_i$, and $\mathbf{q}_j$ correspond to any non-origin 3D point, while $\mathbf{x}_1$, $\mathbf{x}_2$, $\mathbf{q}_1$, and $\mathbf{q}_2$ specifically denote the direction vectors and moment vectors of the Plücker lines associated with the world origin point $\mathbf{P}$.
\subsection{Epipolar Constraint}
Using the same depth parameterization, we can also derive the epipolar constraint directly in terms of $\lambda_1$ and $\lambda_2$. The relative transformation between cameras $\mathbf{c}_i$ and $\mathbf{c}_j$ is:
\begin{equation}
	\begin{aligned}
{{{\mathbf{\bar{T}}}}_{\mathbf{c}j\mathbf{c}i}}& ={{{\mathbf{\bar{T}}}}_{\mathbf{c}j\mathbf{f}2}}\mathbf{\bar{T}}_{\mathbf{f}2\mathbf{f}1}{{{\mathbf{\bar{T}}}}_{\mathbf{f}1\mathbf{c}i}} =\left[ \begin{matrix}
   {{\mathbf{Q}}_{j}}^{\mathrm{T}} & -{{\mathbf{Q}}_{j}}^{\mathrm{T}}{{\mathbf{s}}_{j}}  \\
   \mathbf{0} & 1  \\
\end{matrix} \right]\left[ \begin{matrix}
   {{\mathbf{R}}} & {{\mathbf{t}}}  \\
   \mathbf{0} & 1  \\
\end{matrix} \right]\left[ \begin{matrix}
   {{\mathbf{Q}}_{i}} & {{\mathbf{s}}_{i}}  \\
   \mathbf{0} & 1  \\
\end{matrix} \right] \\ 
 & =\left[ \begin{matrix}
   {{\mathbf{Q}}_{j}}^{\mathrm{T}}{{\mathbf{R}}}{{\mathbf{Q}}_{i}} & {{\mathbf{Q}}_{j}}^{\mathrm{T}}({{\mathbf{R}}}{{\mathbf{s}}_{i}}+{{\mathbf{t}}}-{{\mathbf{s}}_{j}})  \\
   \mathbf{0} & 1  \\
\end{matrix} \right] \\ 
        \end{aligned}.
	\label{c1}
\end{equation}
Combining \eqref{a2} with \eqref{c1}, the rotation and translation components are derived as:
\begin{equation}
	\begin{aligned}
& {{\mathbf{R}}_{\mathbf{c}j\mathbf{c}i}}={{\mathbf{Q}}_{j}}^{\mathrm{T}}{{\mathbf{R}}}{{\mathbf{Q}}_{i}} \\ 
 & {{\mathbf{t}}_{\mathbf{c}j\mathbf{c}i}}={{\mathbf{Q}}_{j}}^{\mathrm{T}}({{\mathbf{R}}}({{\mathbf{s}}_{i}}-{{\mathbf{t}}_{1}})+{{\mathbf{t}}_{2}}-{{\mathbf{s}}_{j}}) \\ 
        \end{aligned}.
	\label{c2}
\end{equation}
Then the corresponding essential matrix 
${{\mathbf{E}}_{\mathbf{c}j\mathbf{c}i}}={{[{{\mathbf{t}}_{\mathbf{c}j\mathbf{c}i}}]}_{\times }}{{\mathbf{R}}_{\mathbf{c}j\mathbf{c}i}}$ can be expressed as:
\begin{equation} 
\begin{aligned} 
	{\mathbf{E}}_{\mathbf{c}j\mathbf{c}i} &= {{\mathbf{Q}}_{j}}^{\mathrm{T}} \left({{\mathbf{R}}} [{{\mathbf{s}}_{i}} - \mathbf{t}_1]_\times + [\mathbf{t}_2 - {{\mathbf{s}}_{j}}]_\times {{\mathbf{R}}} \right) {{\mathbf{Q}}_{i}}. 
	\label{eq:essential_matrix}
\end{aligned}
\end{equation}

Substituting \eqref{a5} into \eqref{c2} then yields the following expression:
\begin{equation}
	\begin{aligned} 
  {{\mathbf{t}}_{\mathbf{c}j\mathbf{c}i}}=&{{\mathbf{Q}}_{j}}^{\mathrm{T}}({{\mathbf{R}}}({{\mathbf{s}}_{i}}-{{\mathbf{x}}_{1}}\times {{\mathbf{q}}_{1}}-{{\lambda }_{1}}{{\mathbf{x}}_{1}})+{{\mathbf{x}}_{2}}\times {{\mathbf{q}}_{2}}+{{\lambda }_{2}}{{\mathbf{x}}_{2}}-{{\mathbf{s}}_{j}}) \\ 
        \end{aligned}.
	\label{c3}
\end{equation}
The essential matrix ${{\mathbf{E}}_{\mathbf{c}j\mathbf{c}i}}$ between cameras $\mathbf{c}_i$ and $\mathbf{c}_j$ can then be written as:
\begin{equation}
	\begin{aligned}
{{\mathbf{E}}_{\mathbf{c}j\mathbf{c}i}}&=-\lambda_{1}\mathbf{Q}_{j}^{\mathrm{T}}\mathbf{R}[\mathbf{x}_{1}]_{\times}\mathbf{Q}_{i}+\lambda_{2}\mathbf{Q}_{j}^{\mathrm{T}}[\mathbf{x}_{2}]_{\times}\mathbf{R}\mathbf{Q}_{i}\\&+ \mathbf{Q}_{j}^{\mathrm{T}}\left(\mathbf{R}[\mathbf{s}_{i}-\mathbf{x}_{1}\times\mathbf{q}_{1}]_{\times}+[\mathbf{x}_{2}\times\mathbf{q}_{2}-\mathbf{s}_{j}]_{\times}\mathbf{R}\right)\mathbf{Q}_{i}
\end{aligned}.
	\label{c4}
\end{equation}
Applying the epipolar constraint $\mathbf{u}_{j}^{\mathrm{T}}{{\mathbf{E}}_{\mathbf{c}j\mathbf{c}i}}{{\mathbf{u}}_{i}}=0$ gives the depth-parameterized epipolar constraint:
\begin{equation}
	\begin{aligned}
&-\lambda_{1}\mathbf{u}_{j}^{\mathrm{T}}\mathbf{Q}_{j}^{\mathrm{T}}\mathbf{R}[\mathbf{x}_{1}]_{\times}\mathbf{Q}_{i}{{\mathbf{u}}_{i}}+\lambda_{2}\mathbf{u}_{j}^{\mathrm{T}}\mathbf{Q}_{j}^{\mathrm{T}}[\mathbf{x}_{2}]_{\times}\mathbf{R}\mathbf{Q}_{i}{{\mathbf{u}}_{i}}+\\& \mathbf{u}_{j}^{\mathrm{T}}\mathbf{Q}_{j}^{\mathrm{T}}\left(\mathbf{R}[\mathbf{s}_{i}-\mathbf{x}_{1}\times\mathbf{q}_{1}]_{\times}+[\mathbf{x}_{2}\times\mathbf{q}_{2}-\mathbf{s}_{j}]_{\times}\mathbf{R}\right)\mathbf{Q}_{i}{{\mathbf{u}}_{i}}=0
\end{aligned}.
	\label{c5}
\end{equation}

Although Equations \eqref{b4} and \eqref{c5} exhibit different forms, they are mathematically equivalent. The primary distinction lies in how point correspondences are represented within the generalized epipolar constraint versus the epipolar constraint. In the generalized epipolar constraint, all point correspondences are expressed using Plücker line representation. Conversely, for the epipolar constraint, only the correspondence of the world origin $\mathbf{P}$ is represented by the Plücker line representation, while other point correspondences utilize conventional normalized image coordinates.

\section{Proposed Minimal Solvers}\label{ourmethod}
This section introduces three novel minimal solvers for generalized camera pose estimation, each exploiting a specific motion prior. Sections \ref{sec:4pt_vertical} and \ref{sec:4pt_axis} present 4-point solvers for cases with known vertical direction and rotation axis direction, respectively. Section \ref{sec:3-Point} develops a 3-point solver under the assumption of planar motion. All solvers build upon the geometric constraints derived in Section \ref{theory} and employ a small-angle approximation to enable efficient polynomial solutions.
\subsection{The 4-Point solver with Known vertical direction}\label{sec:4pt_vertical} 
\begin{figure}[H]        
     \centering    \includegraphics[width=0.7\linewidth]{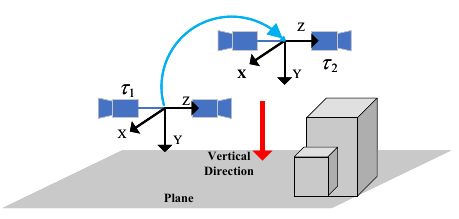}
    \centering
    \caption{The relative pose estimation for multi-camera systems with known vertical direction.}
    \label{Vertical-Direction}
\end{figure}

When the platform is equipped with IMUs, the platform's pitch and roll angles can be accurately estimated. Through IMU alignment, the platform's vertical direction (typically the Y-axis) is aligned with the direction of gravity, as shown in Fig.~\ref{Vertical-Direction}.

Let $\mathbf{R}_{imu1}$ and $\mathbf{R}_{imu2}$ denote the IMU-derived alignment matrices at time instances $\tau_1$ and $\tau_2$, respectively. The platform rotation matrix $\mathbf{R}$ relates to its gravity-aligned counterpart ${{\mathbf{R}}_{\mathbf{v}2\mathbf{v}1}}$ through the following expression:
\begin{equation}
	\begin{aligned}
{{\mathbf{R}}}={{\mathbf{R}}_{\mathbf{f}2\mathbf{f}1}}={{\mathbf{R}}_{imu2}}^{\mathrm{T}}{{\mathbf{R}}_{\mathbf{v}2\mathbf{v}1}}{{\mathbf{R}}_{imu1}}
\end{aligned}.
	\label{d0}
\end{equation}
The problem thus reduces to solving for $\mathbf{R}_{\mathbf{v}2\mathbf{v}1}$, which depends solely on the yaw angle $s$:
\begin{equation}
	\begin{aligned}
\mathbf{R}_{\mathbf{v}2\mathbf{v}1}=\left[ \begin{matrix}
   \cos(s)  & 0 & \sin(s)  \\
   0 & 1 & 0  \\
   -\sin(s)  & 0 & \cos(s)  \\
\end{matrix} \right]
\end{aligned}.
	\label{d1}
\end{equation}

Under high frame rate operation, the rotational displacement of the moving platform between consecutive frames is small, allowing for the application of first-order rotation approximation:
\begin{equation}
	\begin{aligned}
\mathbf{R}_{\mathbf{v}2\mathbf{v}1}=\left[ \begin{matrix}
   1  & 0 & s  \\
   0 & 1 & 0  \\
   -s  & 0 & 1  \\
\end{matrix} \right]
\end{aligned},
	\label{d1_1}
\end{equation}
where $\sin(s) \approx s$ and $\cos(s) \approx 1$.
By incorporating \eqref{d0} into the generalized epipolar constraint \eqref{b4}, we derive the following modified formulation:
\begin{equation}
	\begin{aligned}
      & -{{\lambda }_{1}}{{\mathbf{\tilde{x}}}_{j}}^{\mathrm{T}}\mathbf{R}_{\mathbf{v}2\mathbf{v}1}{{[{{\mathbf{\tilde{x}}}_{1}}]}_{\times }}{{\mathbf{\tilde{x}}}_{i}}+{{\lambda }_{2}}{{\mathbf{\tilde{x}}}_{j}}^{\mathrm{T}}({{[{{\mathbf{\tilde{x}}}_{2}}]}_{\times }}\mathbf{R}_{\mathbf{v}2\mathbf{v}1}){{\mathbf{\tilde{x}}}_{i}}+{{\mathbf{\tilde{x}}}_{j}}^{\mathrm{T}}\mathbf{R}_{\mathbf{v}2\mathbf{v}1}{{\mathbf{\tilde{q}}}_{i}} \\ 
 &+{{\mathbf{\tilde{x}}}_{i}}^{\mathrm{T}}{{\mathbf{R}_{\mathbf{v}2\mathbf{v}1}}^{\mathrm{T}}}{{\mathbf{\tilde{q}}}_{j}}+{{\mathbf{\tilde{x}}}_{j}}^{\mathrm{T}}({{[{{\mathbf{\tilde{x}}}_{2}}\times {{\mathbf{\tilde{q}}}_{2}}]}_{\times }}\mathbf{R}_{\mathbf{v}2\mathbf{v}1}-\mathbf{R}_{\mathbf{v}2\mathbf{v}1}{{[{{\mathbf{\tilde{x}}}_{1}}\times {{\mathbf{\tilde{q}}}_{1}}]}_{\times }}){{\mathbf{\tilde{x}}}_{i}}=0 \\ 
    \end{aligned}.
	\label{d2}
\end{equation}
Comparing \eqref{d2} with \eqref{b4} reveals their structural similarity, with the key distinction being that the variables in \eqref{d2} have been adjusted by alignment matrices. The detailed adjustments include:
${{\mathbf{\tilde{x}}}_{i}}={{\mathbf{R}}_{imu1}}{{\mathbf{x}}_{i}}$, ${{\mathbf{\tilde{x}}}_{j}}={{\mathbf{R}}_{imu2}}{{\mathbf{x}}_{j}}$, ${{\mathbf{\tilde{q}}}_{i}}={{\mathbf{R}}_{imu1}}{{\mathbf{q}}_{i}}$, ${{\mathbf{\tilde{q}}}_{j}}={{\mathbf{R}}_{imu2}}{{\mathbf{q}}_{j}}$.

Similarly, substituting \eqref{d0} into the epipolar constraint \eqref{c5}, we can obtain:
\begin{equation}
	\begin{aligned}
&-\lambda_{1}\mathbf{u}_{j}^{\mathrm{T}}\mathbf{\tilde{Q}}_{j}^{\mathrm{T}}\mathbf{R}_{\mathbf{v}2\mathbf{v}1}[\mathbf{\tilde{x}}_{1}]_{\times}\mathbf{\tilde{Q}}_{i}{{\mathbf{u}}_{i}}+\lambda_{2}\mathbf{u}_{j}^{\mathrm{T}}\mathbf{\tilde{Q}}_{j}^{\mathrm{T}}[\mathbf{\tilde{x}}_{2}]_{\times}\mathbf{R}_{\mathbf{v}2\mathbf{v}1}\mathbf{\tilde{Q}}_{i}{{\mathbf{u}}_{i}}\\&+ \mathbf{u}_{j}^{\mathrm{T}}\mathbf{\tilde{Q}}_{j}^{\mathrm{T}}(\mathbf{R}_{\mathbf{v}2\mathbf{v}1}[\mathbf{\tilde{s}}_{i}-\mathbf{\tilde{x}}_{1}\times\mathbf{\tilde{q}}_{1}]_{\times}+[\mathbf{\tilde{x}}_{2}\times\mathbf{\tilde{q}}_{2}-\mathbf{\tilde{s}}_{j}]_{\times}\mathbf{R}_{\mathbf{v}2\mathbf{v}1})\mathbf{\tilde{Q}}_{i}{{\mathbf{u}}_{i}}=0
\end{aligned}.
	\label{d3}
\end{equation}
In addition to the above adjustments specified in \eqref{d2}, we also need to make the following adjustments, including: 
${{\mathbf{\tilde{Q}}}_{i}}={{\mathbf{R}}_{imu1}}{{\mathbf{Q}}_{i}}$, 
${{\mathbf{\tilde{Q}}}_{j}}={{\mathbf{R}}_{imu2}}{{\mathbf{Q}}_{j}}$, ${{\mathbf{\tilde{s}}}_{i}}={{\mathbf{R}}_{imu1}}{{\mathbf{s}}_{i}}$, ${{\mathbf{\tilde{s}}}_{j}}={{\mathbf{R}}_{imu2}}{{\mathbf{s}}_{j}}$.

Substituting the linearized rotation \eqref{d1_1} into \eqref{d2} or \eqref{d3} yields a linear equation in the unknowns $s$, $\lambda_1$, and $\lambda_2$:
\begin{equation}
	\begin{aligned}
a_1+a_2s+a_3{\lambda }_{1}+a_4{\lambda }_{2}+a_5s{\lambda }_{1}+a_6s{\lambda }_{2}=0
    \end{aligned},
	\label{d3_1}
\end{equation}
where coefficients $a_1$ to $a_6$ are constructed from Plücker
line correspondences.
By leveraging three non-origin point correspondences with corresponding coefficients $a_0$ to $a_6$, $b_0$ to $b_6$, and $c_0$ to $c_6$, we construct a system of three equations based on \eqref{d2} or \eqref{d3}. The world origin correspondence $\mathbf{P}$ is excluded since its constraint equation reduces to 0, offering no meaningful contribution to parameter estimation. These selected equations can be compactly represented in matrix formulation as:
\begin{equation}
	\begin{aligned}
\underbrace{\begin{bmatrix}a_3+a_5s &a_4+a_6s &a_1+a_2s\\b_3+b_5s &b_4+b_6s &b_1+b_2s\\c_3+c_5s &c_4+c_6s &c_1+c_2s\end{bmatrix}}_{{{\mathbf{F}}_{3\times 3}}(s)}\left[ \begin{matrix}
   {{\lambda }_{1}}  \\
   {{\lambda }_{2}}  \\
   1  \\
\end{matrix} \right]={{\mathbf{0}}_{3\times 1}}
\end{aligned}.
	\label{d4}
\end{equation}
Given that the equation system \eqref{d4} has a non-trivial solution, the matrix $\mathbf{F}(s)$ must be rank-deficient with $\text{rank}(\mathbf{F}(s)) \leq 2$, which necessarily leads to the determinant condition:
\begin{equation}
	\begin{aligned}
\text{det}({\mathbf{F}}(s))= 0
\end{aligned}.
	\label{d5}
\end{equation}
This determinant equation yields a 3rd-degree univariate polynomial in $s$, potentially generating up to three distinct real roots.

Once the solutions for $s$ are obtained, substitute them back into \eqref{d1} rather than its simplified version \eqref{d1_1} to strictly maintain the orthogonality of the rotation matrix $\mathbf{R}_{\mathbf{v}2\mathbf{v}1}$. Then, substitute the resulting $\mathbf{R}_{\mathbf{v}2\mathbf{v}1}$ into \eqref{d0} to compute the complete rotation matrix $\mathbf{R}$.

\subsection{The 4-Point solver with Known Rotation Axis Direction}\label{sec:4pt_axis}
\begin{figure}[H]           
     \centering    \includegraphics[width=0.7\linewidth]{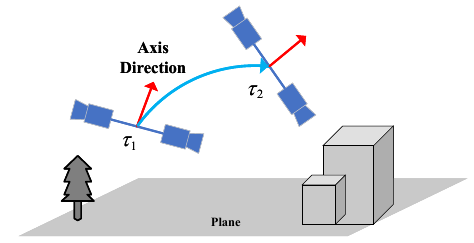}
    \centering
    \caption{The relative pose estimation for multi-camera systems with known rotation axis direction.}
    \label{Rotation-Axis}
\end{figure}

The rotation-axis-based approach provides another useful prior for relative pose estimation in autonomous driving. For instance, during steering maneuvers, the vehicle's rotational motion is largely confined to a specific axis determined by the steering geometry, as shown in Fig. \ref{Rotation-Axis}. In systems with steer-by-wire or well-defined kinematic models, this axis direction can be derived as a strong prior. Additionally, in platforms equipped with IMUs, the rotation axis direction can also be obtained through inertial measurements.

In this section, we adopt the axis-angle representation to describe rotation. 
This elegant representation uses two key components: a unit vector $\mathbf{n}$ specifying the rotation axis direction and a scalar angle $w$ representing the rotation angle magnitude. 
The combination $w\mathbf{n}$ provides a minimal parameterization that compactly encodes both the direction and magnitude of rotation.
 Given $\mathbf{n}$ and $w$, the corresponding rotation matrix $\mathbf{R}$ can be derived using the Rodrigues rotation formula:
\begin{equation}
\begin{aligned}
\mathbf{R} = \cos(w) \cdot \mathbf{I} + (1 - \cos(w)) \mathbf{n}\mathbf{n}^\top + \sin(w) \cdot [\mathbf{n}]_\times
\end{aligned}.
	\label{g1}
\end{equation}
For small rotation angles, we can simplify the Rodrigues formula by applying first-order approximations. When the rotation angle $w$ approaches zero, the trigonometric terms reduce to their linear approximations: $\sin(w) \approx w$ and $\cos(w) \approx 1$. Under this small-angle assumption, the Rodrigues formula simplifies to:

\begin{equation}
\begin{aligned}
\mathbf{R} \approx \mathbf{I} + w[\mathbf{n}]_\times
\end{aligned}.
	\label{g2}
\end{equation}
 
 Substituting \eqref{g2} into either the generalized epipolar constraint \eqref{b4} or the epipolar constraint \eqref{c5} yields a constraint linear in $w$, $\lambda_1$, and $\lambda_2$:
\begin{equation}
	\begin{aligned}
      & -{{\lambda }_{1}}{{\mathbf{x}}_{j}}^{\mathrm{T}}(\mathbf{I} + w[\mathbf{n}]_\times){{[{{\mathbf{x}}_{1}}]}_{\times }}{{\mathbf{x}}_{i}}+{{\lambda }_{2}}{{\mathbf{x}}_{j}}^{\mathrm{T}}({{[{{\mathbf{x}}_{2}}]}_{\times }}(\mathbf{I} + w[\mathbf{n}]_\times)){{\mathbf{x}}_{i}}\\ &+{{\mathbf{x}}_{j}}^{\mathrm{T}}(\mathbf{I} + w[\mathbf{n}]_\times){{\mathbf{q}}_{i}} +   {{\mathbf{x}}_{i}}^{\mathrm{T}}{{(\mathbf{I} + w[\mathbf{n}]_\times)}^{\mathrm{T}}}{{\mathbf{q}}_{j}}+{{\mathbf{x}}_{j}}^{\mathrm{T}} \\&({{[{{\mathbf{x}}_{2}}\times {{\mathbf{q}}_{2}}]}_{\times }}(\mathbf{I} + w[\mathbf{n}]_\times)-(\mathbf{I} + w[\mathbf{n}]_\times){{[{{\mathbf{x}}_{1}}\times {{\mathbf{q}}_{1}}]}_{\times }}){{\mathbf{x}}_{i}}=0 \\ 
    \end{aligned}.
	\label{g3}
\end{equation}
Building upon the formulations in \eqref{g3}, we can also establish a system from three non-origin point correspondences:
 \begin{equation}
	\begin{aligned}
{{\mathbf{F}}_{3\times 3}}(w)\left[ \begin{matrix}
   {{\lambda }_{1}}  \\
   {{\lambda }_{2}}  \\
   1  \\
\end{matrix} \right]={{\mathbf{0}}_{3\times 1}}
\end{aligned}.
	\label{g5}
\end{equation}
Analogous to \eqref{d5}, this system yields the determinant condition:
\begin{equation}
	\begin{aligned}
\text{det}({\mathbf{F}}(w))= 0
\end{aligned}.
	\label{g6}
\end{equation}

Consistent with the previous finding for $s$, equation \eqref{g6} produces a 3rd-degree univariate polynomial in $w$, which can have up to three real solutions.
By substituting the estimated rotation angle magnitude $w$ and the known axis direction $\mathbf{n}$ into Equation \eqref{g1}, we can reconstruct the rotation matrix $\mathbf{R}$.

\subsection{The 3-Point solver with Planar Motion Constraints}\label{sec:3-Point} 
\begin{figure}[H]       
     \centering    \includegraphics[width=0.7\linewidth]{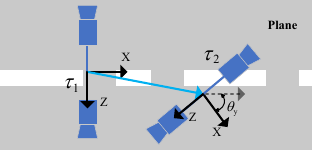}
    \centering
    \caption{The relative pose estimation for multi-camera systems with planar motion constraints.}
    \label{Planar-Motion}
\end{figure}

In typical autonomous driving scenarios, vehicles predominantly operate on structured roads or approximately flat ground. This motion characteristic provides an important prior constraint for pose estimation: the 6-DoF rigid-body motion of the platform can be simplified to a planar motion model. Specifically, only the yaw rotation around the vertical axis (Y-axis) is retained, while the translational motion is constrained to the horizontal plane (with zero vertical displacement, i.e., $t_y=0$), see Fig. \ref{Planar-Motion}. The rotation matrix $\mathbf{R}$ and translation vector $\mathbf{t}$ under planar motion constraints can be expressed as:
\begin{equation}
\mathbf{R}=\left[ \begin{matrix}
   \cos(\theta_y)  & 0 & \sin(\theta_y)  \\
   0 & 1 & 0  \\
   -\sin(\theta_y)  & 0 & \cos(\theta_y)  \\
\end{matrix} \right], \mathbf{t}=\left[ \begin{matrix}
   {{t}_{x}}  \\
   0  \\
   {{t}_{z}}  \\
\end{matrix} \right],
	\label{h1}
\end{equation}
where, $\theta_y$ denotes the yaw rotation angle.

The rotation estimation can directly follow the formulation in Section \ref{sec:4pt_vertical}, leading to an equation similar to \eqref{d3_1}. The key to reducing the minimal point requirement from four to three lies in exploiting the translation constraint $t_y=0$. From \eqref{a5} and the relation $\mathbf{t} = -\mathbf{R} \mathbf{t}_1 + \mathbf{t}_2$, we have:
\begin{equation}
	\begin{aligned}
\mathbf{t} = -\mathbf{R} (\mathbf{x}_1 \times \mathbf{q}_1 + \lambda_1 \mathbf{x}_1) + (\mathbf{x}_2 \times \mathbf{q}_2 + \lambda_2 \mathbf{x}_2). 
\end{aligned}
	\label{h2}
\end{equation}
Enforcing $t_y=0$ and noting that for the yaw-only rotation in \eqref{d1}, $[\mathbf{R}\mathbf{v}]_y = [\mathbf{v}]_y$ for any vector $\mathbf{v}$, we simplify \eqref{h2} to:
\begin{equation}
	\begin{aligned}
0 = -[\mathbf{x}_1 \times \mathbf{q}_1 + \lambda_1 \mathbf{x}_1]_y + [\mathbf{x}_2 \times \mathbf{q}_2 + \lambda_2 \mathbf{x}_2]_y.
\end{aligned}
	\label{h4}
\end{equation}
Solving \eqref{h4} for $\lambda_2$ yields a linear relation:
\begin{equation}
	\begin{aligned}
\lambda_2 = k_1 \lambda_1 + k_2, 
\end{aligned}
	\label{h5}
\end{equation}
where the coefficients are given by:
\begin{equation}
	\begin{aligned}
k_1 = \frac{[\mathbf{x}_1]_y}{[\mathbf{x}_2]_y}, \quad k_2 = \frac{[\mathbf{x}_1 \times \mathbf{q}_1]_y - [\mathbf{x}_2 \times \mathbf{q}_2]_y}{[\mathbf{x}_2]_y}.
\end{aligned}
	\label{h6}
\end{equation}
Here, $[\cdot]_y$ denotes the Y-component of a vector.
Substituting \eqref{h5} into the constraint equation derived from \eqref{d3_1} eliminates $\lambda_2$. After simplification, we obtain a single equation per point correspondence:
\begin{equation}
	\begin{aligned}
A(\theta_y) \lambda_1 + B(\theta_y) = 0,
\end{aligned}
	\label{h9}
\end{equation}
where $A(\theta_y)$ and $B(\theta_y)$ are at most quadratic in $\theta_y$ due to the small-angle approximation.

With two non-origin point correspondences, we form a $2 \times 2$ homogeneous system in $[\lambda_1, 1]^\top$:
\begin{equation}
	\begin{aligned}
\underbrace{\begin{bmatrix}
A^{(1)}(\theta_y) & B^{(1)}(\theta_y) \\
A^{(2)}(\theta_y) & B^{(2)}(\theta_y)
\end{bmatrix}
}_{{{\mathbf{F}}_{2\times 2}}(\theta_y)}\begin{bmatrix}
\lambda_1 \\ 1
\end{bmatrix} = \mathbf{0}. 
\end{aligned}
	\label{h10}
\end{equation}
Analogous to \eqref{d5}, this system yields the determinant condition:
\begin{equation}
	\begin{aligned}
\text{det}({\mathbf{F}}(\theta_y))= 0
\end{aligned}.
	\label{h11}
\end{equation}
Expanding \eqref{h11} yields a quartic equation in $\theta_y$.
Solving this quartic equation yields up to four real solutions for $\theta_y$.

\subsection{Translation Estimation}
Once rotation parameters ($s$, $w$, or $\theta_y$) are estimated, the depth parameters $\lambda_1$ and $\lambda_2$ can be recovered. For the 4-point solvers with known vertical or rotation axis direction, each candidate solution is substituted into the corresponding $3 \times 3$ matrix $\mathbf{F}(\cdot)$ from \eqref{d4} or \eqref{g5}, yielding a homogeneous linear system in $\lambda_1$ and $\lambda_2$. The non-trivial solution is obtained via Singular Value Decomposition (SVD). For the planar motion case, $\lambda_1$ is computed from \eqref{h10} via SVD for each candidate $\theta_y$, after which $\lambda_2$ follows directly from the linear relation \eqref{h5}.

Subsequently, the derived $\lambda_1$ and $\lambda_2$ are substituted into the expressions for $\mathbf{t}_1$ and $\mathbf{t}_2$ given in \eqref{a5}, enabling the explicit calculation of these translation components. Finally, the complete translation vector $\mathbf{t}$ is obtained through the following relationship:
\begin{equation}
	\begin{aligned} 
  \mathbf{t}=-{{\mathbf{R}}}{{\mathbf{t}}_{1}}+{{\mathbf{t}}_{2}}
        \end{aligned},
	\label{g7}
\end{equation}
where, $\mathbf{R}$ is the previously determined rotation matrix. 

The three solvers are designed for distinct prior conditions typical in autonomous driving. \ref{sec:4pt_vertical} requires known vertical directions (e.g., from IMUs) and is applicable whenever gravity alignment is available, but its accuracy degrades under significant pitch/roll noise. \ref{sec:4pt_axis} uses known rotation axis directions (e.g., from steering or IMUs) and excels during turns, though it is less reliable in straight‑line motion. \ref{sec:3-Point} assumes planar motion and performs well on flat terrain but is sensitive to vertical vibrations or non‑planar motion. In practice, the appropriate solver can be selected based on available sensor data and motion context, with the most constrained variant preferred for higher efficiency.

\section{\label{sec:Degeneracy}Degeneracy Analysis}
\subsection{Point Correspondence Classification}
\begin{figure}[H]
	\centering	
		\subfigure[\centering Intra-camera]
		{
			\label{fig:Intra-camera}
        \includegraphics[width=0.45\linewidth]{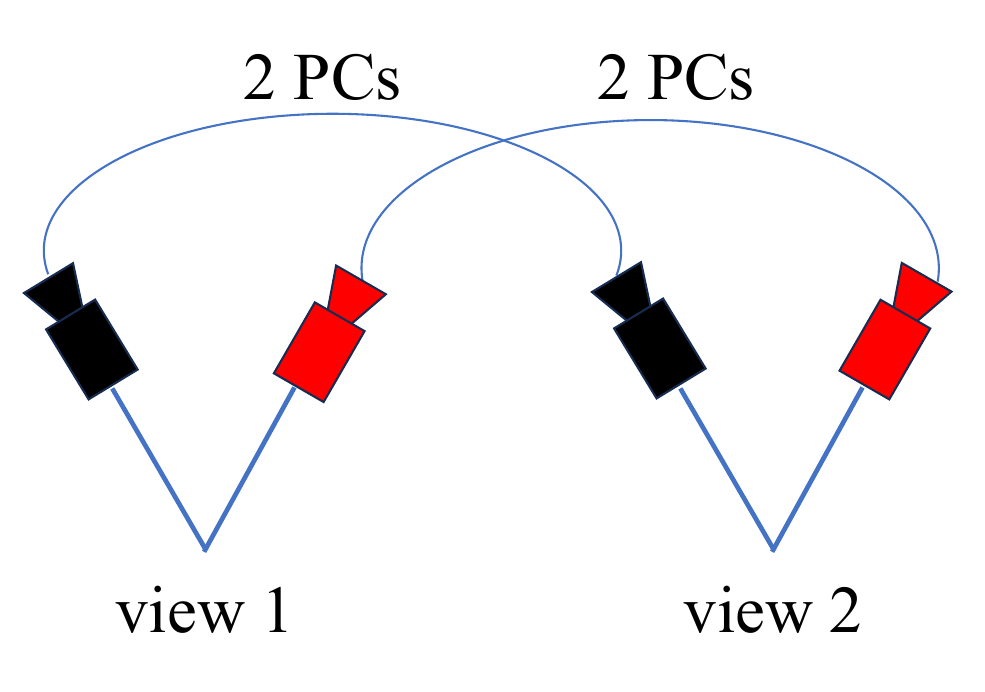}
		}
		\subfigure[\centering Inter-camera]
		{
			\includegraphics[width=0.45\linewidth]{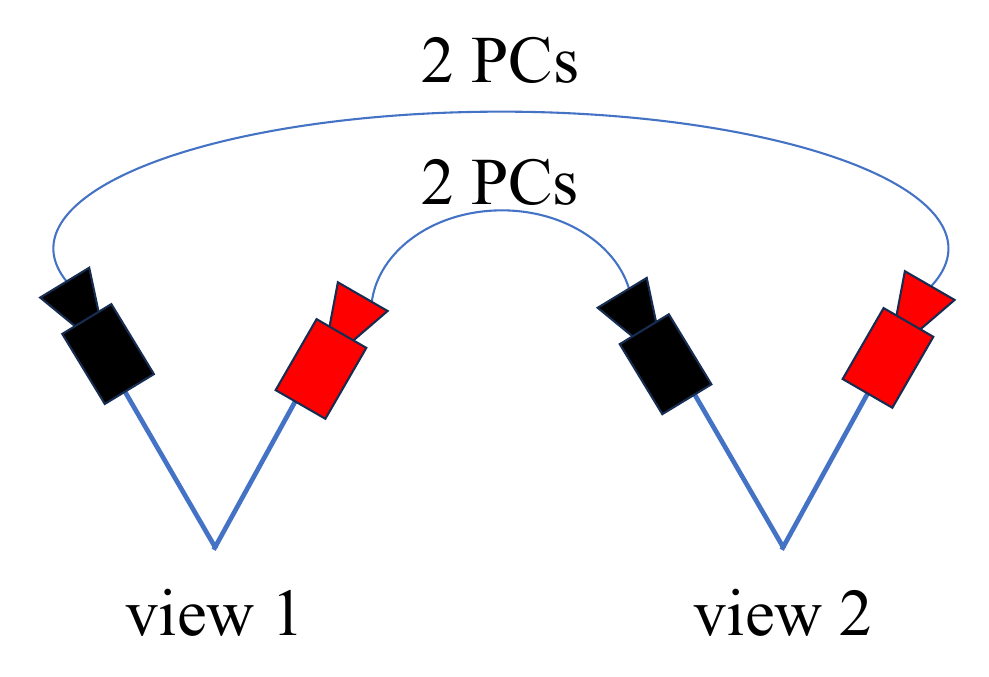}
		}
	\caption{Relative pose estimation from four point correspondences for multi-camera systems with two cameras. (a)Intra-camera point correspondences refer to point features that are seen by the same camera across two consecutive
views. (b)Inter-camera point correspondences refer to point features that are seen by different cameras across two consecutive
views.}
	\label{fig:Intra-Inter}
\end{figure}

Based on the attributes of point features, point correspondences can be categorized into intra-camera point correspondences and inter-camera point correspondences, see Fig.~\ref{fig:Intra-Inter}.
Intra-camera point correspondences refer to point features that are observed by the same camera across two consecutive views. 
This correspondence type is particularly well-suited for multi-camera systems with non-overlapping or minimally overlapping fields of view, where feature tracking within a single camera's perspective provides enhanced stability and reliability.
Inter-camera point correspondences, on the other hand, describe point features that are detected by different cameras across two consecutive views. This approach is optimal for multi-camera systems with significantly overlapping fields of view, enabling robust feature matching across different camera perspectives.
The selection between these two types of correspondences depends on the camera configuration and the degree of overlap in their respective viewing ranges.

\subsection{Degeneracy in Generalized Epipolar Constraint}
Consider the case of pure translation where the relative rotation $\mathbf{R} = \mathbf{I}$, combined with intra-camera correspondences where $\mathbf{s}_i = \mathbf{s}_j$. Under these conditions, the cross-term in the generalized epipolar constraint \eqref{b3} vanishes:
\begin{equation}
	\begin{aligned}
{{\mathbf{x}}_{i}}^{\text{T}}{{\mathbf{R}}}^{\text{T}}{{\mathbf{q}}_{j}}+{{\mathbf{x}}_{j}}^{\text{T}}{{\mathbf{R}}}{{\mathbf{q}}_{i}}={{\mathbf{x}}_{i}}^{\text{T}}{{\mathbf{s}}_{i}}\times {{\mathbf{x}}_{j}}+{{\mathbf{x}}_{j}}^{\text{T}}{{\mathbf{s}}_{i}}\times {{\mathbf{x}}_{i}}=\mathbf{0}
    \end{aligned}.
	\label{k1}
\end{equation}
Consequently, the generalized epipolar constraint \eqref{b3} reduces to the following form:
\begin{equation}
	\begin{aligned}
{{\mathbf{x}}_{j}}^{\mathrm{T}}({{[{{\mathbf{t}}_{2}-\mathbf{t}}_{1}}]}_{\times }){{\mathbf{x}}_{i}}=0
    \end{aligned}.
	\label{k2}
\end{equation}
In this degenerate case, the constraint matrix becomes homogeneous with respect to the relative translation $\mathbf{t}_2 - \mathbf{t}_1$ between the two views. Introducing a free parameter $\kappa$, it can be verified that $\kappa(\mathbf{t}_2 - \mathbf{t}_1)$ always satisfies equation \eqref{k2}, indicating the loss of scale information.
\subsection{Degeneracy in Epipolar Constraint}
Similarly, for the epipolar constraint, substituting $\mathbf{R}=\mathbf{I}$ and $\mathbf{s}_i = \mathbf{s}_j$ into the essential matrix formulation \eqref{eq:essential_matrix}, we obtain:
\begin{equation}
\begin{aligned} 
	{\mathbf{E}}_{\mathbf{c}j\mathbf{c}i} &= {{\mathbf{Q}}_{j}}^{\mathrm{T}} \left( [{{\mathbf{t}}_{2}} - \mathbf{t}_1]_\times  \right) {{\mathbf{Q}}_{i}}. 
	\label{eq:essential_matrix_deg}
\end{aligned}
\end{equation}
Applying the epipolar constraint $\mathbf{u}_{j}^{\mathrm{T}}{{\mathbf{E}}_{\mathbf{c}j\mathbf{c}i}}{{\mathbf{u}}_{i}}=0$ yields an identical constraint to equation \eqref{k2}. 
The resulting constraint matrix remains homogeneous with respect to $\mathbf{t}_2 - \mathbf{t}_1$, confirming that $\kappa(\mathbf{t}_2 - \mathbf{t}_1)$ satisfies the constraint for any scalar $\kappa$. 

\section{Experiment}\label{Experiment}
To comprehensively validate our proposed methods, we conducted extensive performance evaluations on both synthetic and real-world datasets.
The \texttt{4pt-Approx} corresponds to the method presented in Section~\ref{sec:4pt_vertical}, \texttt{4pt-Axis-Approx} corresponds to the method described in Section~\ref{sec:4pt_axis}, while \texttt{3pt-Approx} corresponds to the method described in Section~\ref{sec:3-Point}.
For comparative analysis, we benchmark our solvers against both state-of-the-art first-order rotation approximation methods, including \texttt{4pt-Liu}~\cite{liu2017robust} and \texttt{6pt-Ventura}~\cite{ventura2015efficient}, and classic non-first-order rotation approximation methods, including \texttt{4pt-Lee}~\cite{hee2014relative}, \texttt{17pt-Li}~\cite{li2008linear}, and \texttt{6pt-Stew}~\cite{henrikstewenius2005solutions}. 
Rotation and translation errors are measured using the following metrics \cite{guan2022affine}:
 \begin{equation}
	\begin{aligned} 
&\varepsilon_{\mathbf{R}}=\arccos((\mathrm{trace}(\mathbf{R}_{gt}\mathbf{R}^T)-1)/2) \\  
&\varepsilon_{\mathbf{t}, \mathrm{dir}}=\arccos((\mathbf{t}_{gt}^T\mathbf{t})/(\|\mathbf{t}_{gt}\|\cdot\|\mathbf{t}\|))
\end{aligned}.
\label{erro_metrics}
\end{equation}
Here, $\mathbf{R}_{gt}$ and $\mathbf{t}_{gt}$ denote the ground truth values, while $\mathbf{R}$ and $\mathbf{t}$ represent the estimated counterparts. $\varepsilon_{\mathbf{R}}$ represents the angular error of rotation, while $\varepsilon_{\mathbf{t}, \mathrm{dir}}$ measures the directional error of translation, neglecting its scale.
The main reason that we focus solely on the translation directional error is that our experiments use only intra-camera correspondences and assume small-angle rotations akin to pure translation. As Section \ref{sec:Degeneracy} analyzes, these conditions inherently prevent reliable translation scale recovery. Therefore, translation accuracy is assessed using only directional information.

\subsection{Efficiency Comparison}
To ensure a fair comparison, all solvers were evaluated on an AMD R9-7945HX 2.50 GHz processor using C++ implementations. Table~\ref{tab:SolverTime_4pt} reports execution times across 10,000 trials, with mean values outside parentheses and median values inside parentheses. The results indicate that the 4-point solvers achieve substantially higher computational efficiency than 6-DoF solvers such as \texttt{17pt-Li}~\cite{li2008linear}, \texttt{6pt-Stew}~\cite{henrikstewenius2005solutions} and \texttt{6pt-Ventura}~\cite{ventura2015efficient}.
Among the 4-point solvers, \texttt{4pt-Liu}~\cite{liu2017robust} achieves the fastest execution time, followed by \texttt{4pt-Axis-Approx}, and \texttt{4pt-Approx}, which are approximately 1.7× faster than the non-approximation solver \texttt{4pt-Lee}~\cite{hee2014relative}. The 3-point solver \texttt{3pt-Approx} is slightly slower than the 4-point methods due to the need to solve a quartic polynomial, yet it still maintains high computational efficiency. Collectively, these experimental results validate the computational advantages of our proposed methods.

\begin{table}[H]
	\caption{Runtime comparison of relative pose estimation solvers for multi-camera systems (unit:~$\mu s$).}
	\centering
		\setlength{\tabcolsep}{8pt}{
			\scalebox{1}{
				\begin{tabular}{ccccc}
					\hline					
					Methods & 17pt-Li \cite{li2008linear}  & 6pt-Stew \cite{henrikstewenius2005solutions}  &6pt-Ventura \cite{ventura2015efficient} & 4pt-Lee \cite{hee2014relative}     \\
					\hline
					Runtime &50.2704(48) & 3759.8892(3744) & 57.4035(53)  & 28.0242(27)    \\ \hline
                    Methods 
                            &4pt-Liu \cite{liu2017robust} & 4pt-Approx  & 4pt-Axis-Approx &3pt-Approx \\ \hline
					Runtime  & \textbf{8.2684(7)} & 16.2423(10) & 13.6533(8) & 18.6852(12) \\
					\hline
		\end{tabular}}}
	\label{tab:SolverTime_4pt}
\end{table}

\subsection{Experiments on Synthetic Data}
We evaluated the proposed solvers on synthetic data using a virtual multi-camera system consisting of two cameras. 
The position and orientation of the multi-camera system varied over time, with a focal length of 400 pixels. 
All points were ensured to be visible in both views, though only intra‑camera correspondences were employed for pose estimation, reflecting the typical configuration of multi‑camera systems with non‑overlapping fields of view. All solvers were integrated into the RANSAC framework, and the solution yielding the largest number of inliers was selected for final error computation. Two distinct motion scenarios were simulated: general 6-DoF non-planar motion for evaluating the proposed \texttt{4pt-Approx} and \texttt{4pt-Axis-Approx} solvers, and strictly constrained planar motion for evaluating \texttt{3pt-Approx}.

\subsubsection{\label{sec:noplane}Non-Planar Motion Estimation}
\begin{figure}[H]
	\centering
		\includegraphics[width=0.8\linewidth]{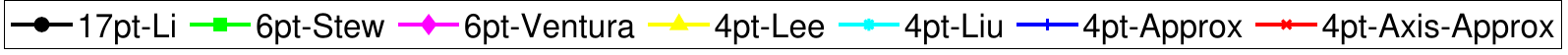}\\
		\subfigure[\centering $\varepsilon_{\mathbf{R}}$]
		{
		\includegraphics[width=0.31\linewidth]{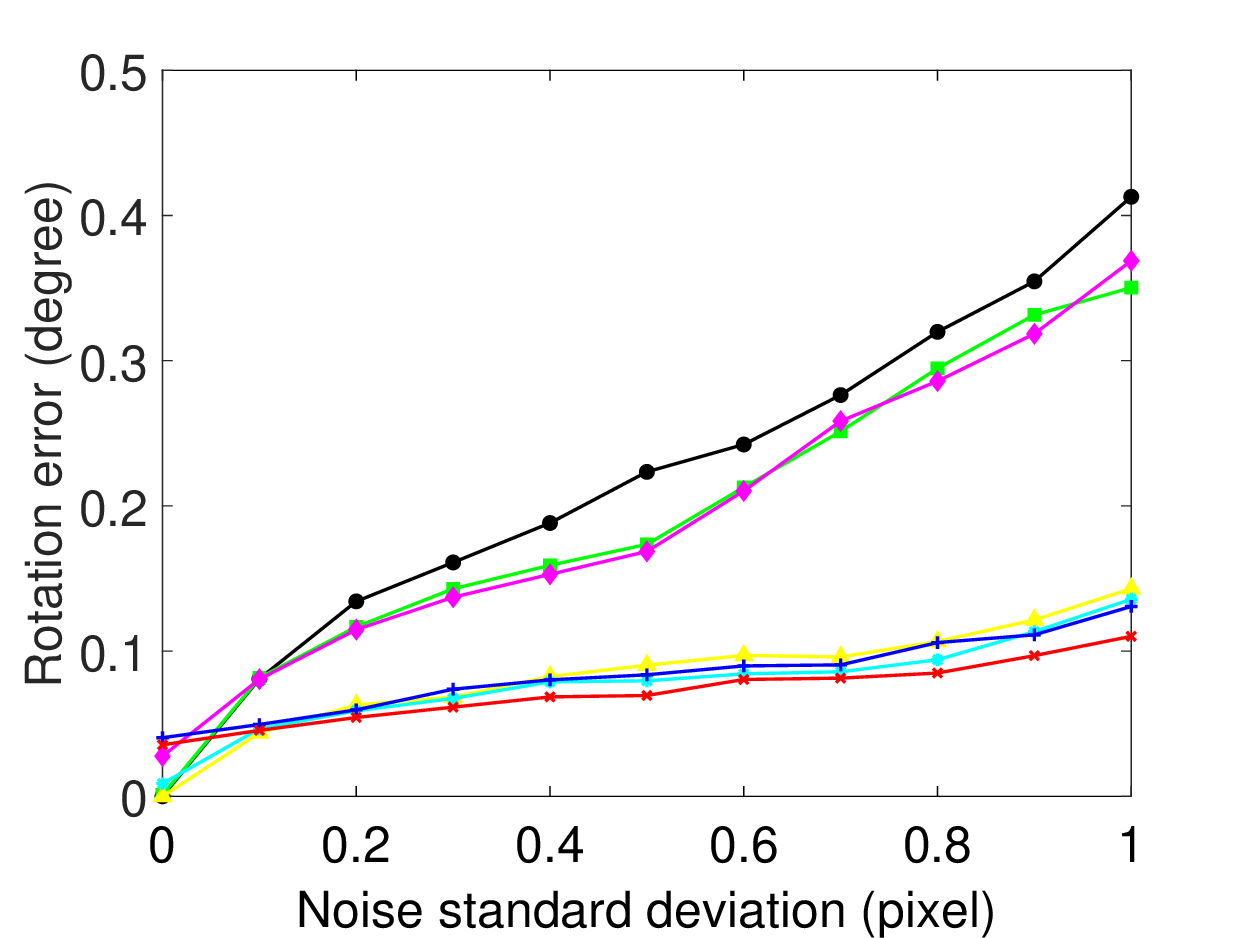}
		}
    \subfigure[\centering ${\varepsilon_{\bf{R}}}$]
		{
		\includegraphics[width=0.31\linewidth]{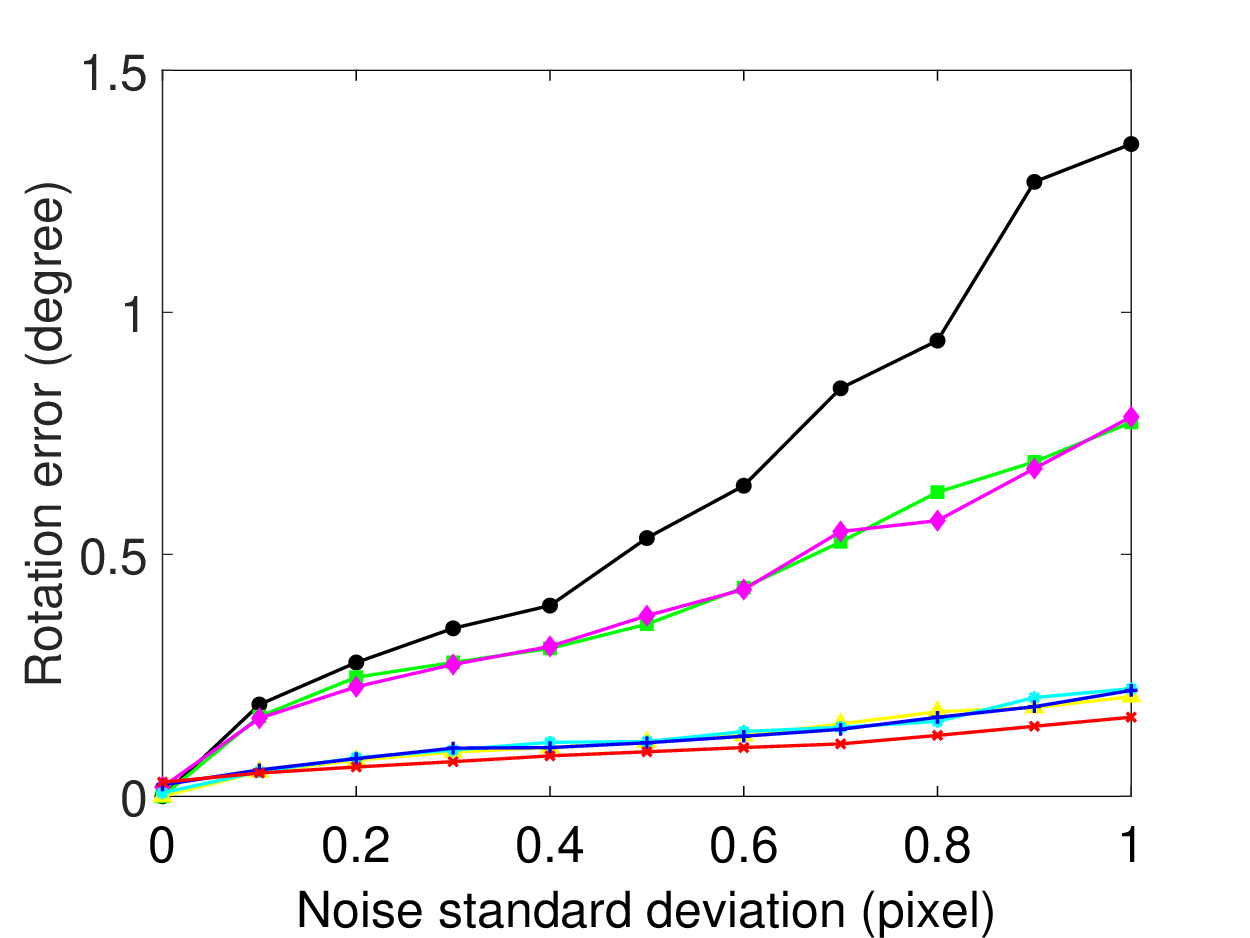}
		}
\subfigure[\centering ${\varepsilon_{\bf{R}}}$]
		{
		\includegraphics[width=0.31\linewidth]{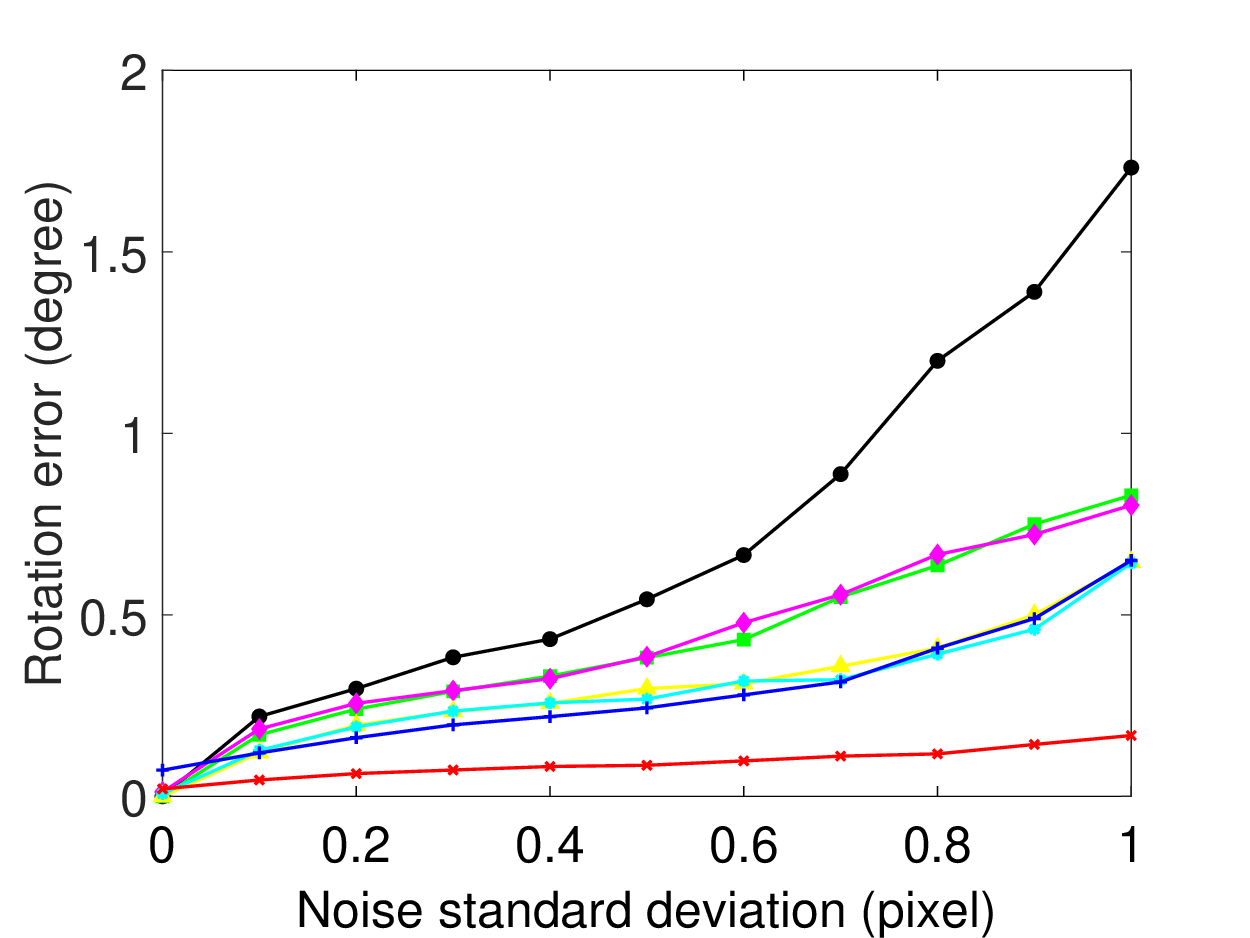}
		}
\\ 
		\subfigure[\centering $\varepsilon_{\mathbf{t},\text{dir}}$]
		{
		\includegraphics[width=0.31\linewidth]{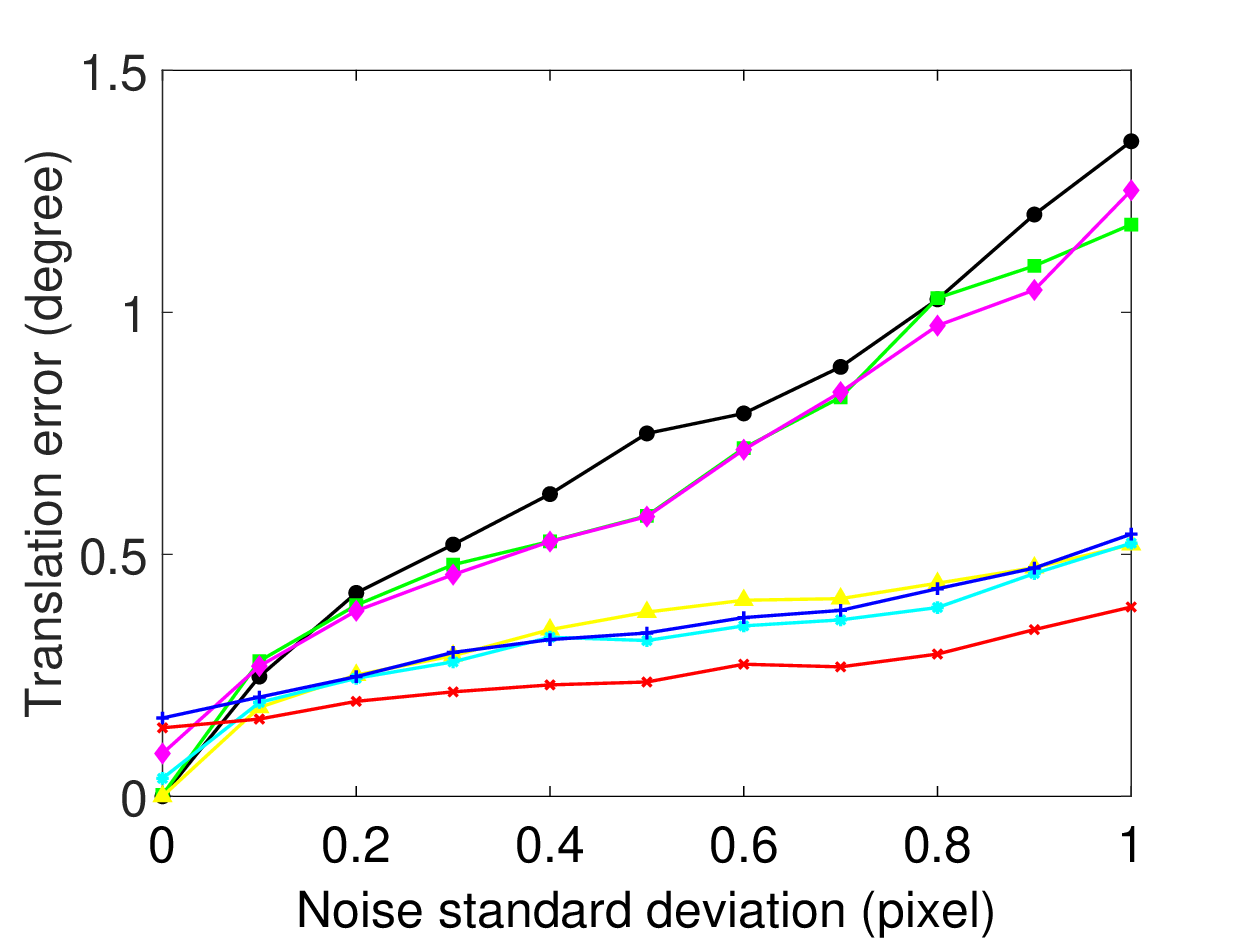}
		}
		\subfigure[\centering $\varepsilon_{\mathbf{t},\text{dir}}$]
		{
		\includegraphics[width=0.31\linewidth]{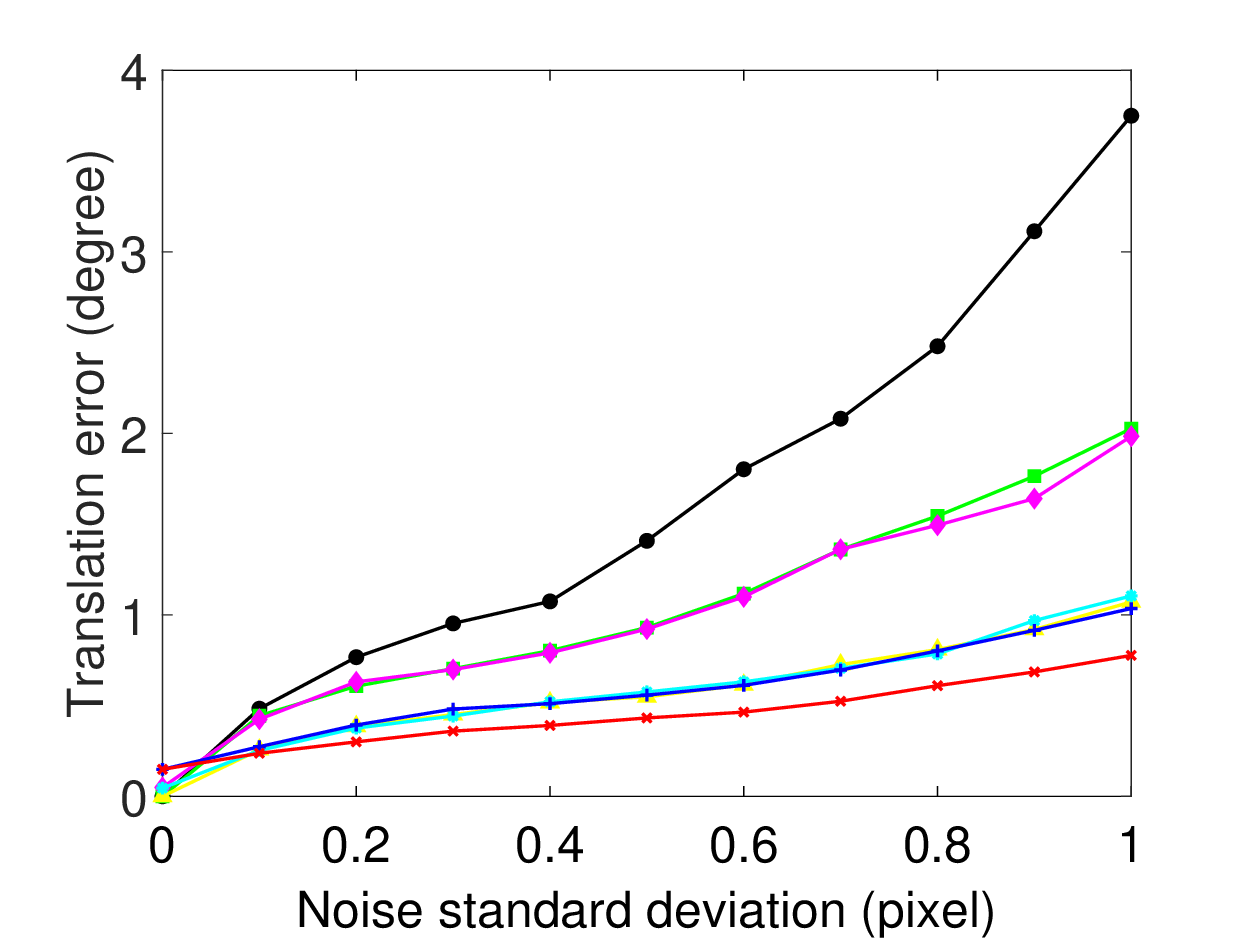}
		}	 		
	\subfigure[\centering $\varepsilon_{\mathbf{t},\text{dir}}$]
		{
		\includegraphics[width=0.31\linewidth]{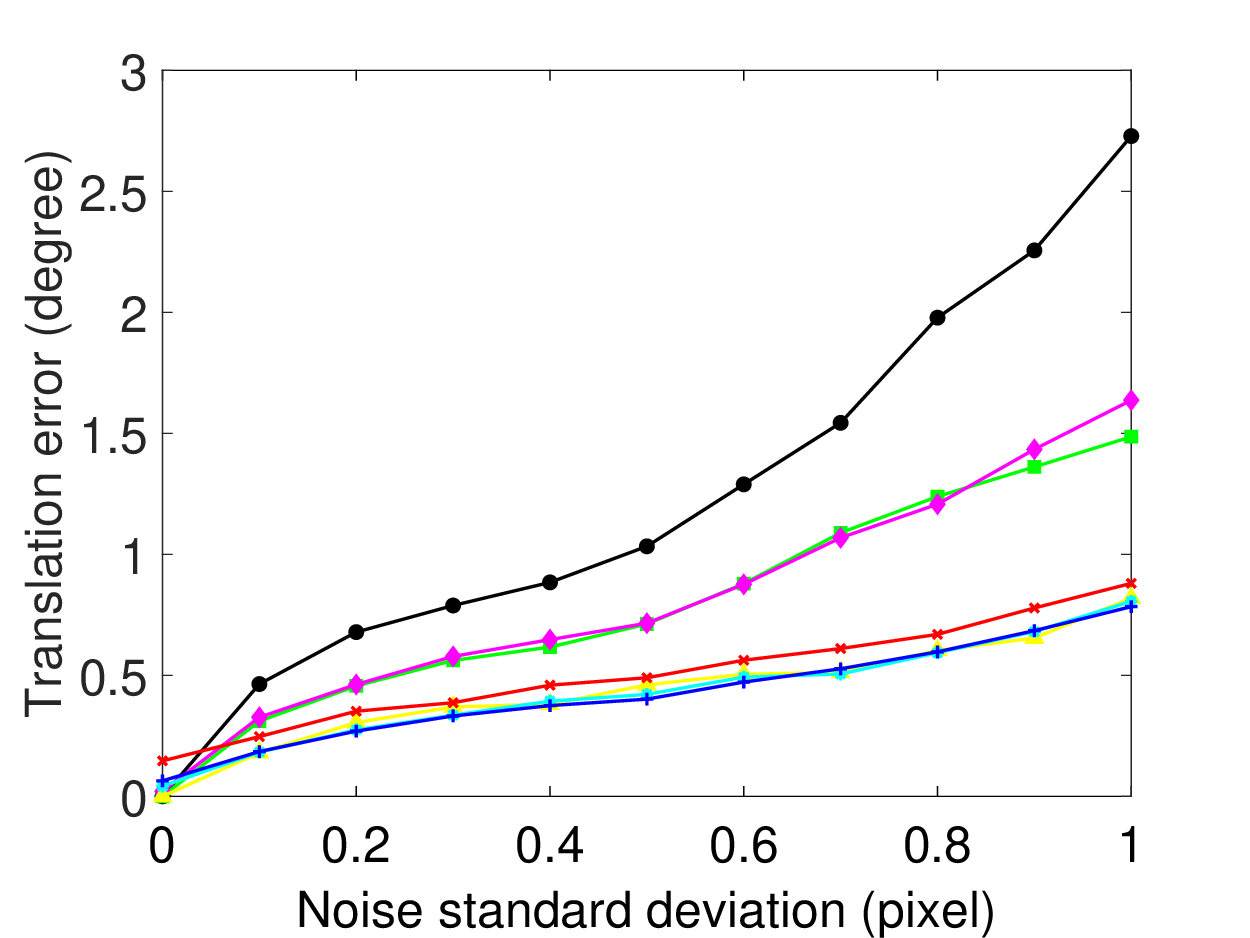}
		}
	
	\caption{Rotation and translation errors for multi-camera systems with increasing image noise under non-planar motions. The first, second, and third columns present the performance of different solvers under forward, random, and sideways motions, respectively.}
	\label{fig:image_noise}
\end{figure}

In this section, we evaluate pose estimation under general 6-DoF motion, while assuming the vertical direction or the rotation axis direction is known—a common prior in many practical systems equipped with IMUs or gravity sensors. 
\textbf{1) Image Noise Evaluation.} 
In this experiment, Gaussian noise with a standard deviation ranging from 0 to 1.0 pixel was added to image coordinates. The ground-truth rotation angle was set to $1^\circ$, reflecting small-angle rotation scenarios suitable for first-order rotation approximation. 
 Three motion patterns were tested, including forward motion, random motion, and sideways motion. For each pattern, we evaluate both rotation and translation estimation errors. As shown in Fig.~\ref{fig:image_noise}, the top row presents rotation errors $\varepsilon_{\mathbf{R}}$, and the bottom row presents translation direction errors $\varepsilon_{\mathbf{t},\text{dir}}$. Columns correspond to forward, random, and sideways motions, respectively.

Experimental results in Fig.~\ref{fig:image_noise} demonstrate that 4-point solvers exhibit substantially greater robustness against image noise compared to the 6-DoF solvers such as \texttt{17pt-Li}~\cite{li2008linear}, \texttt{6pt-Stew}~\cite{henrikstewenius2005solutions}, and \texttt{6pt-Ventura}~\cite{ventura2015efficient}.
Among all the 4-point solvers, our proposed \texttt{4pt-Axis-Approx} consistently achieves higher accuracy across various motion conditions, with the exception of sideways translation, highlighting its superior noise resilience.
For 4-point solvers utilizing IMU vertical direction priors, our proposed \texttt{4pt-Approx} exhibited accuracy comparable to \texttt{4pt-Lee}\cite{hee2014relative} and \texttt{4pt-Liu}\cite{liu2017robust}, with similarly strong noise resistance. Overall, the proposed solvers demonstrate competitive robustness to image noise, with \texttt{4pt-Axis-Approx} exhibiting particularly well.

\textbf{2) IMU Noise Evaluation.}
In practical applications, high-end IMUs can achieve angular measurement accuracy as precise as $0.02^\circ$, while even low-cost IMUs can maintain errors below $0.5^\circ$~\cite{kukelova2010closed}. The IMUs found in modern vehicles or smartphones typically have an accuracy around $0.06^\circ$~\cite{Ding2021Globally}. 
 To thoroughly evaluate the algorithm's robustness against IMU inaccuracies, extensive simulations were performed with angular noise ranging from $0^\circ$ to $1^\circ$. This range spans the performance of both commercial and industrial-grade IMUs. 
The ground-truth rotation angle was kept at $1^\circ$, and image noise was fixed at 0.5 pixels.

\begin{figure}[H]
	\centering
				\includegraphics[width=0.8\linewidth]{fig/simulation/legend.pdf}\\
		\subfigure[\centering ${\varepsilon_{\bf{R}}}$]
		{
			\includegraphics[width=0.31\linewidth]{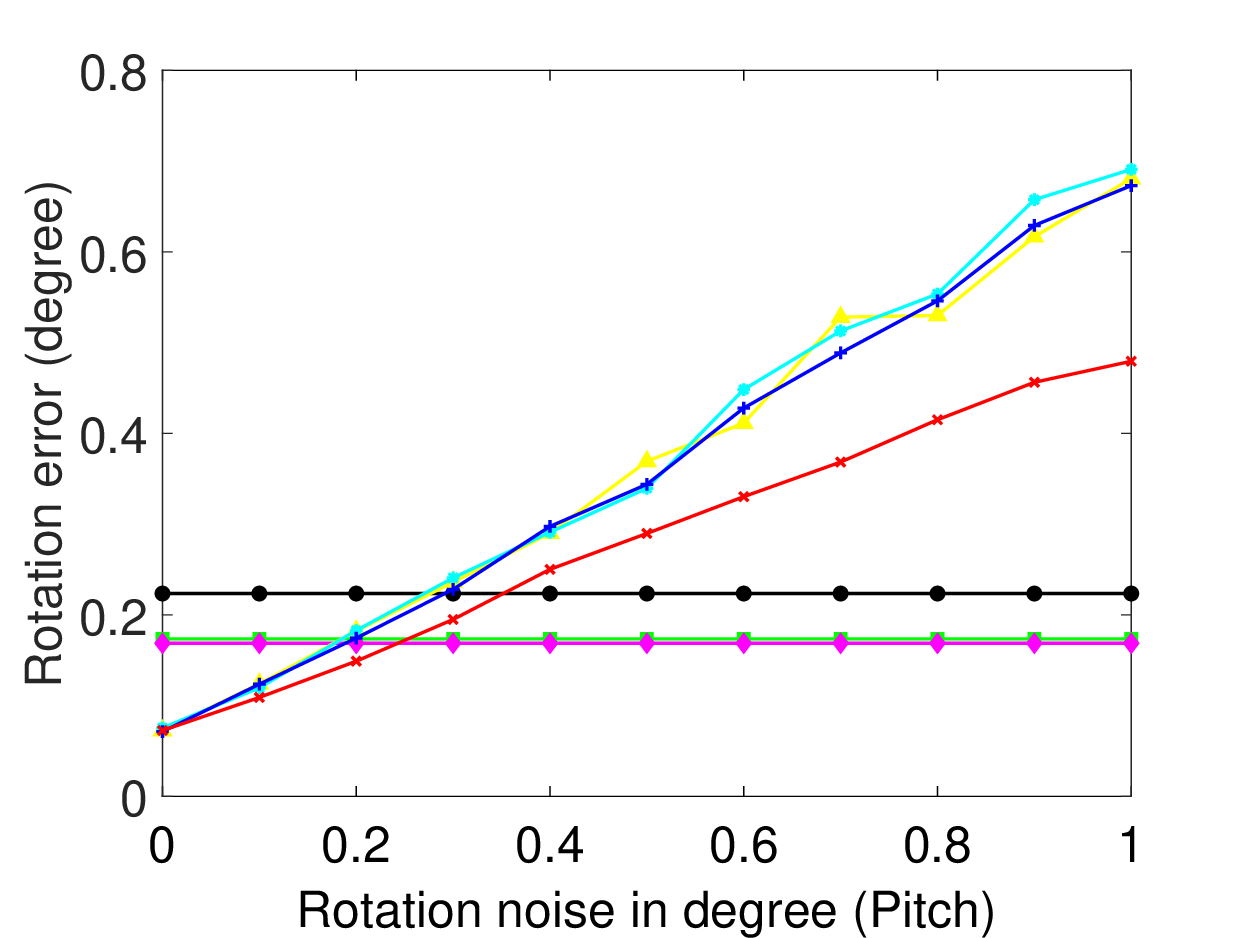}
		}
\subfigure[\centering ${\varepsilon_{\bf{R}}}$]
		{
			\includegraphics[width=0.31\linewidth]{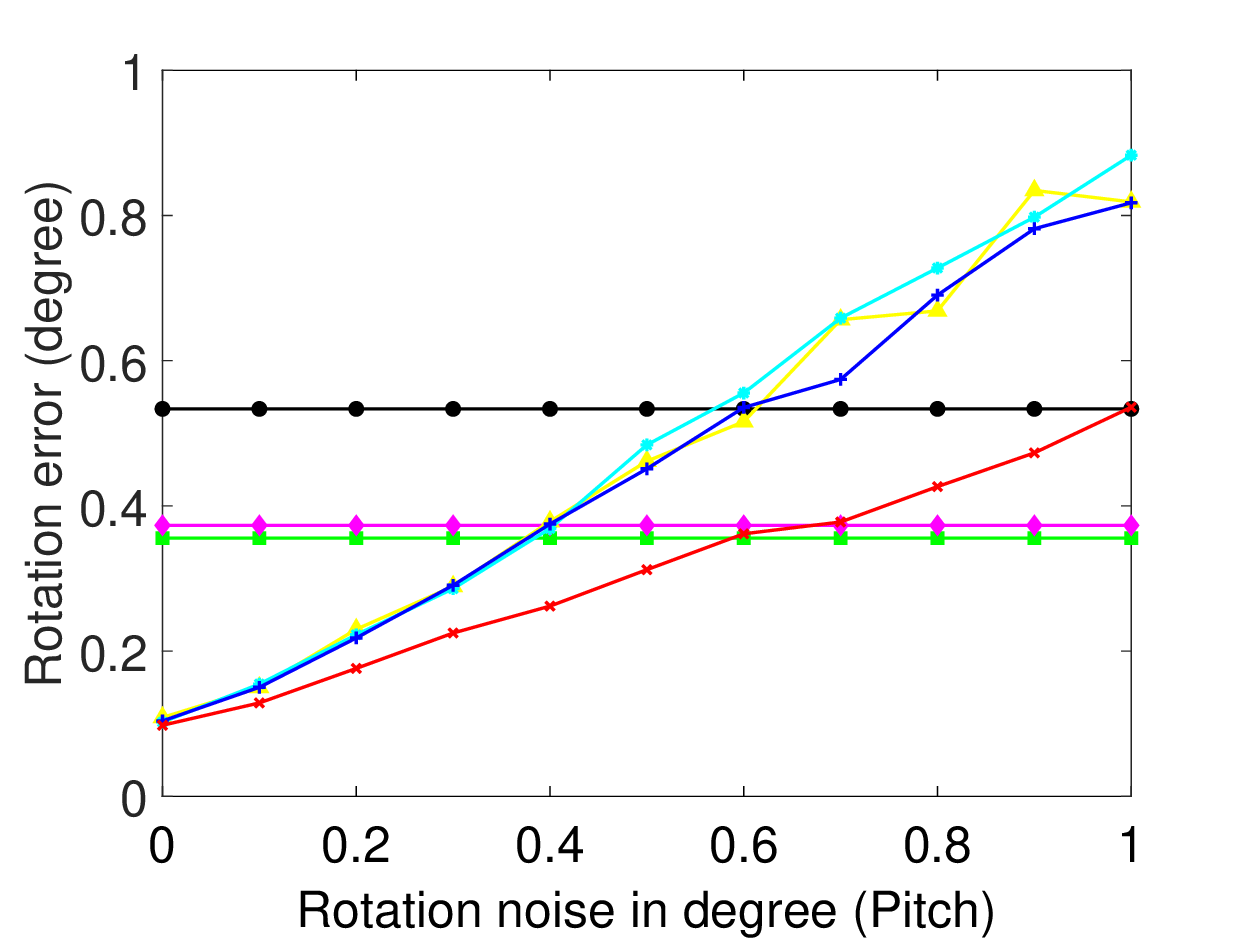}
		}
        \subfigure[\centering ${\varepsilon_{\bf{R}}}$]
		{
			\includegraphics[width=0.31\linewidth]{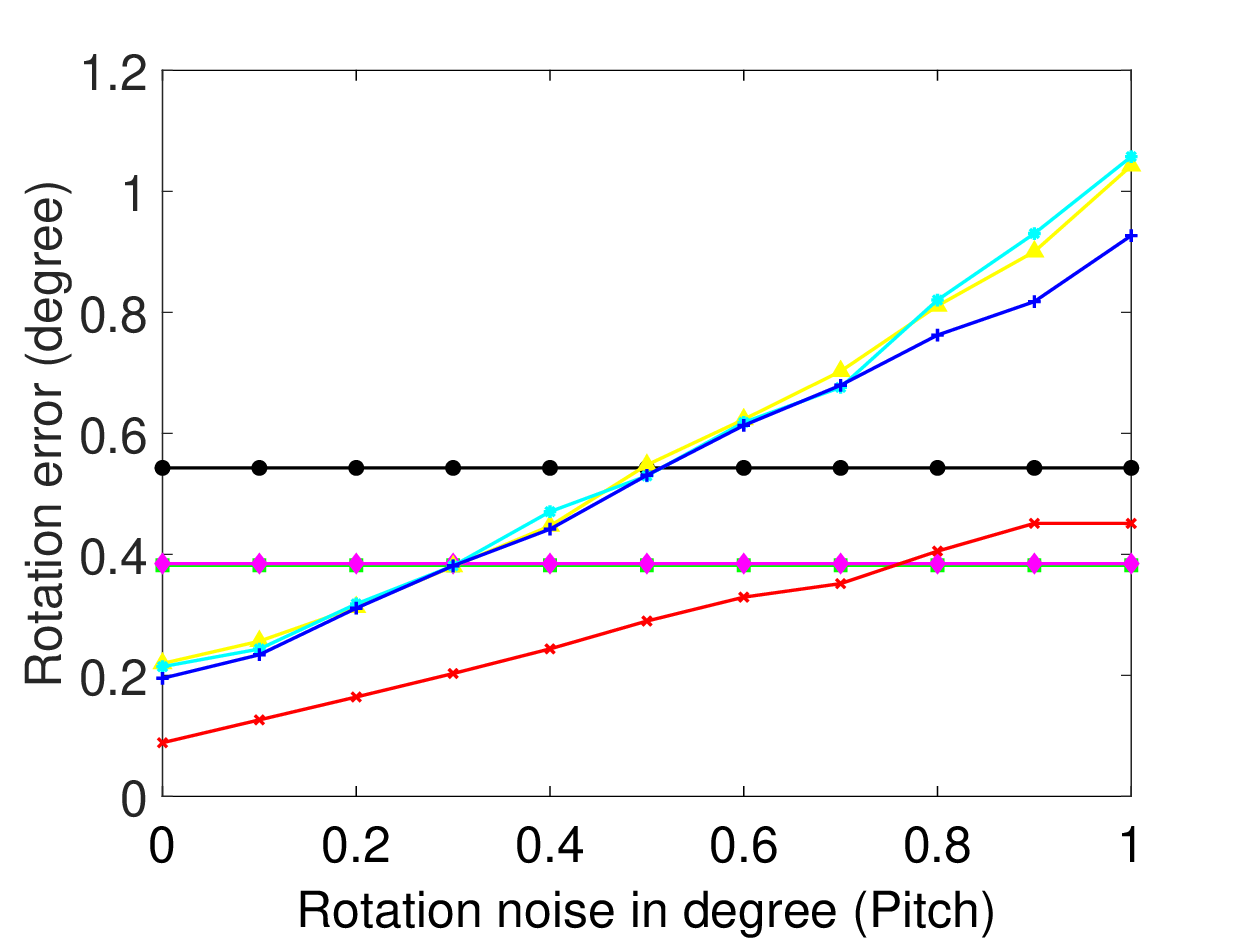}
		}
        \\
\subfigure[\centering $\varepsilon_{\mathbf{t},\text{dir}}$]
		{
			\includegraphics[width=0.31\linewidth]{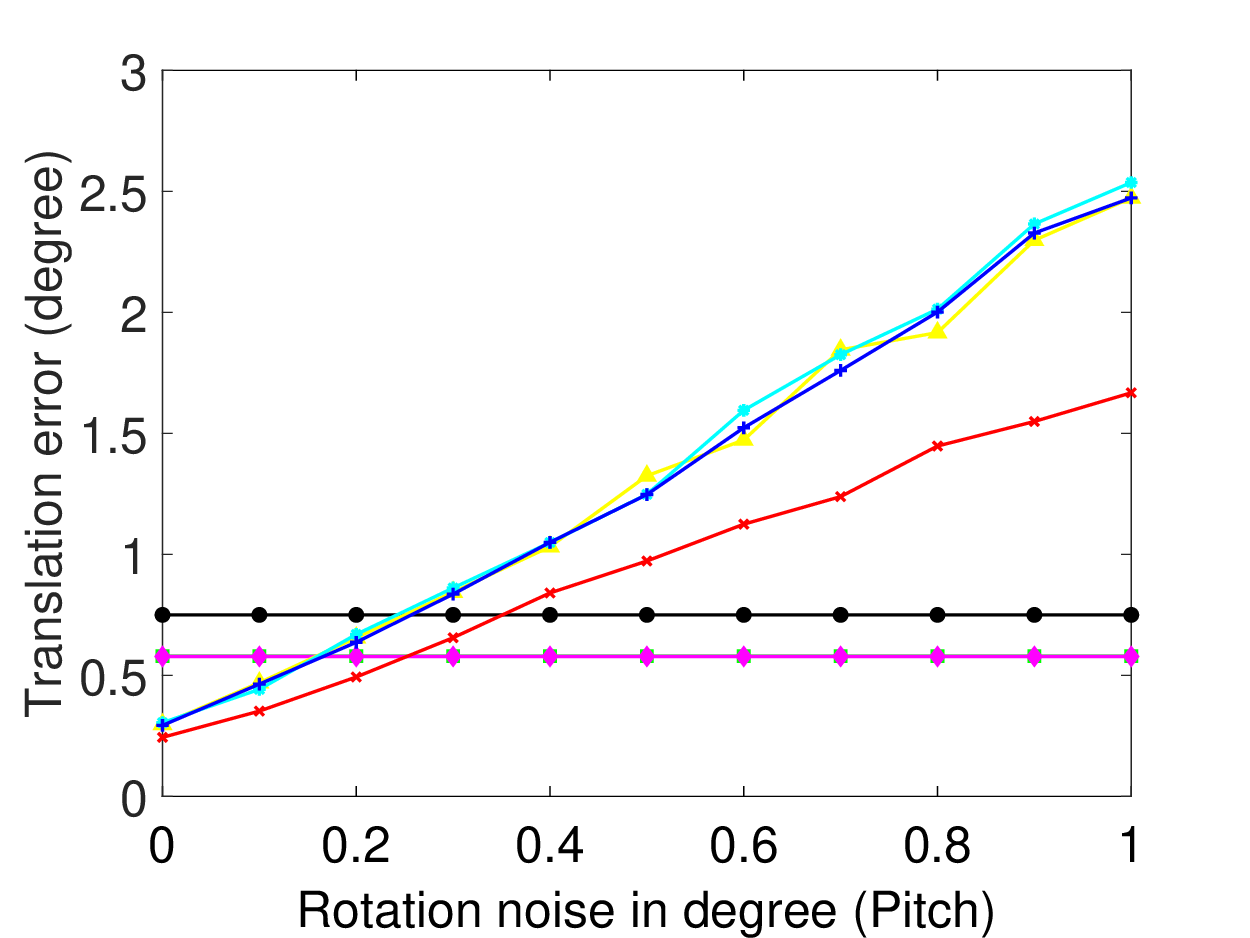}
		}
\subfigure[\centering $\varepsilon_{\mathbf{t},\text{dir}}$]
		{
			\includegraphics[width=0.31\linewidth]{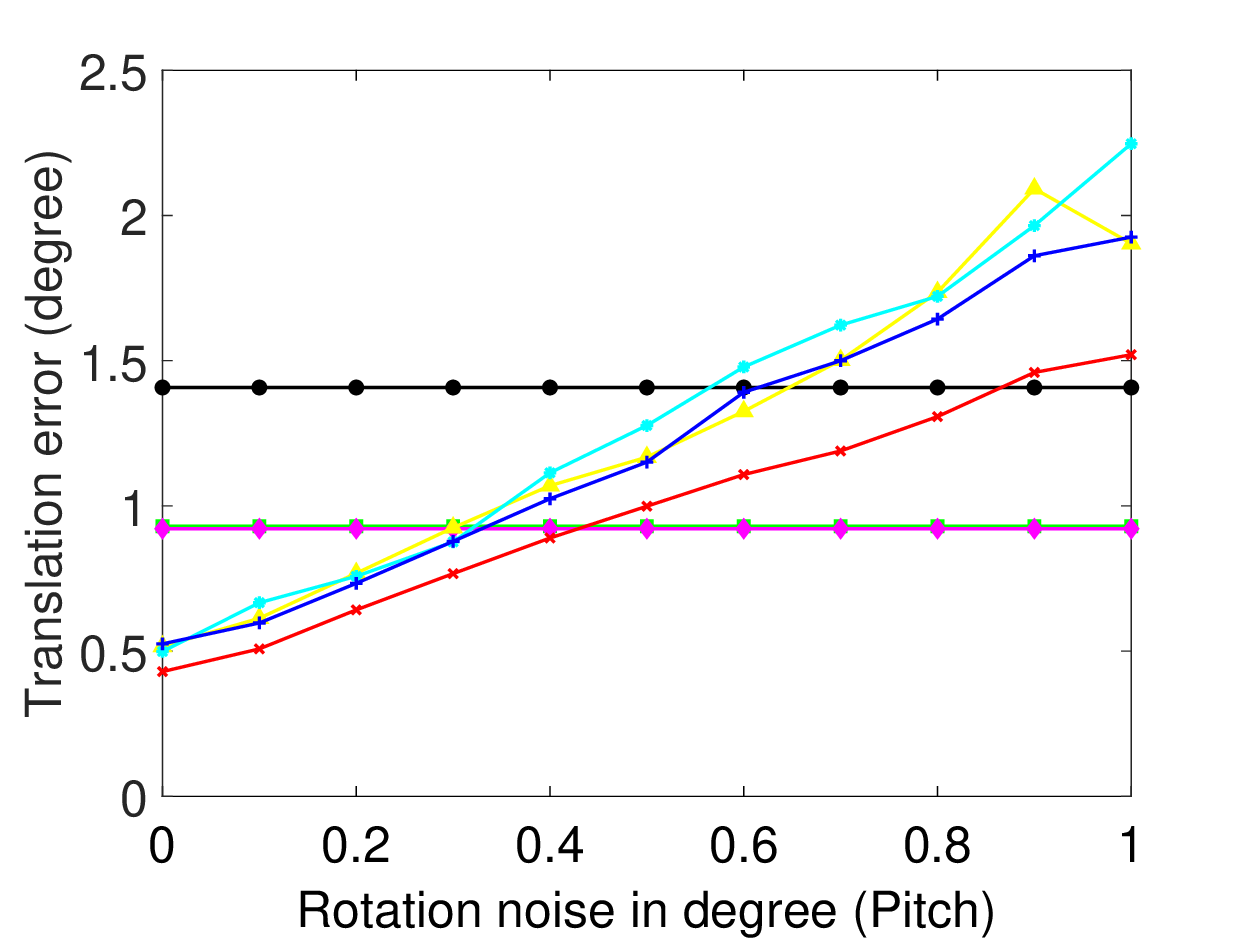}
		}
\subfigure[\centering $\varepsilon_{\mathbf{t},\text{dir}}$]
		{
			\includegraphics[width=0.31\linewidth]{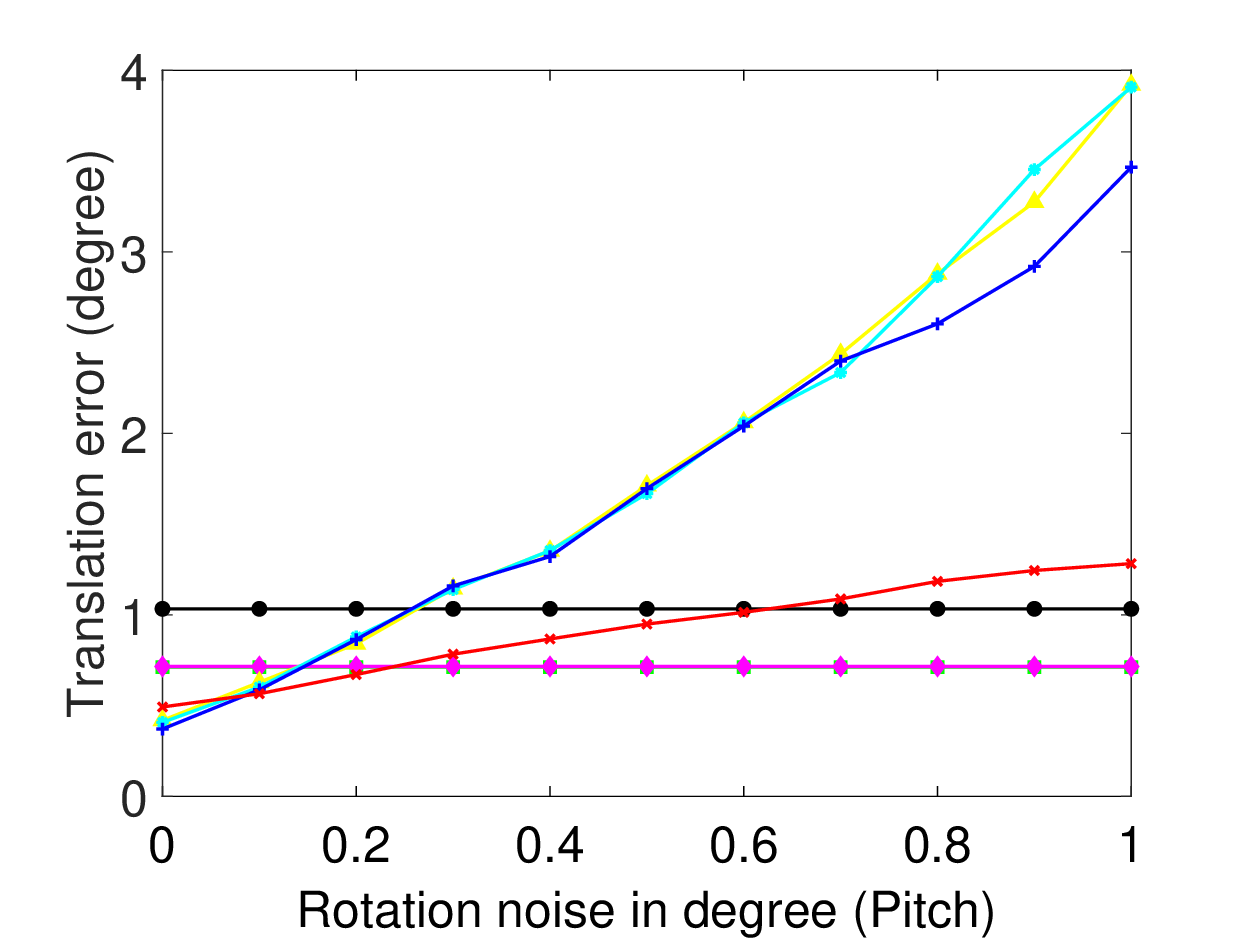}
		}
	\caption{Rotation and translation errors for multi-camera systems with increasing IMU pitch angle noise. The first, second, and third columns present the performance of different solvers under forward, random, and sideways motions, respectively.}
	\label{fig:pitch_angle_noise}
\end{figure}

Fig.~\ref{fig:pitch_angle_noise} shows the performance of different solvers under increasing IMU pitch angle noise for forward, random, and sideways motion. 
As expected, 6-DoF solvers remain unaffected by IMU pitch angle noise, since they do not incorporate IMU angle measurements. In contrast, the accuracy of 4-point solvers gradually declines as IMU noise increases. 
 Notably, under low noise conditions (below $0.2^\circ$), 4-point solvers still achieve significantly higher accuracy than 6-DoF solvers.
Among the 4-point solvers that utilize IMU priors, our \texttt{4pt-Axis-Approx} clearly outperforms the others, demonstrating superior resistance to pitch angle noise. This enhanced performance can be attributed to its rotation representation: pitch angle noise directly perturbs the estimated vertical direction in other methods, while such angular errors are generally non-orthogonal to the rotation axis direction used in \texttt{4pt-Axis-Approx}, thereby reducing error propagation and yielding higher accuracy under identical noise levels.
Furthermore, our \texttt{4pt-Approx} performs comparably to the state-of-the-art solvers \texttt{4pt-Liu}~\cite{liu2017robust} and \texttt{4pt-Lee}~\cite{hee2014relative}.

\begin{figure}[H]
	\centering
		\includegraphics[width=0.8\linewidth]{fig/simulation/legend.pdf}\\
		\subfigure[\centering ${\varepsilon_{\bf{R}}}$]
		{
			\includegraphics[width=0.31\linewidth]{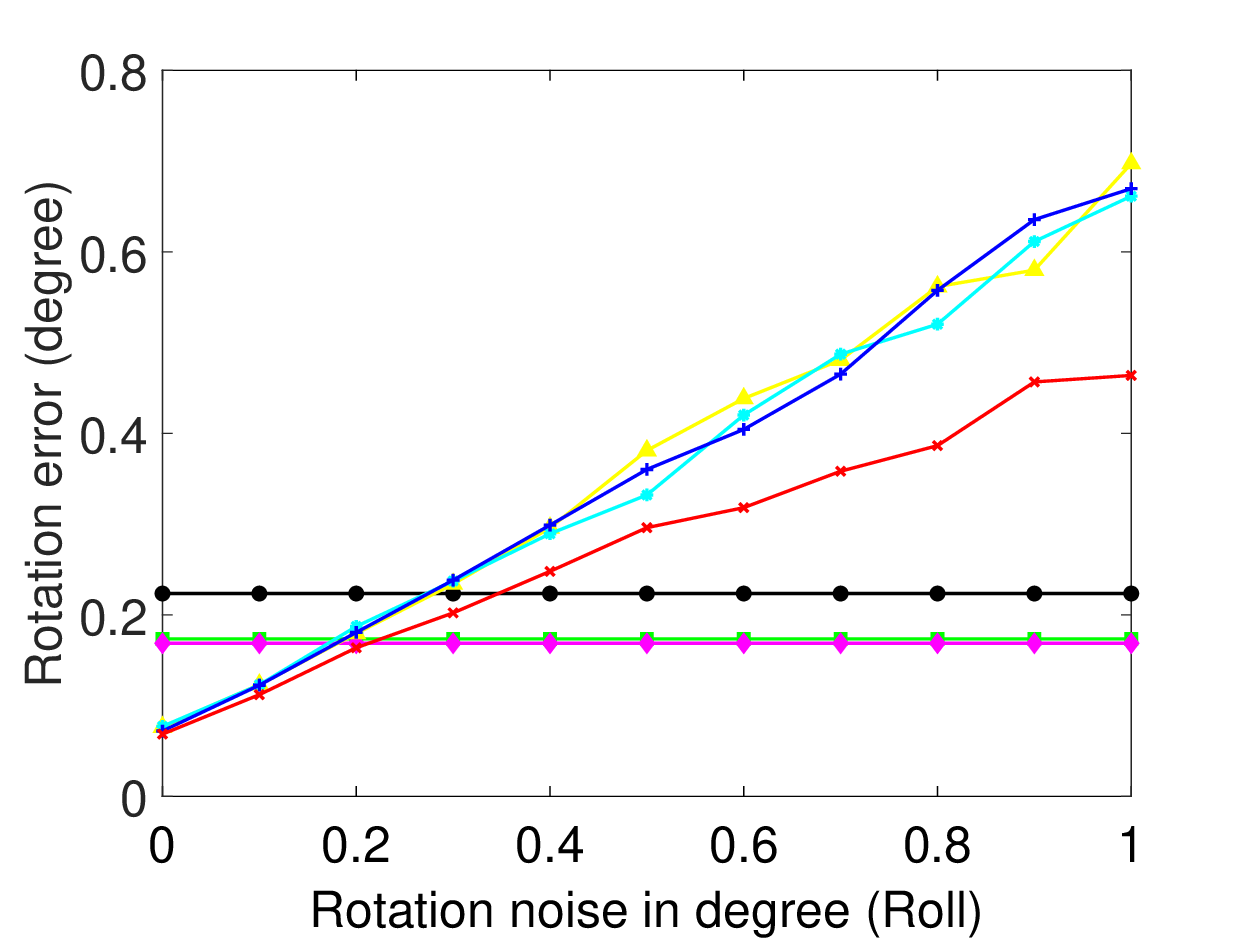}
		}
		\subfigure[\centering ${\varepsilon_{\bf{R}}}$]
		{
			\includegraphics[width=0.31\linewidth]{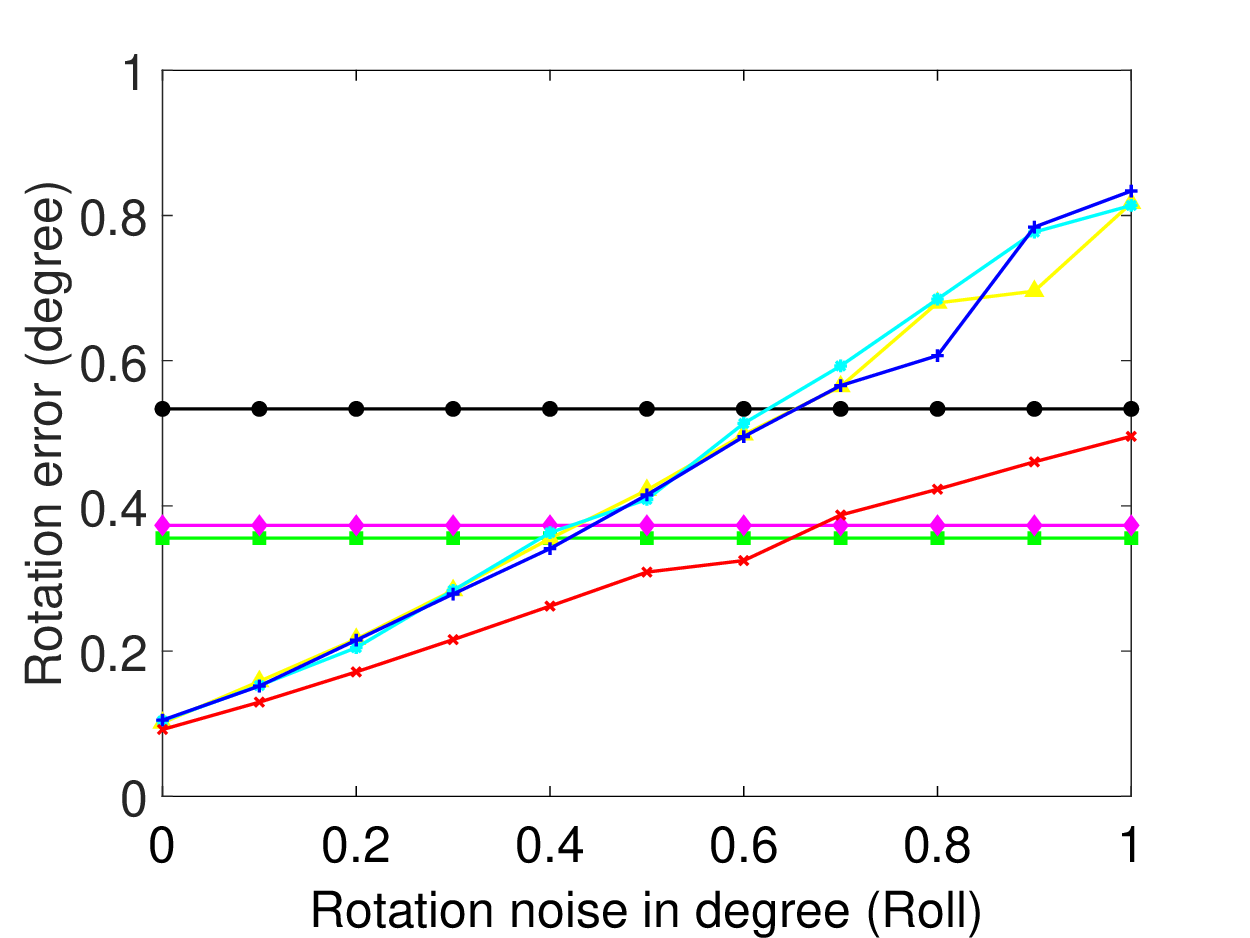}
		}
		\subfigure[\centering ${\varepsilon_{\bf{R}}}$]
		{
		\includegraphics[width=0.31\linewidth]{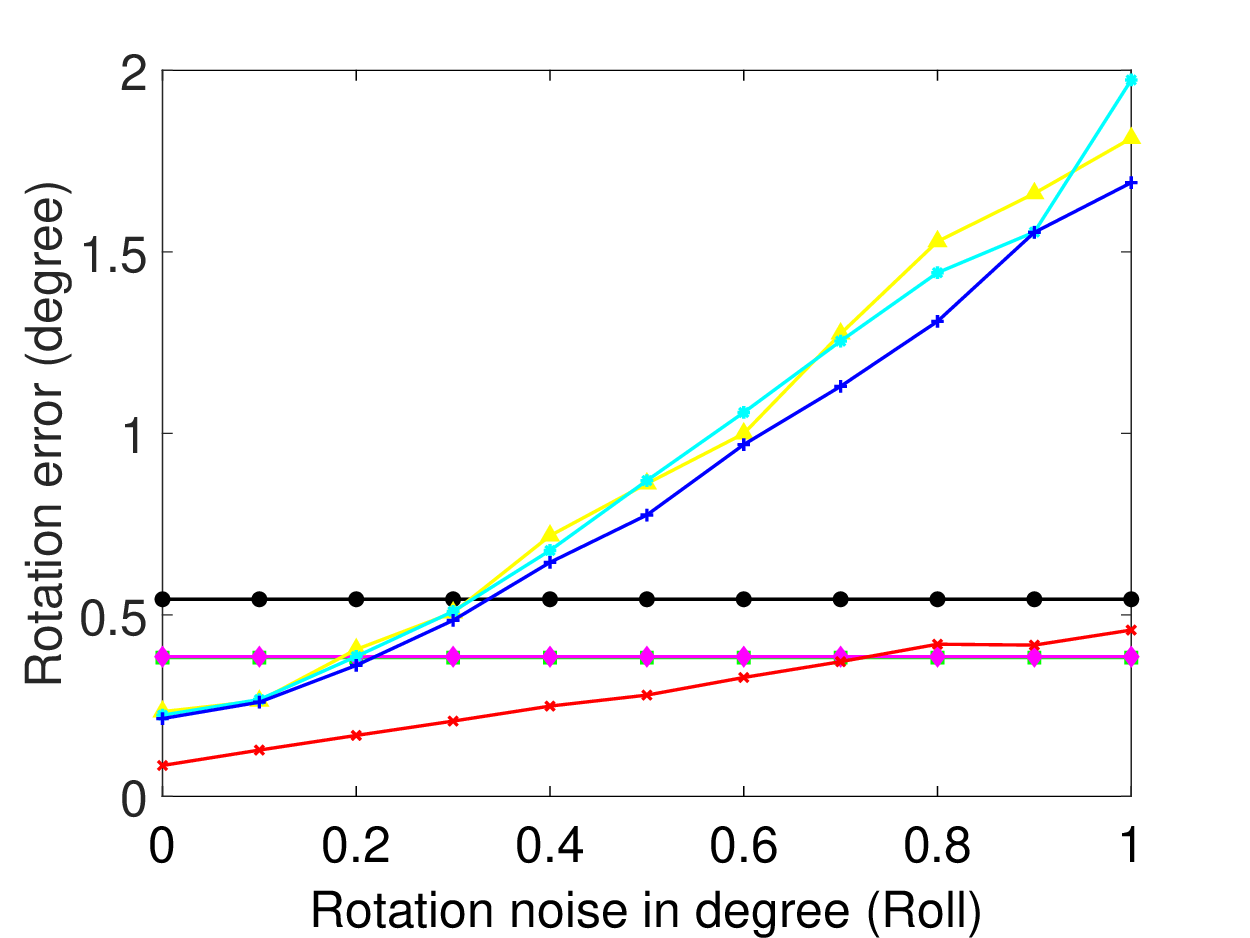}
		}
        \\
\subfigure[\centering $\varepsilon_{\mathbf{t},\text{dir}}$]
		{
		\includegraphics[width=0.31\linewidth]{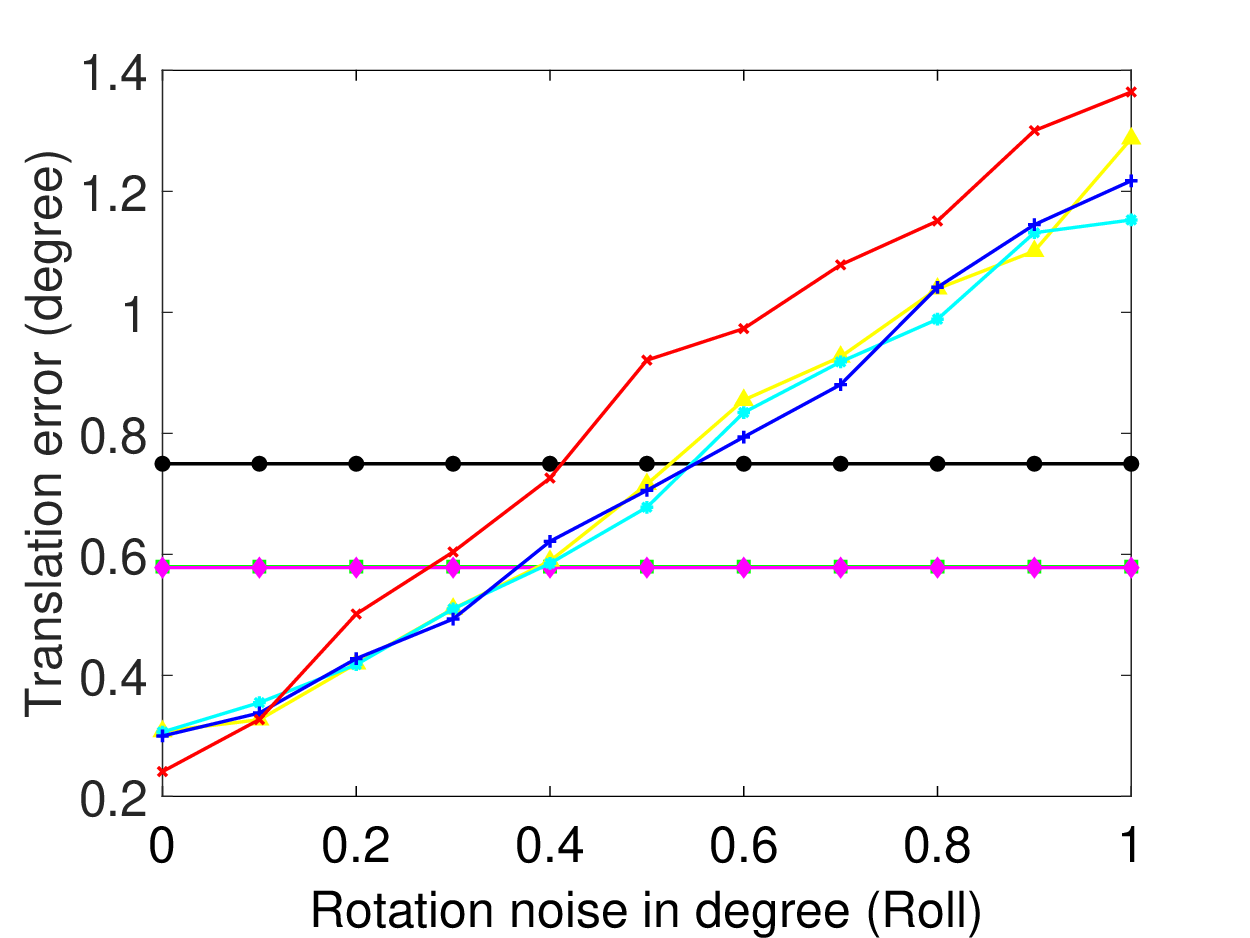}
		}
\subfigure[\centering $\varepsilon_{\mathbf{t},\text{dir}}$]
		{
		\includegraphics[width=0.31\linewidth]{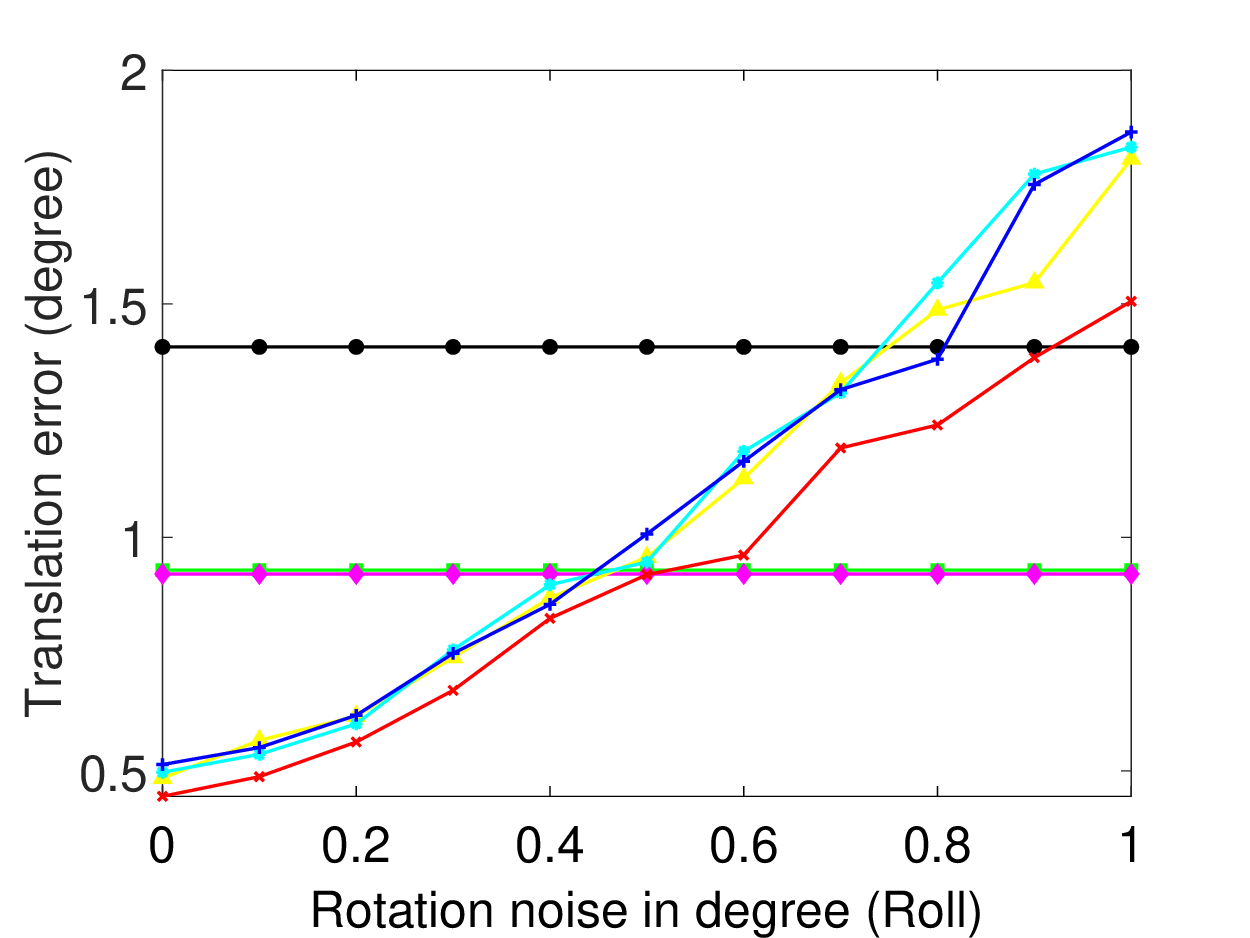}
		}
\subfigure[\centering $\varepsilon_{\mathbf{t},\text{dir}}$]
		{
		\includegraphics[width=0.31\linewidth]{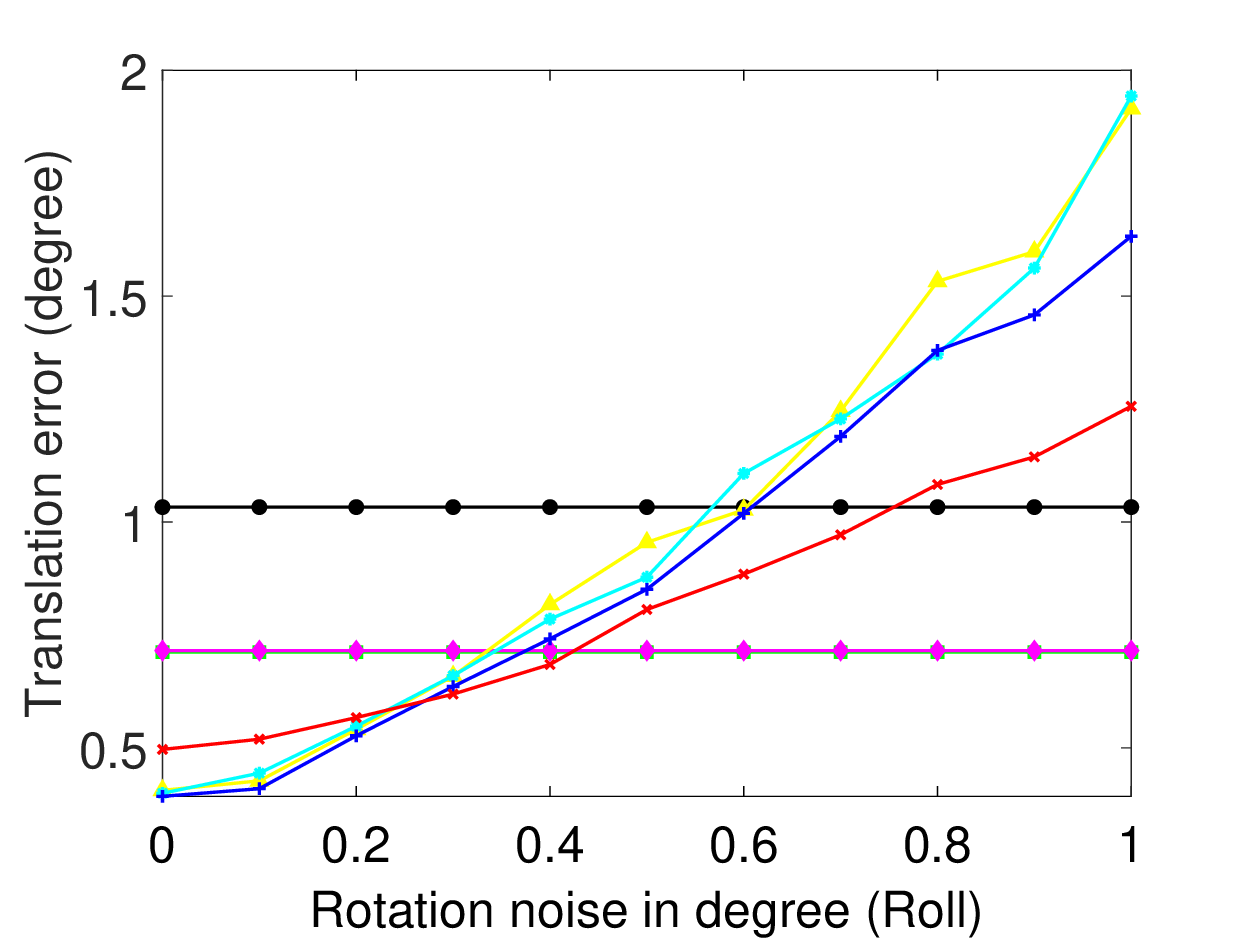}
		}
	
	\caption{Rotation and translation errors for multi-camera systems with increasing IMU roll angle noise. The first, second, and third columns present the performance of different solvers under forward, random, and sideways motions, respectively.}
	\label{fig:roll_angle_noise}
\end{figure}

Fig.~\ref{fig:roll_angle_noise} shows similar trends for IMU roll angle noise. It shows that 4-point methods outperform 6-DoF solvers in low-noise conditions, maintaining higher accuracy until noise reaches $0.2^\circ$ for forward/sideways motion and $0.4^\circ$ for random motion. 
\texttt{4pt-Axis-Approx} maintains superior accuracy across all motion patterns except for forward translation, while \texttt{4pt-Approx}, \texttt{4pt-Liu}
\cite{liu2017robust}, and \texttt{4pt-Lee}~\cite{hee2014relative} exhibit similar resistance against IMU roll angle noise, further confirming the overall robustness of these methods under diverse angular noise conditions.

\subsubsection{\label{sec:plane}Planar motion estimation}
This section evaluates pose estimation under strictly constrained planar motion, where the platform undergoes only yaw rotation and horizontal translation. In addition to the multi‑camera solvers, we include the monocular planar method \texttt{2pt‑Choi} \cite{choi2018fast} for comparison. Note that \texttt{2pt‑Choi} \cite{choi2018fast} estimates the motion of the individual camera rather than the full platform motion.

\begin{figure}[H]
	\centering
		\includegraphics[width=0.7\linewidth]{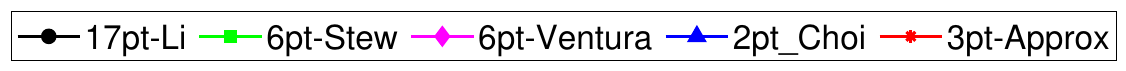}\\
		\subfigure[\centering $\varepsilon_{\mathbf{R}}$]
		{
		\includegraphics[width=0.31\linewidth]{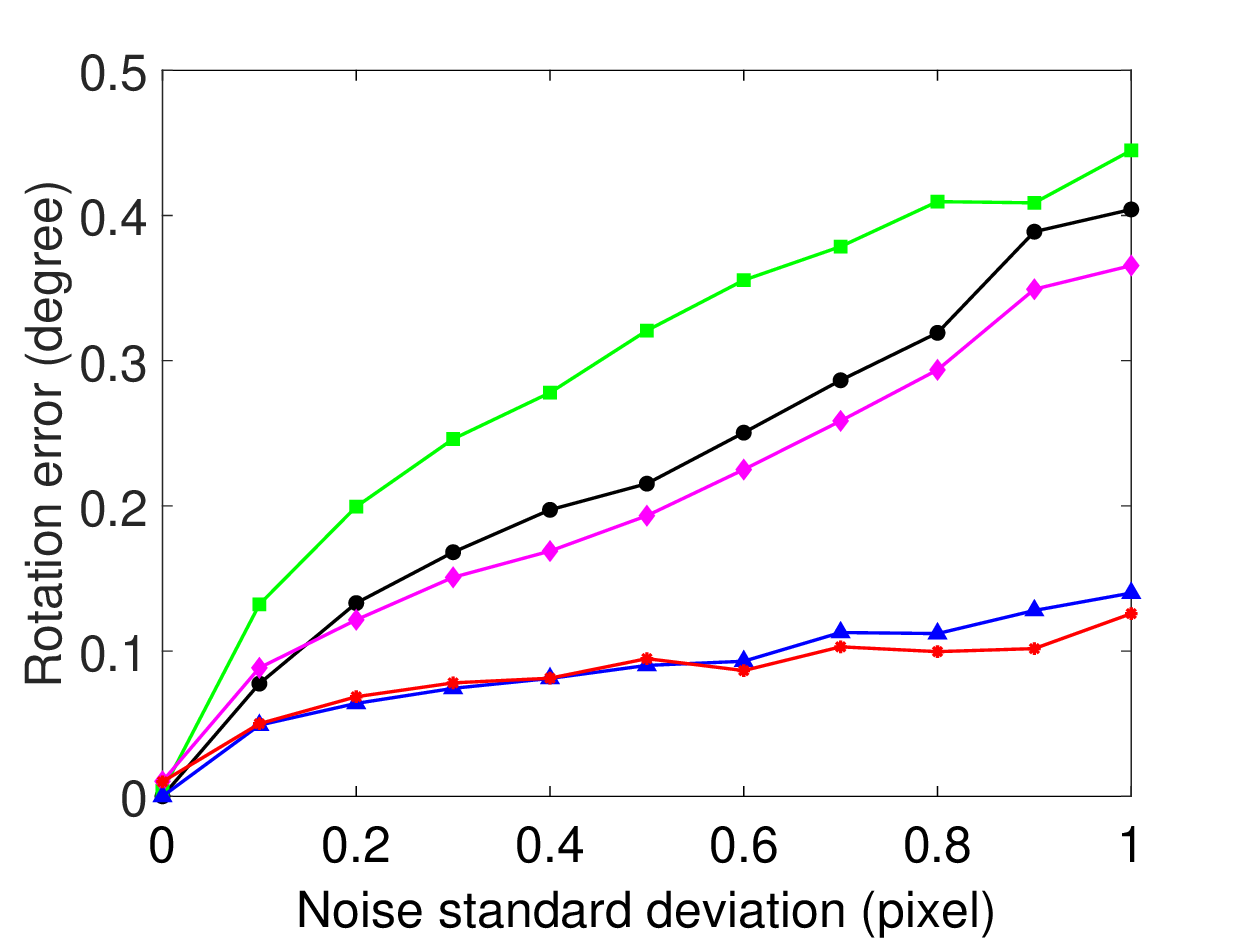}
		}
    \subfigure[\centering ${\varepsilon_{\bf{R}}}$]
		{
		\includegraphics[width=0.31\linewidth]{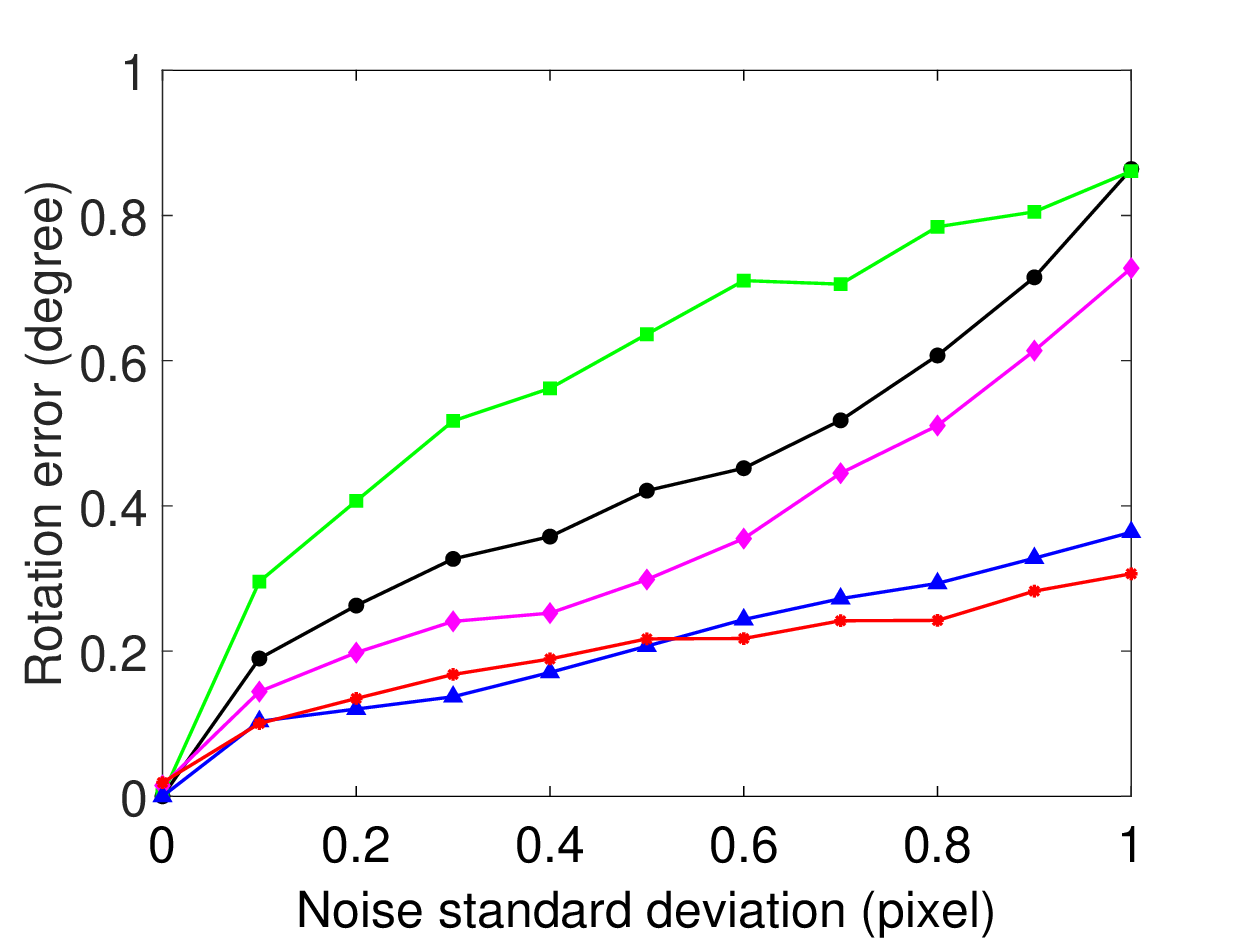}
		}
\subfigure[\centering ${\varepsilon_{\bf{R}}}$]
		{
		\includegraphics[width=0.31\linewidth]{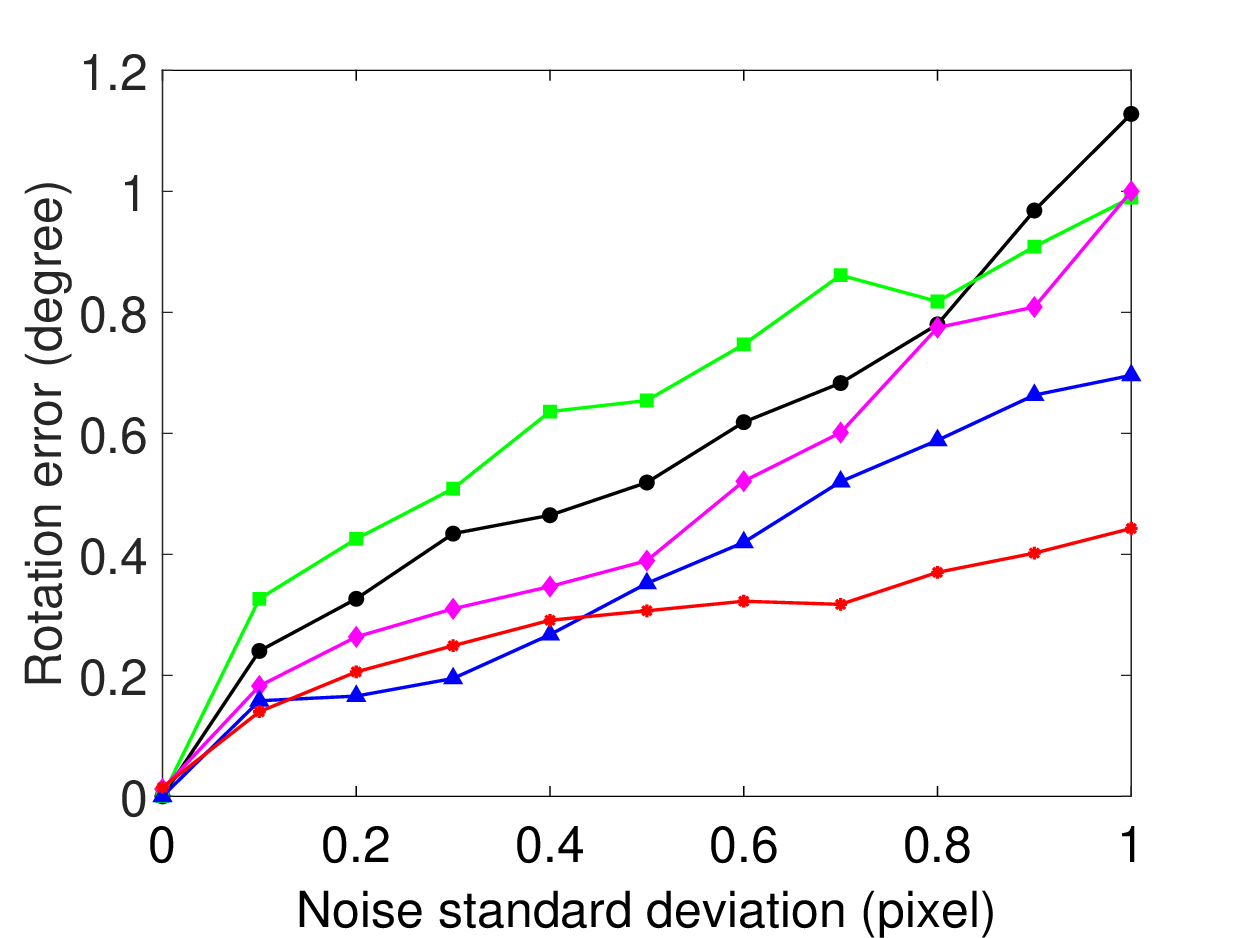}
		}
\\ \vspace{-10pt}
		\subfigure[\centering $\varepsilon_{\mathbf{t},\text{dir}}$]
		{
		\includegraphics[width=0.31\linewidth]{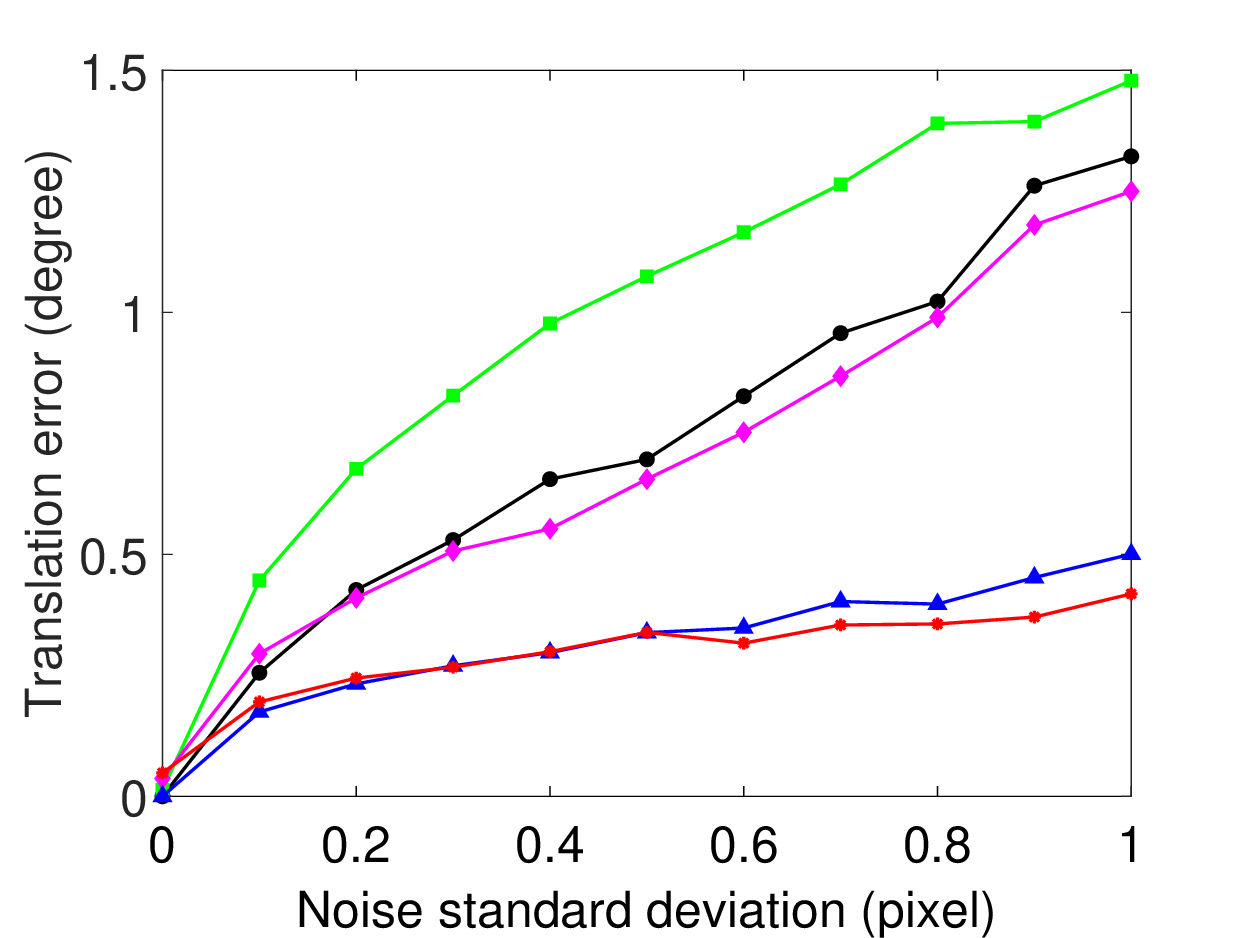}
		}	 
		\subfigure[\centering $\varepsilon_{\mathbf{t},\text{dir}}$]
		{
		\includegraphics[width=0.31\linewidth]{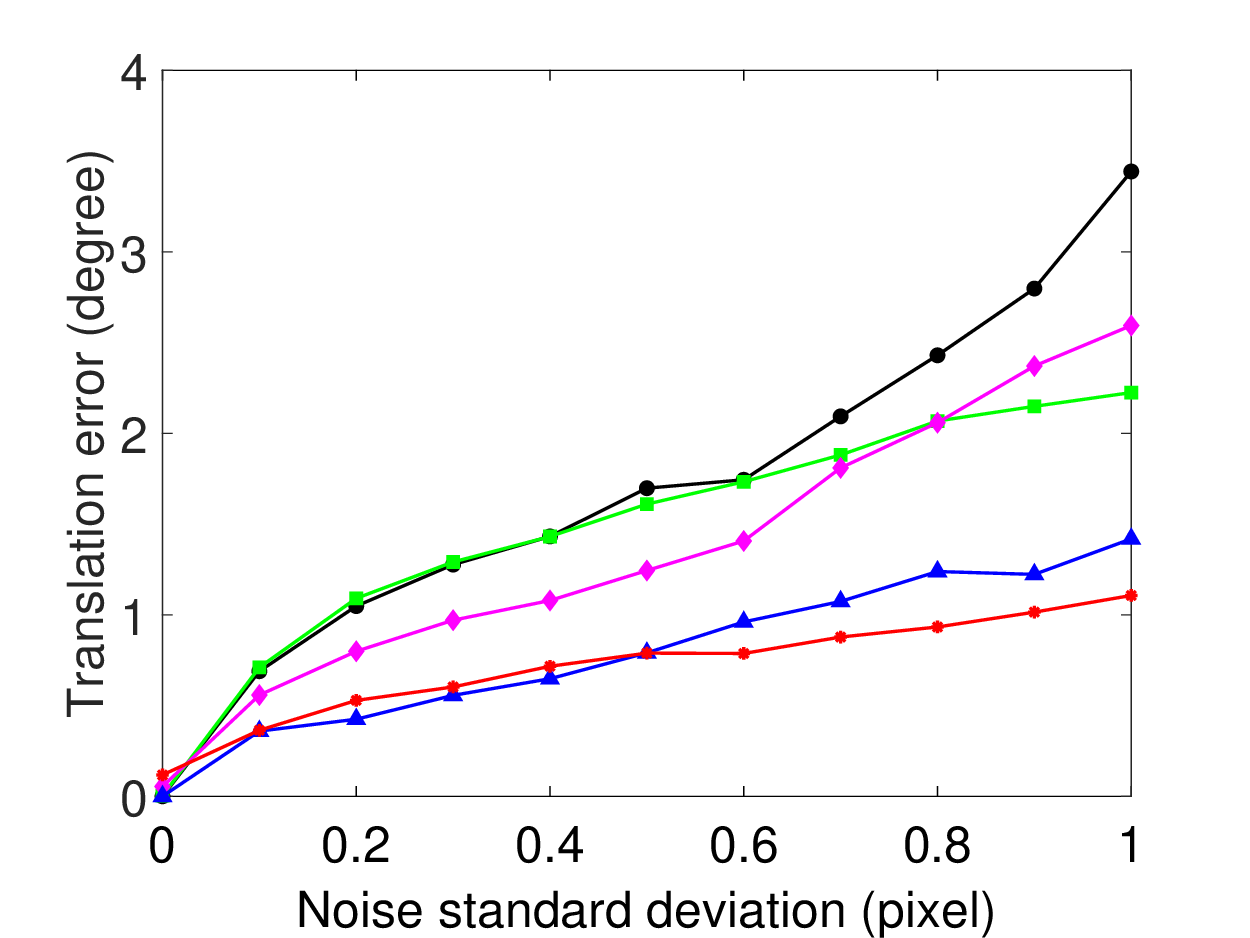}
		}	 		
	\subfigure[\centering $\varepsilon_{\mathbf{t},\text{dir}}$]
		{
		\includegraphics[width=0.31\linewidth]{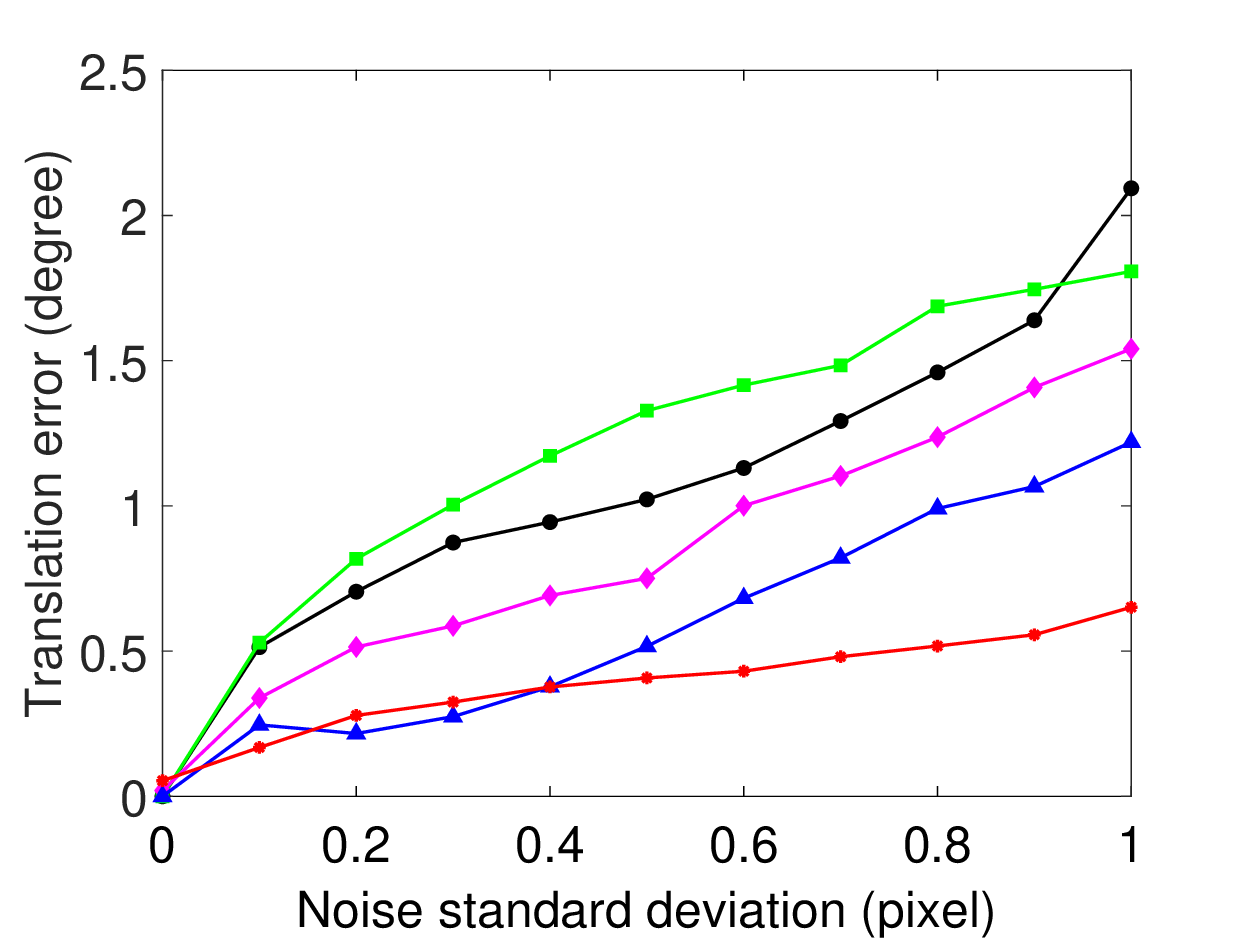}
		}
	
	\caption{Rotation and translation errors for multi-camera systems with increasing image noise under planar motions. The first, second, and third columns present the performance of different solvers under forward, random, and sideways motions, respectively.}
	\label{fig:image_noise_plane}
\end{figure}

\textbf{1) Image Noise Evaluation.} 
Image noise was introduced in the same manner as in the non‑planar case. Results in Fig.~\ref{fig:image_noise_plane} demonstrate that the proposed \texttt{3pt-Approx} method exhibits excellent accuracy stability. When noise is below 0.4 pixels, its rotation and translation errors are comparable to those of \texttt{2pt-Choi}, and significantly outperform \texttt{17pt-Li} and \texttt{6pt-Stew}. As noise exceeds 0.4 pixels, errors increase for all methods, but 3pt‑Approx degrades more slowly and achieves the smallest overall error, demonstrating superior robustness to image noise.
\begin{figure}[H]
	\centering
		\includegraphics[width=0.7\linewidth]{fig/simulation/plane_lengend.pdf}
        \\
		\subfigure[\centering $\varepsilon_{\mathbf{R}}$]
		{
		\includegraphics[width=0.31\linewidth]{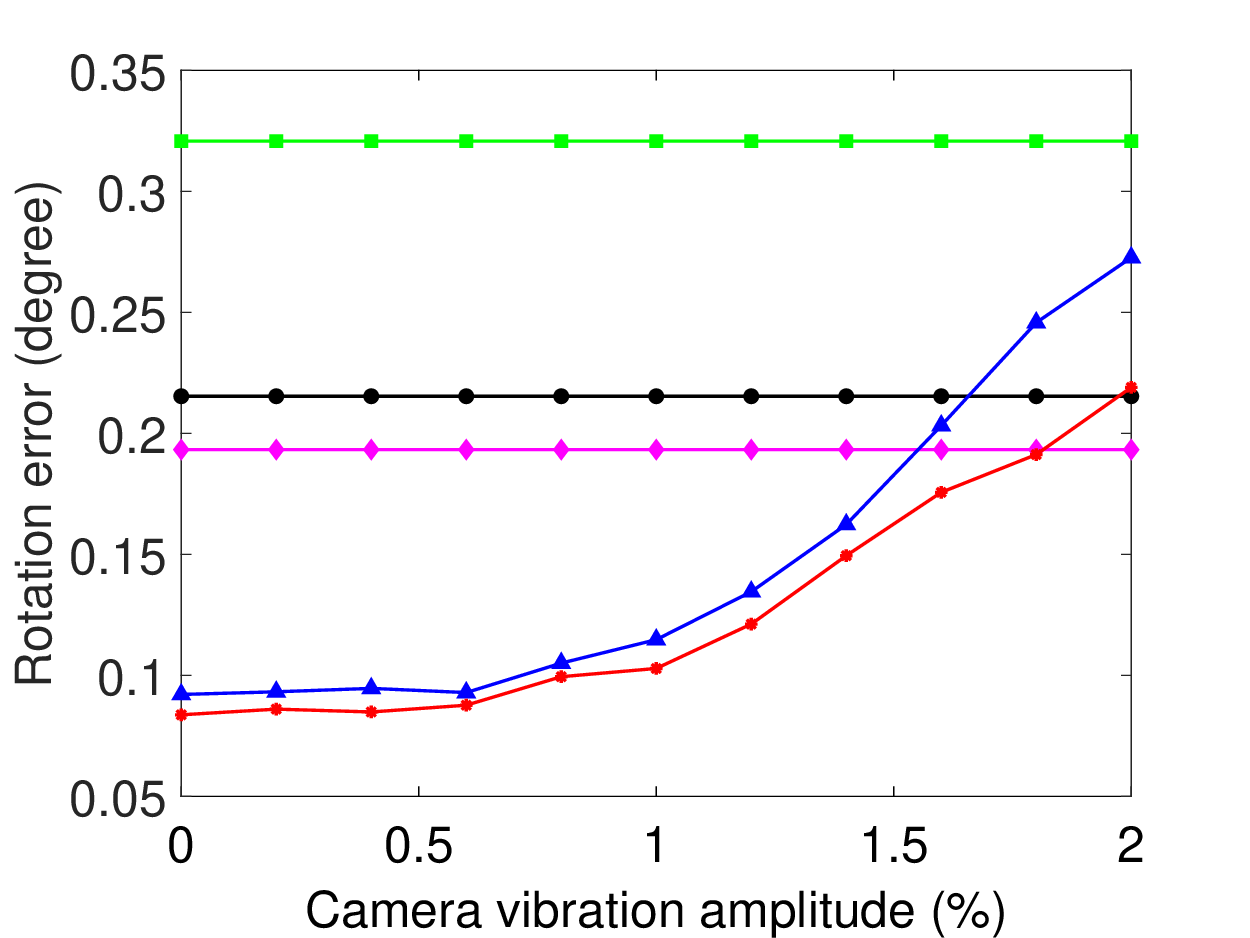}
		}
    \subfigure[\centering ${\varepsilon_{\bf{R}}}$]
		{
		\includegraphics[width=0.31\linewidth]{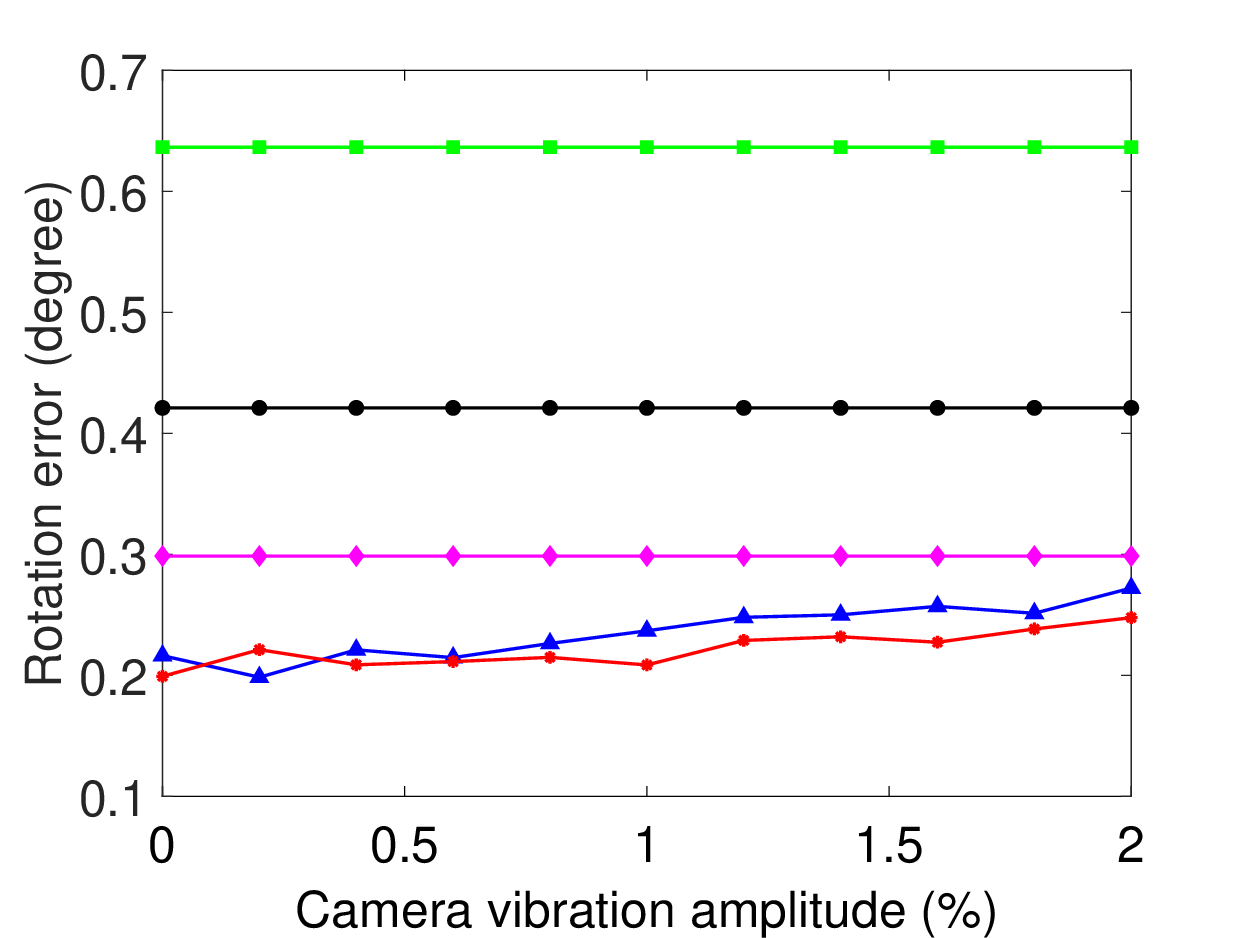}
		}
\subfigure[\centering ${\varepsilon_{\bf{R}}}$]
		{
		\includegraphics[width=0.31\linewidth]{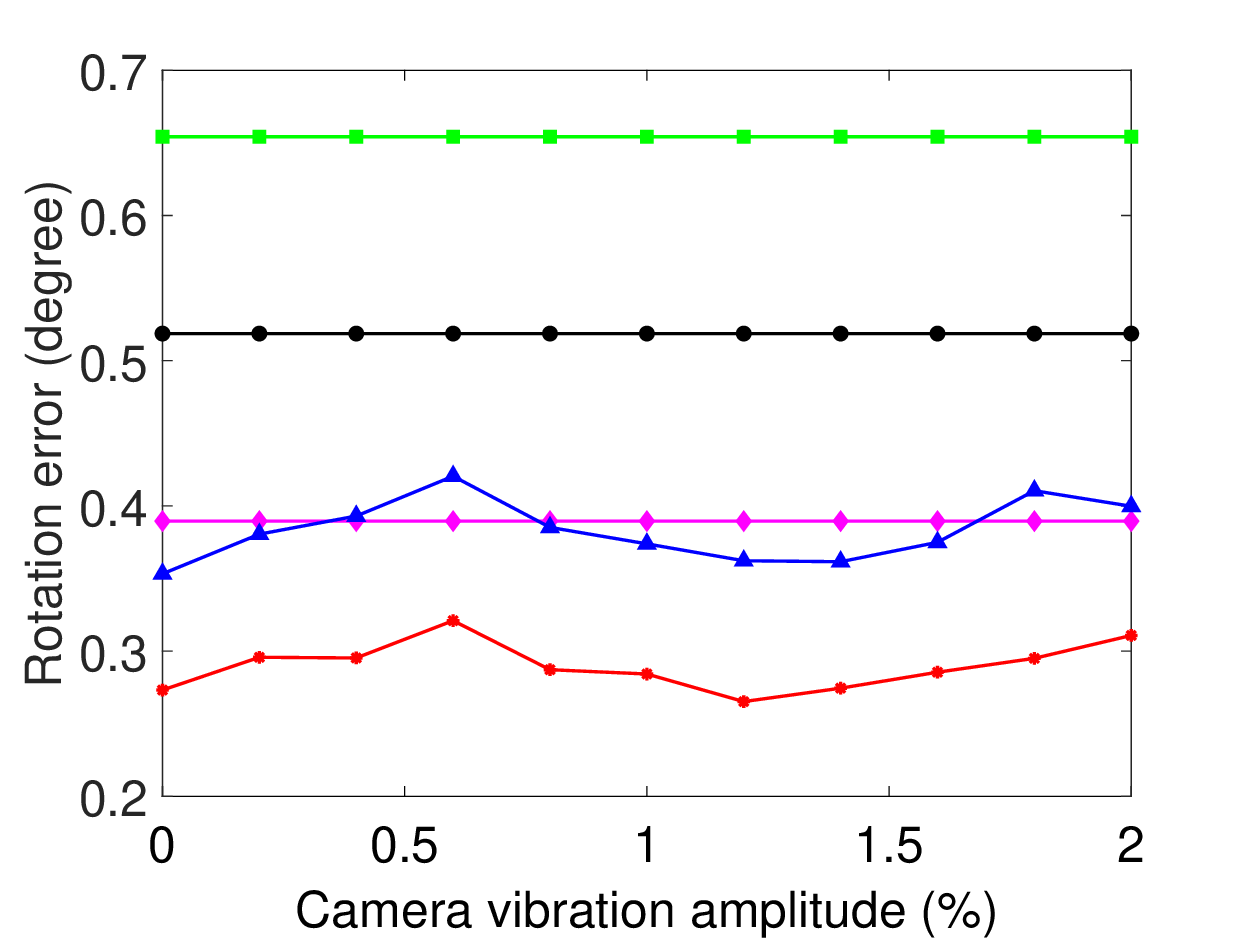}
		}
\\ \vspace{-10pt}
		\subfigure[\centering $\varepsilon_{\mathbf{t},\text{dir}}$]
		{
		\includegraphics[width=0.31\linewidth]{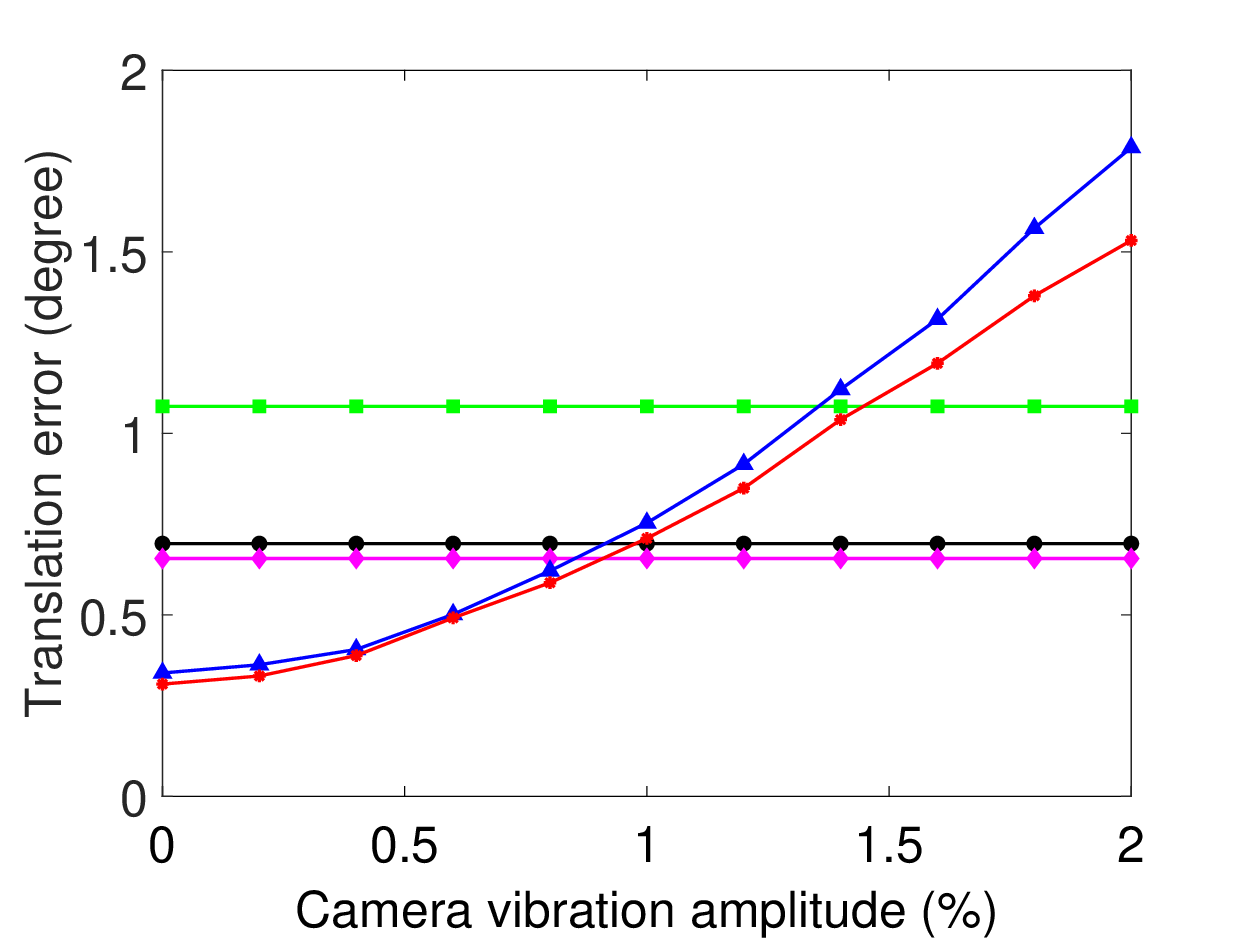}
		}
		\subfigure[\centering $\varepsilon_{\mathbf{t},\text{dir}}$]
		{
		\includegraphics[width=0.31\linewidth]{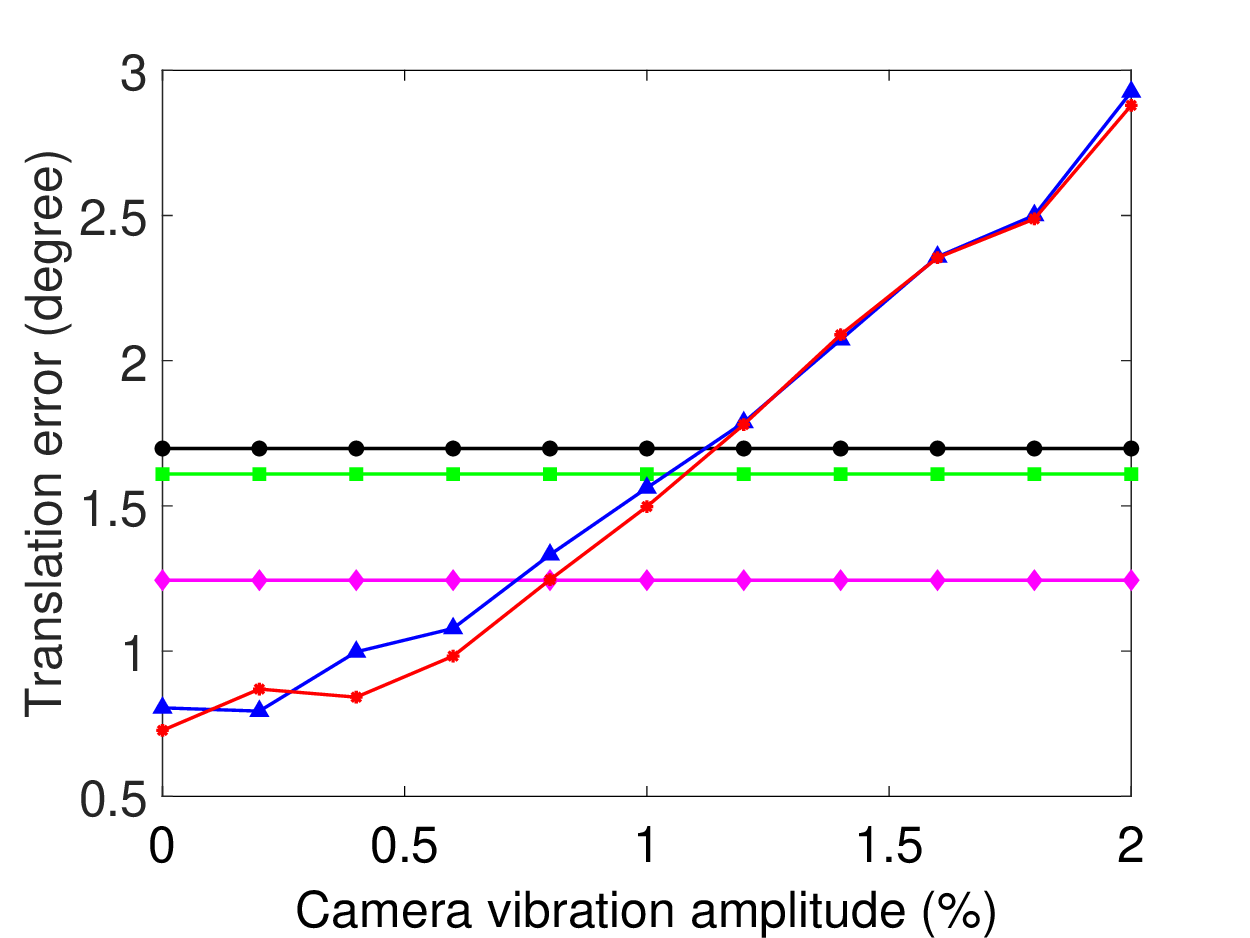}
		}	 		
	\subfigure[\centering $\varepsilon_{\mathbf{t},\text{dir}}$]
		{
		\includegraphics[width=0.31\linewidth]{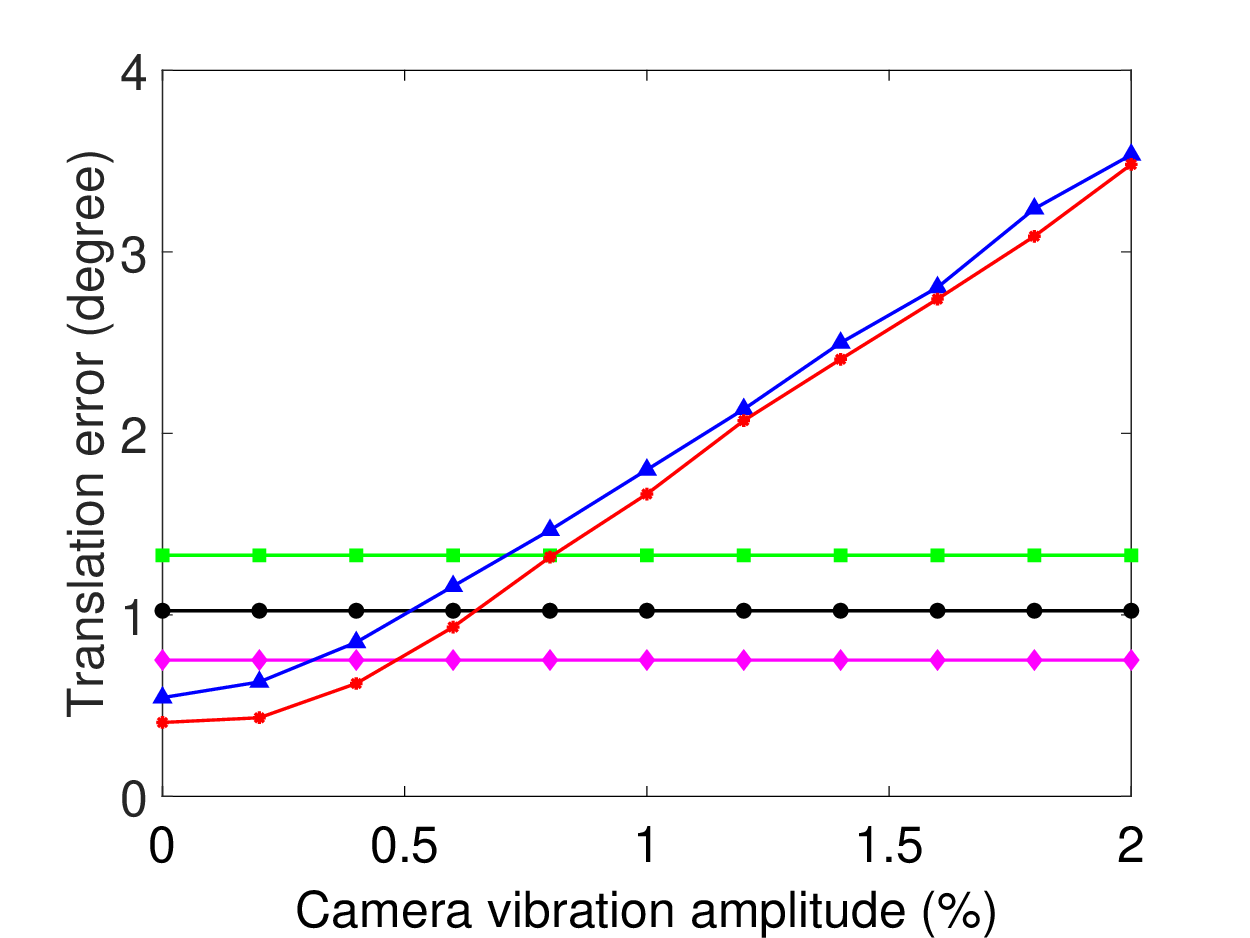}
		}
	
	\caption{Rotation and translation errors for multi-camera systems with increasing vibration noise. The first, second, and third columns present the performance of different solvers under forward, random, and sideways motions, respectively.}
	\label{fig:vibrations}
\end{figure}

\textbf{2) Vibration Noise Evaluation.}
In real-world scenarios, moving platforms often experience minor vibrations due to uneven road surfaces or mechanical oscillations. To evaluate this effect, we modeled vibration magnitude as a percentage of the platform's translation norm $||\mathbf{t}||$, ranging from $0\%$ to $2\%$, while keeping image noise fixed at 0.5 pixels. Note that general 6-DoF solvers are inherently unaffected by this perturbation, as they do not assume planar motion. 
 In typical driving conditions, vertical vibration displacement is negligible compared to the travel distance, usually below $1\%$ of $||\mathbf{t}||$.

As shown in Fig.~\ref{fig:vibrations}, the proposed \texttt{3pt-Approx} method maintains superior performance over general 6-DoF solvers within the realistic vibration range. Though both \texttt{3pt-Approx} and \texttt{2pt-Choi} exhibit a gradual decline in accuracy with increased vibration, their errors remain acceptable. Notably, \texttt{3pt-Approx} consistently achieves higher precision than \texttt{2pt-Choi} across most test conditions, highlighting its robustness and practical suitability for automotive applications.

\subsection{Experiments on Real Data}
\begin{figure}[H]
  \centering
     \subfigure[\centering View 1]{
     \centering
     \includegraphics[width=0.49\linewidth]{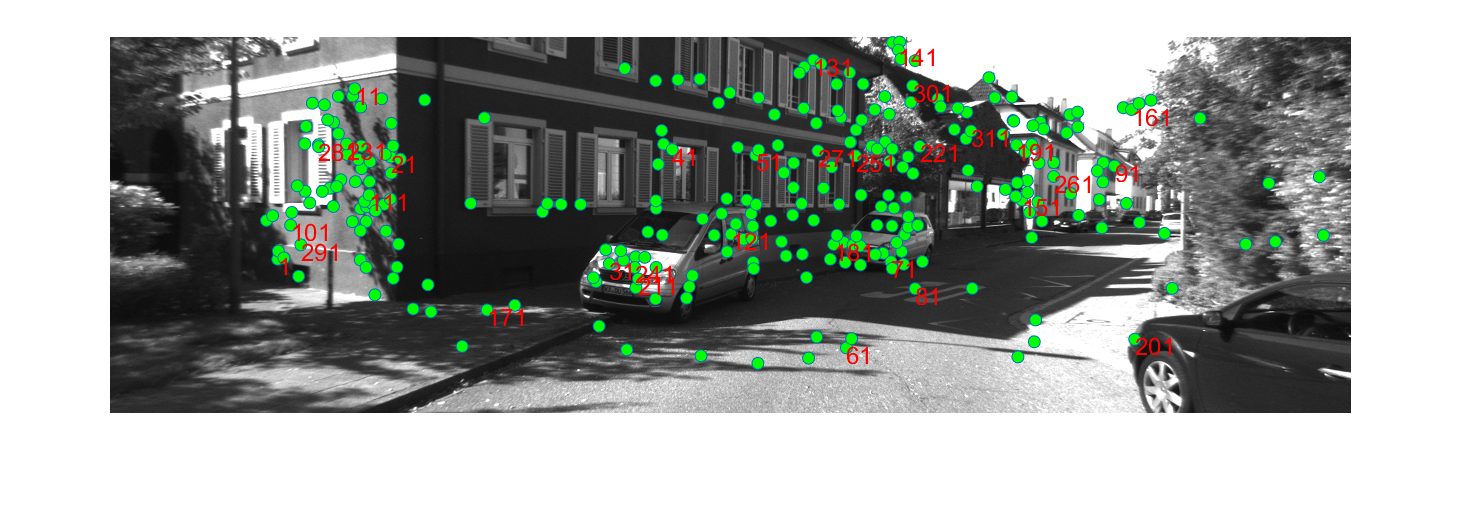}
     \label{fig.MATCH1}
      }%
      \subfigure[\centering View 2]{
     \centering
     \includegraphics[width=0.49\linewidth]{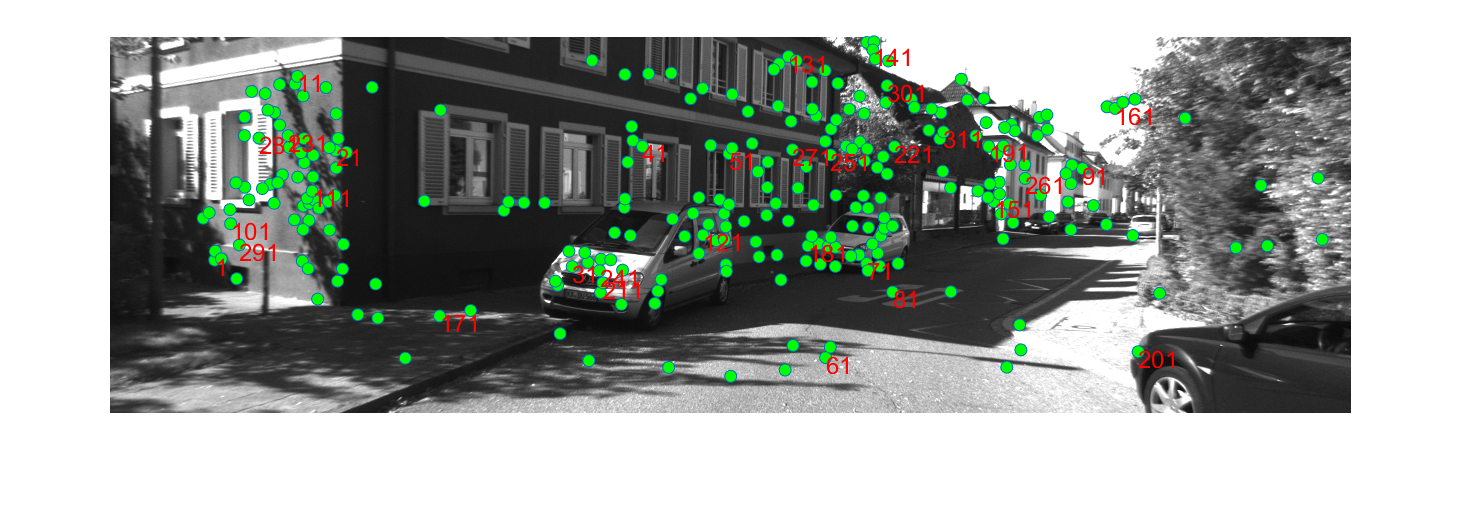}
     \label{fig.MATCH2}
      }%
          \centering
    \caption{Point correspondences between between consecutive frames. The images were extracted from frames 3000-3001 of sequence 00 in the KITTI dataset.}
    \label{fig:point_correspondences}
\end{figure}

For real-world evaluation, we employed the KITTI dataset~\cite{geiger2013vision}, a standard benchmark for autonomous driving research. Sequences 00-10 include GPS/IMU ground‑truth data, and we evaluated our method across all 11 sequences, covering approximately 23,000 image pairs.
 Fig.~\ref{fig:point_correspondences} illustrates an example with 626 successfully matched feature points, marked in green dots. Correspondences are labeled with red numbers, only a subset is shown to preserve visual clarity.

\begin{table}[H]
	\caption{Average runtime comparison of relative pose estimation solvers within RANSAC framework on KITTI dataset (unit:~$s$).}
	\centering
		\setlength{\tabcolsep}{4pt}{
			\scalebox{1}{
				\begin{tabular}{cccccccccc}
					\hline					
	Method & \begin{tabular}[c]{@{}c@{}}17pt-Li\\ ~\cite{li2008linear}\end{tabular} & \begin{tabular}[c]{@{}c@{}}6pt-St.\\ ~\cite{henrikstewenius2005solutions}\end{tabular} & \begin{tabular}[c]{@{}c@{}}6pt-Ven.\\ ~\cite{ventura2015efficient}\end{tabular} & \begin{tabular}[c]{@{}c@{}}4pt-Lee\\ ~\cite{hee2014relative}\end{tabular} & \begin{tabular}[c]{@{}c@{}}4pt-Liu\\ ~\cite{liu2017robust}\end{tabular}  & \begin{tabular}[c]{@{}c@{}}4pt-\\Approx \end{tabular} & \begin{tabular}[c]{@{}c@{}}4pt-Axis-\\Approx \end{tabular} & \begin{tabular}[c]{@{}c@{}}3pt-\\Approx \end{tabular} \\  \hline 
Runtime &0.7764 & 23.4727 & 6.4544 & 1.0483 & 0.6316 & 1.0424 & \textbf{0.4772} & 1.2945  \\ 
					\hline
		\end{tabular}}}
	\label{tab:SolverTime_4pt_ransac}
\end{table}
Following the simulation setup, all methods were embedded in the RANSAC framework. 
Table~\ref{tab:SolverTime_4pt_ransac} shows the average runtime comparison of relative pose estimation solvers within the RANSAC framework on the KITTI dataset.
The results indicate that our proposed methods achieve high computational efficiency, with \texttt{4pt-Axis-Approx} achieving the fastest runtime. 
Unlike the other methods, the iteration counts for the 6-DoF solvers, namely \texttt{6pt-Stew}~\cite{henrikstewenius2005solutions}, \texttt{17pt-Li}~\cite{li2008linear}, and \texttt{6pt-Ventura}~\cite{ventura2015efficient}, reached the predefined maximum limit. The reason \texttt{17pt-Li}~\cite{li2008linear} runs faster than \texttt{4pt-Lee}~\cite{hee2014relative}, \texttt{4pt-Approx}, and \texttt{3pt-Approx} is that it returns a single solution, whereas the others require additional verification for multiple solutions. 

\begin{table}[H]
    \caption{ Median rotation errors for KITTI sequences (unit: degree)}
    \label{tab:realR}
        \setlength\tabcolsep{6pt}
    \centering
\begin{tabular}{cccccccccc}
\hline
Seq. & \begin{tabular}[c]{@{}c@{}}17pt-Li\\ ~\cite{li2008linear}\end{tabular} & \begin{tabular}[c]{@{}c@{}}6pt-St.\\ ~\cite{henrikstewenius2005solutions}\end{tabular} & \begin{tabular}[c]{@{}c@{}}6pt-Ven.\\ ~\cite{ventura2015efficient}\end{tabular} & \begin{tabular}[c]{@{}c@{}}4pt-Lee\\ ~\cite{hee2014relative}\end{tabular} & \begin{tabular}[c]{@{}c@{}}4pt-Liu\\ ~\cite{liu2017robust}\end{tabular}  & \begin{tabular}[c]{@{}c@{}}4pt-\\Approx \end{tabular} & \begin{tabular}[c]{@{}c@{}}4pt-Axis-\\Approx \end{tabular} & \begin{tabular}[c]{@{}c@{}}3pt-\\Approx \end{tabular} \\  \hline 
00    & 0.0928 & 0.1924 & 0.0595 & \textbf{0.0308} & 0.0315 & 0.0333 & 0.0369 & 0.0347 \\
    01    & 0.0998 & 0.458 & 0.0497 & 0.0368 & 0.0375 & 0.0438 & \textbf{0.0284} & 0.0519 \\
    02    & 0.0826 & 0.1801 & 0.0545 & 0.0289 & \textbf{0.0287} & 0.0297 & 0.0345 & 0.0329 \\
    03    & 0.0762 & 0.2207 & 0.0516 & \textbf{0.0360} & 0.0361 & 0.0384 & 0.0384 & 0.0387 \\
    04    & 0.0737 & 0.2037 & 0.0434 & 0.0220 & 0.0232 & \textbf{0.0212} & 0.0229 & 0.0268 \\
    05    & 0.0742 & 0.1968 & 0.0456 & 0.0268 & 0.0270 & \textbf{0.0266} & 0.0283 & 0.0292 \\
    06    & 0.0693 & 0.2059 & 0.0421 & 0.0264 & 0.0264 & \textbf{0.0239} & 0.0249 & 0.0299 \\
    07    & 0.0708 & 0.1912 & 0.0497 & 0.0270 & \textbf{0.0269} & 0.0292 & 0.0299 & 0.0297 \\
    08    & 0.0777 & 0.1901 & 0.0498 & \textbf{0.0254} & 0.0255 & 0.0284 & 0.0317 & 0.0292 \\
    09    & 0.0857 & 0.1823 & 0.0541 & \textbf{0.0277} & 0.0289 & 0.0309 & 0.0323 & 0.0339 \\
    10    & 0.0885 & 0.1935 & 0.0572 & \textbf{0.0276} & 0.0281 & 0.0279 & 0.0331 & 0.0326 \\ \hline
\end{tabular}
 \end{table}

\begin{table}[H]
    \caption{ Median translation direction errors for KITTI sequences (unit: degree)}
    \label{tab:realT}
        \setlength\tabcolsep{6pt}
    \centering
\begin{tabular}{cccccccccc}
\hline
Seq. & \begin{tabular}[c]{@{}c@{}}17pt-Li\\ ~\cite{li2008linear}\end{tabular} & \begin{tabular}[c]{@{}c@{}}6pt-St.\\ ~\cite{henrikstewenius2005solutions}\end{tabular} & \begin{tabular}[c]{@{}c@{}}6pt-Ven.\\ ~\cite{ventura2015efficient}\end{tabular}& \begin{tabular}[c]{@{}c@{}}4pt-Lee\\ ~\cite{hee2014relative}\end{tabular} & \begin{tabular}[c]{@{}c@{}}4pt-Liu\\ ~\cite{liu2017robust}\end{tabular}  & \begin{tabular}[c]{@{}c@{}}4pt-\\Approx \end{tabular} & \begin{tabular}[c]{@{}c@{}}4pt-Axis-\\Approx \end{tabular} & \begin{tabular}[c]{@{}c@{}}3pt-\\Approx \end{tabular} \\  \hline 
00    & 1.8917 & 3.4859 & 1.1230 & 1.1374 & 1.1186 & 1.2419 & \textbf{1.1088} & 1.5012 \\
    01    & 3.9064 & 6.8551 & 1.3609 & 1.3383 & 1.3408 & 2.2743 & \textbf{1.2823} & 2.385 \\
    02    & 1.4328 & 2.3983 & \textbf{0.9234} & 0.9630 & 0.9621 & 1.0481 & 1.0051 & 1.4019 \\
    03    & 2.1875 & 6.3038 & 1.3101 & \textbf{1.2063} & 1.2508 & 1.3521 & 1.2170 & 1.4543 \\
    04    & 1.4304 & 3.1938 & 0.6006 & \textbf{0.5948} & 0.6084 & 0.6400  & 0.6295 & 1.4000 \\
    05    & 1.6829 & 3.6314 & 0.7776 & 0.8465 & 0.7951 & 0.8900  & \textbf{0.7687} & 1.4447 \\
    06    & 1.5178 & 3.0878 & 0.6101 & 0.6235 & 0.6330 & 0.6432 & \textbf{0.5544} & 1.2872 \\
    07    & 2.1587 & 4.719 & 1.1337 & 1.0859 & \textbf{1.016} & 1.2362 & 1.1084 & 1.6828 \\
    08    & 2.0057 & 4.0091 & 1.2802 & \textbf{1.2133} & 1.2476 & 1.3881 & 1.2634 & 1.8108 \\
    09    & 1.4466 & 2.7625 & 0.8632 & \textbf{0.7735} & 0.7972 & 0.9051 & 0.8708 & 1.4341 \\
    10    & 1.6177 & 3.0233 & 0.9877 & 0.9822 & 0.949 & 1.0758 & \textbf{0.9462} & 1.5479 \\ \hline
\end{tabular}
 \end{table}

The rotation and translation errors are shown in Table~\ref{tab:realR} and Table~\ref{tab:realT}, respectively. The 4-point solvers substantially outperform all the 6-DoF solvers in rotation accuracy, and similarly surpass both \texttt{17pt-Li}~\cite{li2008linear} and \texttt{6pt-Stew}~\cite{henrikstewenius2005solutions} in translation accuracy.
Among the 4-point methods, \texttt{4pt-Lee}~\cite{hee2014relative} achieves slightly higher rotation accuracy in partial sequences. In contrast, our proposed \texttt{4pt-Axis-Approx} method performs marginally better in translation, achieving superior accuracy in sequences 00, 01, 05, 06, and 10. It is worth noting that before applying the \texttt{3pt-Approx} algorithm, we aligned the image data with approximately planar motion to ensure local consistency with the planar model. After this alignment, \texttt{3pt-Approx} achieves rotation accuracy comparable to that of the 4-point solvers, and its translation accuracy also exceeds that of both \texttt{17pt-Li}~\cite{li2008linear} and \texttt{6pt-Stew}~\cite{henrikstewenius2005solutions}.

\section{Conclusion}\label{Conclusion}
In this paper, we introduce a unified framework for efficient relative pose estimation, based on a novel translation parameterization and first-order rotation approximation. Within this framework, we develop three minimal solvers tailored to typical automotive scenarios: one incorporating IMU-based vertical direction, one using rotation axis direction, and one assuming planar vehicle motion.
While designed for autonomous driving, these solvers are also applicable to other structured navigation contexts, such as indoor mobile robots, gravity-aligned aerial vehicles, or platforms with constrained kinematics. In unstructured or highly dynamic environments, general 6-DoF solvers remain necessary, our methods offer an efficient alternative when motion priors are available.
While the proposed methods achieve notable efficiency and accuracy, they are inherently dependent on the precision of IMU measurements or the validity of planar motion assumptions. Future work will focus on developing more adaptive solvers that can relax these dependencies, for instance by estimating the vertical direction or rotation axis direction from visual information when IMU data is noisy or unavailable.
Overall, our solution offers an efficient and effective baseline for real-time pose estimation in autonomous driving scenarios. Extensive evaluation on both synthetic and real-world data confirms that our approach achieves high computational speed while maintaining accuracy comparable to state-of-the-art multi-camera ego-motion estimation methods.

\section*{Disclosures}
The authors declare no conflicts of interest.
\section*{Data Availability}
Data underlying the results presented in this paper are not publicly available at this time but may be obtained from the authors upon reasonable request.
\bibliography{references}
\end{document}